\documentclass[journal]{IEEEtran}

\usepackage{hyperref}       
\usepackage{url}            
\usepackage{booktabs}       
\usepackage{amsfonts}       
\usepackage{nicefrac}       
\usepackage{microtype}      
\usepackage{color}
\usepackage{amsmath}
\usepackage{cite}
\usepackage{amsmath,amssymb,amsfonts,amsthm}
\usepackage{bm,bbm}
\usepackage{algorithm}
\usepackage{algpseudocode}
\usepackage{footnote}
\usepackage{comment}
\usepackage{multirow}
\usepackage{graphicx}
\usepackage{epstopdf}
\usepackage{caption,subcaption}
\usepackage{lipsum}
\usepackage{mwe}
\usepackage[normalem]{ulem}
\allowdisplaybreaks
\def\BibTeX{{\rm B\kern-.05em{\sc i\kern-.025em b}\kern-.08em
    T\kern-.1667em\lower.7ex\hbox{E}\kern-.125emX}}

\algrenewcommand\algorithmicrequire{\textbf{Input:}}
\algrenewcommand\algorithmicensure{\textbf{Initialize:}}

\newtheorem{assumption}{Assumption}
\newtheorem{definition}{Definition}
\newtheorem{lemma}{Lemma}
\newtheorem{theorem}{Theorem}
\ifCLASSINFOpdf
\else
\fi

\begin{document}

\title{Communication-Efficient {Federated}
Learning Using Censored Heavy Ball Descent \\
\thanks{The work was supported by the U. S. Army Research Laboratory and the U. S. Army Research Office under grant number W911NF-17-1-0331 and by the Office of Naval Research grant number  N00014-22-1-2626.}}

\author{Yicheng Chen, Rick S. Blum,
\IEEEmembership{Fellow,~IEEE}, and Brian M. Sadler, \IEEEmembership{Life Fellow,~IEEE}
\thanks{
Yicheng Chen is with Reality Labs Meta, Redmond, WA 98052 USA
(email: chenggejiayou@gmail.com).}
\thanks{
Rick S. Blum is with Lehigh University, Bethlehem, PA 18015 USA
(email: rblum@eecs.lehigh.edu).}
\thanks{
Brian M. Sadler is with DEVCOM ARL, Adelphi, MD 20878 USA (email: brian.m.sadler6.civ@mail.mil).}}


\maketitle

\begin{abstract}
Distributed machine learning enables scalability and computational offloading, but requires significant levels of communication. Consequently, communication efficiency in distributed learning settings is an important consideration, especially when the communications are wireless and battery-driven devices are employed. In this paper we develop a censoring-based heavy ball (CHB) method for distributed learning in a server-worker architecture.  Each worker self-censors unless its local gradient is sufficiently different from the previously transmitted one. The significant practical advantages of the HB method for learning problems are well known, but the question of reducing communications has not been addressed. CHB takes advantage of the HB smoothing to eliminate reporting small changes, 
and provably achieves a linear convergence rate equivalent to that of the classical HB method for smooth and strongly convex objective functions. The convergence guarantee of CHB is theoretically justified for both convex and nonconvex cases. In addition we prove that, under some conditions, at least half of all communications can be eliminated without any impact on convergence rate. Extensive numerical results validate the communication efficiency of CHB on both synthetic and real datasets, for convex, nonconvex, and nondifferentiable cases. Given a target accuracy, CHB can significantly reduce the number of communications compared to existing algorithms, achieving the same accuracy without slowing down the optimization process.
\end{abstract}


%
\IEEEpeerreviewmaketitle

\section{Introduction}

Traditional centralized optimization algorithms for machine learning applications have a number of limitations when applied in networked settings.  This is especially clear in some applications where abundant distributed processing resources are inherently available such as vehicular, mobile cellular, and sensor networks. This motivates distributed optimization that balances communications and processing. One popular distributed optimization problem can be formulated as
\begin{align}\label{basicprob}
\min _{\boldsymbol{\theta} \in \mathbb{R}^{d}} f(\boldsymbol{\theta}) \quad \text { with } \quad f(\boldsymbol{\theta}) \buildrel \Delta \over=\sum_{m \in \mathcal{M}} f_{m}(\boldsymbol{\theta})
\end{align}
where $\boldsymbol{\theta} \in \mathbb{R}^{d}$ is the model parameter vector to be optimized, $f(\boldsymbol{\theta})$ is the objective function to be minimized, and $f_{m}(\boldsymbol{\theta})$ is the local function for worker $m \in \mathcal{M}=\{1,2,...,M\}$. The problem in (\ref{basicprob}) has been successfully applied as a model in multi-agent systems \cite{liu2021approximate, turan2022robust,li2021distributed}, sensor networks \cite{zhang2015distributed, chen2019testing, chen2018ordered, chen2020optimal, chen2021ordering}, smart grids \cite{binetti2014distributed}, and distributed learning \cite{park2021communication, vlaski2021distributed, imteaj2021survey, chen2022communication}. \textcolor[rgb]{0.00,0.00,0.00}{To relate (\ref{basicprob}) to a distributed learning setting, let  $f_{m}(\boldsymbol{\theta})\buildrel \Delta \over=\sum_{n=1}^{N_m}\ell(\boldsymbol{\theta};\mathbf{x}_n,y_n) $ denote the loss function $\ell(\boldsymbol{\theta};\mathbf{x}_n,y_n)$ summed over all the training data $\{ (\mathbf{x}_n,y_n), n=1,2,...,N_m  \} $
at worker/node $m$, where $\mathbf{x}_n$ is the $n$-th feature vector and $y_n$ is the corresponding label. 
A common method to solve (\ref{basicprob}) uses a  worker-server architecture where a central server updates the model parameter by aggregating the gradients computed at all workers and then transmits the updated parameter back to the workers.} The learning task is solved after many communications between workers and the server. {However, the communications latency is generally orders of magnitude larger than local memory access, e.g., about $2500$ times larger over a standard network connection \cite{smith2017cocoa}. }
\textcolor[rgb]{0.00,0.00,0.00}{For this reason, the required communications can be a significant bottleneck on overall performance, especially in federated learning and edge AI systems \cite{chen2021ordered,zhou2021communication, ozfatura2021time, jhunjhunwala2021adaptive, li2021communication, gao2021convergence, chen2022distributed, chen2020ordered}.} The communications can also be a large drain on the batteries of mobile nodes. Therefore, it is extremely important to develop distributed optimization algorithms that are efficient in terms of communications. In this paper we develop a communication-efficient server-worker processing scheme to significantly reduce communications without sacrificing convergence speed.

\subsection{Prior Art}

The challenge of communication-efficient distributed optimization has been addressed in several ways.
Data compression is one popular solution to reduce the bandwidth usage per transmission.
The existing approaches mainly exploit quantization and sparsification.
Quantization aims to approximate a continuous-valued quantity
using a finite set of values to simplify processing and storage, 
and has been successfully applied in many distributed optimization algorithms \cite{jhunjhunwala2021adaptive,sattler2021cfd, honig2022dadaquant, fan2021communication,   wang2021quantized, elkordy2022heterosag, chen2021dynamic}.
Sparsification techniques can also reduce bandwidth.  For example some small gradient vector coordinates can be assumed to be zero while the remaining coordinates can be appropriately amplified to ensure the sparsified gradient is unbiased \cite{ozfatura2021time, panda2022sparsefed,mitra2021linear, stich2018sparsified}. However, the existing efforts for quantization and sparsification reduce the bandwidth usage, not the number of communications. Additionally, quantization and sparsification bring noise and errors which slow down the optimization process in general.

In addition to reducing the number of
bits per communication, accelerating convergence is another straightforward method to enhance communication efficiency. 
One-shot parameter averaging methods were proposed in \cite{zinkevich2010parallelized, zhang2013communication, mcdonald2009efficient, mcmahan2016federated} to find the optimal minimizer using only one iteration.   While the idea of reducing the number of iterations is desirable, one-iteration approaches may sacrifice too much performance to accomplish this goal in some cases  \cite{shamir2014communication}. As an alternative to summing gradients, researchers have considered more sophisticated methods where a worker can compute (or approximately compute) its local Hessian.  Examples of these approaches include quasi-Newton methods \cite{agarwal2014reliable, bordes2009sgd, byrd2016stochastic, moritz2016linearly, gower2016stochastic} and approximated Newton methods \cite{shamir2014communication, zhang2015disco, reddi2016aide}. On the other hand, primal-dual methods \cite{smith2017cocoa, duchi2011dual, scaman2018optimal, he2018cola} are also shown to be efficient in distributed optimization where the primal solution is obtained by solving the dual problem. However, these algorithms save communications at the cost of increasing local computation, unlike the method proposed in this paper which does not increase local computation compared to the first-order algorithms where only the local gradient is computed.

As an alternative to speeding up the optimization process, reducing the number of transmissions per iteration is another efficient way to save communications. \textcolor[rgb]{0.00,0.00,0.00}{An approach called censoring has previously been studied for applications other than learning \cite{rago1996censoring, appadwedula2005energy, marano2006cross, patwari2003hierarchical}. Censoring eliminates transmissions of less informative data, thus reducing worker communications. The original work \cite{rago1996censoring, appadwedula2005energy, marano2006cross, patwari2003hierarchical} focused on hypothesis testing and the censoring was applied to the worker likelihood ratios.  Later, the ideas were applied to distributed optimization, 
where the censoring was applied to workers {such that worker gradients are not transmitted if they do not change significantly  compared to previously transmitted ones}.} Distributed optimization approaches employing censoring include the event-triggered zero-gradient-sum algorithms \cite{chen2016event}, and the censoring-based subgradient algorithm \cite{lu2017event}. However, the step size of the algorithms in \cite{chen2016event, lu2017event} is reduced at each iteration to guarantee convergence, which results in relatively slow progress in the optimization process as the step size becomes smaller and smaller. Distributed dual averaging with censoring was proposed in \cite{tsianos2013networked} but it does not have theoretical justification for convergence or communication saving, unlike this paper. \textcolor[rgb]{0.00,0.00,0.00}{Gradient descent (GD) with censoring was proposed 
in \cite{chen2018lag} 
to save communications for machine learning in 
a comprehensive study that our paper builds upon. In \cite{chen2018lag}, each step is chosen according to the approximated steepest descent direction without considering previous search directions, whereas in our paper we propose a heavy ball (HB) method with a momentum term. The significant practical advantages of the HB method for learning problems are well known  \cite{leen1994optimal, sutskever2013importance}, but the question of reducing communications, including convergence, has not been addressed. The following proposed censoring-based heavy ball (CHB) method 
accelerates convergence, as might be expected from 
previous studies of heavy ball, but also takes advantage of the momentum term to smooth the gradient and reduce changes 
resulting in further communication savings. }

\textbf{Our contributions.} In this paper we develop a censoring-based heavy ball (CHB) method for distributed learning in the server-worker architecture {where our censoring threshold for each worker to decide whether to transmit or not at each iteration is proposed in an intuitive and straightforward way. Combining the censoring threshold with the CHB parameter update rule},  \textcolor[rgb]{0.00,0.00,0.00}{a rigorous convergence guarantee is derived in this paper for CHB under strongly convex, convex and nonconvex cases.  Our results also demonstrate that CHB can achieve the same linear convergence rate as the classical HB method for a smooth and strongly convex objective function.
We show that more than half of the communications can be saved without any impact on the convergence rate under certain conditions.}  \textcolor[rgb]{0.00,0.00,0.00}{Numerical experimental results demonstrate that CHB is able to significantly reduce the total number of communications while maintaining a fast convergence rate to a target solution accuracy when 
compared to other distributed learning algorithms for both synthetic and real datasets.}

\textbf{Notation.} \textcolor[rgb]{0.00,0.00,0.00}{Bold lower case letters are used throughout this paper to denote column vectors. The notation $\|\bf{x}\|$ and $\bf{x^{\top}}$ is employed to denote the  $\ell_2$-norm and the transpose of a column vector $\bf{x}$, respectively.}

\section{Censoring-based heavy ball method}
Before introducing our method for reducing communications, we begin with a brief discussion of the classical HB method \cite{Bottou:2018}, a popular iterative optimization algorithm, and focus on its parameter update rule in a distributed system with a server and $M$ workers. Specifically, at iteration $k$ of HB, the server broadcasts the current parameter $\boldsymbol{\theta}^k$ to all workers; each worker $m$ computes the gradient $\nabla f_m(\boldsymbol{\theta}^k)$ based on its own local function $f_m(\boldsymbol{\theta}^k)$ and transmits $\nabla f_m(\boldsymbol{\theta}^k)$ back to the server. \textcolor[rgb]{0.00,0.00,0.00}{Upon receiving $\nabla f_m(\boldsymbol{\theta}^k)$ from each  worker, the server updates  $\boldsymbol{\theta}^{k}$ as  \cite{Bottou:2018}}
\begin{align}\label{HBupdate}
{\textbf { HB-update }}\quad \boldsymbol{\theta}^{k+1}&=\boldsymbol{\theta}^{k}-\alpha \nabla f(\boldsymbol{\theta}^{k})+\beta(\boldsymbol{\theta}^{k}-\boldsymbol{\theta}^{k-1}) \text { with } \notag\\
&\qquad\nabla f(\boldsymbol{\theta}^{k})\buildrel \Delta \over=\sum_{m \in \mathcal{M}} \nabla f_{m}(\boldsymbol{\theta}^{k})
\end{align}
where $\alpha$ is the step size, $\beta$ is a constant, and $\nabla f(\boldsymbol{\theta}^{k})$ is the gradient of $f(\boldsymbol{\theta}^{k})$. To implement (\ref{HBupdate}) requires $M$ communications from the workers to the server during each iteration $k$.

Our communication-efficient CHB algorithm will be described next. 
{Each worker {maintains} 
two vectors concerning iteration $k$. One is the parameter vector $\boldsymbol{\theta}^k$ sent by the server at the start of iteration $k$. The other is the last gradient $\nabla f_{m}(\hat{\boldsymbol{\theta}}_{m}^{k-1})$
transmitted from worker $m$ to the server
{prior} to iteration {$k$}.} An important feature of CHB is that a worker $m$ will not transmit to the server if the current gradient $\nabla f_{m}(\boldsymbol{\theta}^{k})$ is not sufficiently different from the previously transmitted gradient $\nabla f_{m}(\hat{\boldsymbol{\theta}}_{m}^{k-1})$. Define
\begin{align}\label{definediff}
\delta \nabla_{m}^{k}\buildrel \Delta \over=\nabla f_{m}(\boldsymbol{\theta}^{k})-\nabla f_{m}(\hat{\boldsymbol{\theta}}_{m}^{k-1}).
\end{align}
{It is worth mentioning that $\delta \nabla_{m}^{k}$ characterizes the ``newness'' of the information contained in the gradient of worker $m$ at iteration $k$.} To implement the CHB algorithm, at the beginning of iteration $k$ the server broadcasts $\boldsymbol{\theta}^{k}$ to all workers. Then only if worker $m$ has a gradient which is sufficiently different from that previously transmitted will it transmit $\delta \nabla_{m}^{k}$ back to the server. Immediately after transmitting, worker $m$ will update its transmitted gradient  $\nabla f_{m}(\hat{\boldsymbol{\theta}}_{m}^{k})=\nabla f_{m}({\boldsymbol{\theta}}^{k})$ while others keep their previous values $\nabla f_{m}(\hat{\boldsymbol{\theta}}_{m}^{k})=\nabla f_{m}(\hat{\boldsymbol{\theta}}_{m}^{k-1})$. At the end of iteration $k$, the server updates the parameter $\boldsymbol{\theta}^{k}$ via
\begin{align}\label{commuinicationHBupdate}
{\textbf { CHB-update }}\quad
\boldsymbol{\theta}^{k+1}&=\boldsymbol{\theta}^{k}-\alpha\nabla^k +  \beta(\boldsymbol{\theta}^{k}-\boldsymbol{\theta}^{k-1})\text { with }\notag\\
&\nabla^{k}\buildrel \Delta \over=\sum_{m \in \mathcal{M}} \nabla f_{m}(\hat{\boldsymbol{\theta}}^{k})
\end{align}
where $\nabla^{k}$ is an approximation to the gradient at iteration $k$ which is computed recursively using
\begin{align}\label{approximatedgradi}
\nabla^{k}=\nabla^{k-1}+\sum_{m\in\mathcal{M}^{k}}\delta \nabla_{m}^{k}.
\end{align}
Here $\mathcal{M}^{k}$ is a set which collects the indices of workers who have transmitted during iteration $k$ and whose cardinality is denoted $|\mathcal{M}^{k}|$. The number of communications are often reduced to $|\mathcal{M}^{k}|<M$ during iteration $k$ in the CHB algorithm which explains the term $\sum_{m\in\mathcal{M}^{k}}\delta \nabla_{m}^{k}$ in (\ref{approximatedgradi}). In fact, if we replace $\mathcal{M}^{k}$ in (\ref{approximatedgradi}) with $\mathcal{M}$, plugging (\ref{approximatedgradi}) into (\ref{commuinicationHBupdate}) yields the classical HB update rule in (\ref{HBupdate}).

Next we describe when a worker will skip a transmission, which we call the CHB-skip-transmission condition. To make the principles behind the design of the CHB-skip-transmission condition clear, the CHB update rule in (\ref{commuinicationHBupdate}) can be rewritten as
\begin{small}
\begin{align}\label{}
\boldsymbol{\theta}^{k+1}
&= \boldsymbol{\theta}^{k}-\alpha\bigg(\nabla f(\boldsymbol{\theta}^{k}) - \sum_{m\in\mathcal{M}_c^{k}}\delta \nabla_{m}^{k} \bigg) + \beta(\boldsymbol{\theta}^{k}-\boldsymbol{\theta}^{k-1})\label{CHBupdatestep0}\\
&= \boldsymbol{\theta}^{k}- \alpha\nabla f(\boldsymbol{\theta}^{k}) + \beta(\boldsymbol{\theta}^{k}-\boldsymbol{\theta}^{k-1}) + \alpha\sum_{m\in\mathcal{M}_c^{k}}\delta \nabla_{m}^{k}\label{CHBupdatestep1}
\end{align}
\end{small}where $\mathcal{M}_c^{k}$ is a set which collects the
indices of workers who do not transmit during iteration $k$, and $\delta \nabla_{m}^{k}$ is defined in (\ref{definediff}). It is well known \cite{Bottou:2018} that the update steps in HB tend to accumulate contributions in directions of persistent descent while canceling directions that oscillate. This interpretation provides some intuitive explanation as to why the HB method often outperforms gradient descent (GD) \cite{Bottou:2018}.
Since the update rule of CHB in (\ref{commuinicationHBupdate}) is very similar to that of the traditional HB method (just drop less
informative terms), it is not surprising that we have observed that
the CHB algorithm inherits the advantages from the HB method.
 Compared with the classical HB update rule in (\ref{HBupdate}), the result in (\ref{CHBupdatestep1}) {shows that we can save transmissions by not allowing the workers in $\mathcal{M}_c^{k}$ to transmit if $\delta \nabla_{m}^{k}$ for all $m\in\mathcal{M}_c^{k}$ is very small compared to $(\boldsymbol{\theta}^{k}-\boldsymbol{\theta}^{k-1})$. This takes advantage of the HB smoothing and as will see, we can save transmissions with very little affect on the optimization convergence rate. 
 Therefore, the rule to determine if worker $m$ will or will not} transmit is called the {\textbf{CHB-skip-transmission condition}}, 
\begin{align}\label{CHB-stopping condition}
 \|\delta \nabla_{m}^{k}\|^2 \le \varepsilon_1\|\boldsymbol{\theta}^{k}-\boldsymbol{\theta}^{k-1}\|^2
\end{align}
where $\varepsilon_1$ is a positive constant that {allows us to tune the convergence while saving communications.}
{Intuitively, smaller $\varepsilon_1$ leads to less communication censoring, e.g., CHB reduces to the classical HB method when we set $\varepsilon_1=0$. On the other hand, when $\varepsilon_1$ increases, the number of communications per iteration is  reduced at the cost of increasing the number of iterations. A favorable communication-iteration trade-off can be achieved by CHB via tuning $\varepsilon_1$;
see the example 
in Figure \ref{fig:diffThres}. 
Finding a theoretically optimal value of
$\varepsilon_1$ is an interesting open problem. 
Note that using similar steps to those  shown in \cite{chen2018lag}, the condition (8) can also be derived based on the criterion which ensures that CHB has larger per-communication descent than HB.}
At iteration $k$, after receiving $\boldsymbol{\theta}^{k}$ from the server, each worker can independently check the CHB-skip-transmission condition (\ref{CHB-stopping condition}). If the CHB-skip-transmission condition is not satisfied, then worker $m$ will immediately transmit $\delta \nabla_{m}^{k}$ to the server. The CHB
algorithm is summarized as Algorithm \ref{CHB}.

\begin{savenotes}
\begin{algorithm}
\begin{algorithmic}[1]
\caption{{CHB}}\label{CHB}
\Require{step size $\alpha$, positive constants $\varepsilon_1$ and $\beta$.}
\Ensure{${\boldsymbol{\theta}}^1$, \{$\nabla\mathcal{L}_{m}(\hat{\boldsymbol{\theta}}_{m}^{0})$, $\forall m$\}}
\For{$k=1, 2,..., K$}
\State Server broadcasts $\boldsymbol{\theta}^k$ at the starting time of iteration $k$.
\For{$m=1, 2,..., M$}
\If{the CHB-skip-transmission condition in (\ref{CHB-stopping condition}) is not satisfied for worker $m$}
\State Worker $m$ transmits $\delta \nabla_{m}^{k}$ to the server and locally updates $\nabla f_{m}(\hat{\boldsymbol{\theta}}_{m}^{k})=\nabla f_{m}({\boldsymbol{\theta}}^{k})$.
\Else
\State Worker $m$ does not transmit but locally updates $\nabla f_{m}(\hat{\boldsymbol{\theta}}_{m}^{k})=\nabla f_{m}(\hat{\boldsymbol{\theta}}_{m}^{k-1})$.
\EndIf
\EndFor
\State Server updates ${\boldsymbol{\theta}}^{k}$ via (\ref{commuinicationHBupdate}).
\EndFor
\end{algorithmic}
\end{algorithm}
\end{savenotes}

\section{Convergence and communication analysis}

In this section, convergence rate and communication reduction guarantees for CHB are developed for the proper choice of the constants $\alpha$, $\beta$, and $\varepsilon_1$.
The following assumptions and {definitions} are needed for the provided results. 

\begin{assumption}\label{OHBassumption1}
In \emph{(\ref{basicprob})}, $f(\boldsymbol{\theta})$ in is $L$-smooth and coercive. This implies there exists a constant $L>0$ such that $\|\nabla f(\boldsymbol{\theta}_1)-\nabla f(\boldsymbol{\theta}_2)\|\le L\|\boldsymbol{\theta}_1-\boldsymbol{\theta}_2\|,\ \forall \ \boldsymbol{\theta}_1, \boldsymbol{\theta}_2$ \emph{\cite{nesterov2018lectures}} and $\lim_{\|\boldsymbol{\theta}\|\rightarrow\infty}f(\boldsymbol{\theta})=+\infty$ \emph{\cite{peressini1988mathematics}}.
\end{assumption}
{\begin{definition}\label{OHBassumption2}
\textcolor[rgb]{0.00,0.00,0.00}{The objective function  $f(\boldsymbol{\theta})$ in \emph{(\ref{basicprob})} is $\mu$-strongly convex, which implies the existence of a constant $\mu>0$ such that
$f(\boldsymbol{\theta}_1)\ge f(\boldsymbol{\theta}_2) + \nabla f(\boldsymbol{\theta}_2) ^{\top}(\boldsymbol{\theta}_1-\boldsymbol{\theta}_2)  + \frac{\mu}{2}\|\boldsymbol{\theta}_1-\boldsymbol{\theta}_2\|^2, \forall \ \boldsymbol{\theta}_1, \boldsymbol{\theta}_2$ \emph{\cite{nesterov2018lectures}}.}
\end{definition}}
{\begin{definition}\label{OHBassumption3}
\textcolor[rgb]{0.00,0.00,0.00}{The objective function  $f(\boldsymbol{\theta})$ in \emph{(\ref{basicprob})} is convex, which  implies $f(\boldsymbol{\theta})$ satisfies $f(\lambda\boldsymbol{\theta}_1 + (1-\lambda)\boldsymbol{\theta}_2)\le \lambda f(\boldsymbol{\theta}_1) + (1-\lambda)f(\boldsymbol{\theta}_2), \forall \ \boldsymbol{\theta}_1, \boldsymbol{\theta}_2$ and $0\le\lambda\le 1$ \emph{\cite{peressini1988mathematics}}.}
\end{definition}}
\begin{assumption}\label{OHBassumption4}
In \emph{(\ref{basicprob})}, $f_m(\boldsymbol{\theta})$ is $L_m$-smooth for each worker $m$. This implies there exists a constant $L_m>0$ such that
$\|\nabla f_m(\boldsymbol{\theta}_1)-\nabla f_m(\boldsymbol{\theta}_2)\|\le L_m\|\boldsymbol{\theta}_1-\boldsymbol{\theta}_2\|,\ \forall \ \boldsymbol{\theta}_1, \boldsymbol{\theta}_2$ \emph{\cite{nesterov2018lectures}}.
\end{assumption}
\textcolor[rgb]{0.00,0.00,0.00}{Let 
\begin{align}\label{definationLyapunov}
\mathbb{L}(\boldsymbol{\theta}^{k})\buildrel \Delta \over=f(\boldsymbol{\theta}^{k})-f\left(\boldsymbol{\theta}^{*}\right)+ \eta_{1}\|\boldsymbol{\theta}^{k}-\boldsymbol{\theta}^{k-1}\|^{2}
\end{align}
define a Lyapunov function, 
where $\boldsymbol{\theta}^{*}$ is the optimal solution to the optimization problem in (\ref{basicprob}), and $\eta_{1}$ is a non-negative constant.} If we allow all workers to transmit ($\mathcal{M}^{k}=\mathcal{M}$) and set $\eta_{1}=0$, then CHB reduces to the classical HB method and $\mathbb{L}(\boldsymbol{\theta}^{k})$ will describe the optimization process of the HB method. However, when some communications are suppressed, the following lemma describes the behavior of the Lyapunov function defined in (\ref{definationLyapunov}).
\begin{lemma}\label{descentLyapunovLemma}
Under \emph{Assumption \ref{OHBassumption1}}, if the constants $\alpha$, $\beta$, and $\varepsilon_1$ are chosen so that
\begin{align}
\sigma_0&\ge 0 \ \mbox{with}\ \sigma_0\buildrel \Delta \over= \frac{\alpha}{2} - \Big(\eta_1-\frac{1-\alpha L}{2\alpha}\Big)\alpha^2(1+\rho_1)(1+\rho_2),\label{DDScondition1}\\
\sigma_1&\ge 0 \ \mbox{with}\ \sigma_1\buildrel \Delta \over= -\gamma|\mathcal{M}_c^{k}|^2\varepsilon_1 - \frac{\beta^2}{2\alpha}(1+\rho_3^{-1})\notag\\
&\qquad\qquad - \Big(\eta_1-\frac{1-\alpha L}{2\alpha}\Big)\beta^2(1+\rho_1^{-1})+\eta_1,\label{condition2}\\
\gamma&\ge \frac{\alpha}{2}(1+\rho_3)\ \mbox{with}\ \gamma\buildrel \Delta \over= \frac{\alpha}{2}(1+\rho_3) \notag\\
&\qquad\qquad+\Big(\eta_1-\frac{1-\alpha L}{2\alpha}\Big)\alpha^2(1+\rho_1)(1+\rho_2^{-1}), \label{etacondition5}
\end{align}
where $\eta_1$ is the non-negative constant in (\ref{definationLyapunov}), $|\mathcal{M}_c^{k}|$ is the cardinality of $\mathcal{M}_c^{k}$, and $\rho_1$, $\rho_2$, $\rho_3>0$,
then the Lyapunov function follows
\begin{align}\label{descentLyapunovequation}
\mathbb{L}(\boldsymbol{\theta}^{k+1}) -  \mathbb{L}(\boldsymbol{\theta}^{k}) \le  -\sigma_0\left\|\nabla f\left(\boldsymbol{\theta}^{k}\right)\right\|^{2} - \sigma_1\left\| \boldsymbol{\theta}^{k}-\boldsymbol{\theta}^{k-1} \right\|^{2}
\end{align}
where constants $\sigma_0\ge0$ and $\sigma_1\ge0$ depend on $\alpha$, $\beta$ and $\varepsilon_1$. The result in (\ref{descentLyapunovequation}) implies $\mathbb{L}(\boldsymbol{\theta}^{k+1}) \le \mathbb{L}(\boldsymbol{\theta}^{k})$.
\end{lemma}
\begin{IEEEproof}
Please refer to Appendix \ref{descentProof}.
\end{IEEEproof}

{From (\ref{condition2}), we know that $\sigma_1=\sigma_1(|\mathcal{M}_c^{k}|)$ but for simplicity, we omit $|\mathcal{M}_c^{k}|$ here.} Note that if we set $\big(\eta_1-\frac{1-\alpha L}{2\alpha}\big)=0$, then (\ref{DDScondition1})--(\ref{etacondition5}) are equivalent to
\begin{align}\label{addconstraintLayp}
\alpha\le\frac{1}{L}, \quad \beta&\le\sqrt{\frac{1-\alpha L}{1+\rho_3^{-1}}},\quad \mbox{and}\notag\\ &\qquad\varepsilon_1\le\frac{(1-\alpha L)-\beta^2(1+\rho_3^{-1})}{\alpha^2(1+\rho_3)|\mathcal{M}_c^{k}|^2}.
\end{align}
\textcolor[rgb]{0.00,0.00,0.00}{It can be shown that (\ref{addconstraintLayp}) admits an uncountable number of solutions for $\alpha$, $\beta$ and $\varepsilon_1$ which would guarantee (\ref{DDScondition1})--(\ref{etacondition5}). Some examples of other parameter choices can be found at the end of Appendix \ref{descentProof}.
Next, we will show that the Lyapunov function in (\ref{definationLyapunov}) has a linear convergence rate under certain conditions.}
\begin{theorem}(\textbf{strongly convex})\label{convergenceTheorem}
Under \emph{Assumption \ref{OHBassumption1}} and  {\emph{Definition \ref{OHBassumption1}}}, if constants $\alpha$, $\beta$ and $\varepsilon_1$ are properly selected such that (\ref{DDScondition1})--(\ref{etacondition5}) are satisfied with $\sigma_0>0$ in (\ref{DDScondition1}) and $\sigma_1>0$ in (\ref{condition2}), then the Lyapunov function in (\ref{definationLyapunov}) converges Q-linearly; that is, there exists a constant $c(\alpha,\beta,\varepsilon_1)\in(0,1)$ such that at iteration $k$,
\begin{align}\label{linearconveLay1243}
 \mathbb{L}(\boldsymbol{\theta}^{k+1})\le \big(1-c(\alpha,\beta,\varepsilon_1)\big)\mathbb{L}(\boldsymbol{\theta}^{k})
\end{align}
where $c(\alpha,\beta,\varepsilon_1)=\min\big\{ 2\sigma_0\mu,\min_k\{\sigma_1\}/\eta_1 \big\}$ with $\sigma_0$ and $\sigma_1$ defined in (\ref{DDScondition1}) and (\ref{condition2}), respectively.
The result in (\ref{linearconveLay1243}) implies
\begin{align}\label{linerratefunctiondecrease}
f(\boldsymbol{\theta}^k) - f\left(\boldsymbol{\theta}^*\right)\le \big(1-c(\alpha,\beta,\varepsilon_1)\big)^{k}\mathbb{L}(\boldsymbol{\theta}^0).
\end{align}
\end{theorem}
\begin{IEEEproof}
Please refer to Appendix \ref{theorem1Proof}.
\end{IEEEproof}
\textcolor[rgb]{0.00,0.00,0.00}{Theorem \ref{convergenceTheorem} implies CHB can achieve a linear convergence rate which is of the same order as that for HB \cite{Ghadimi:2015} under {Assumption \ref{OHBassumption1} and {Definition \ref{OHBassumption2}}.}  If we set 
$\rho_3=1$, $\delta\in(0,1)$,
$\alpha=\frac{1-\delta}{L}, \eta_1=\frac{1-\alpha L}{2\alpha}, \varepsilon_1=\frac{(1-\alpha L)(1-\alpha \mu)}{4\alpha^2M^2}, \mbox{and}\  \beta=\frac{1}{2}\sqrt{(1-\alpha L)(1-\alpha\mu)},$
then
\begin{align}\label{CHBcomxplx}
c(\alpha,\beta,\varepsilon_1) = \frac{1-\delta}{L/\mu},
\end{align}
which is exactly the same value we obtain in HB if we also choose $\alpha={(1-\delta)}/{L}$  \cite{Ghadimi:2015}. There are an uncountable number of other parameter settings in CHB that provide the same order convergence rate as HB.} However, the just predicted theoretical convergence rates of CHB and HB are very conservative compared to what is observed in numerical examples. Numerical results in Section \ref{numericalresult} indicate that CHB with $\alpha=1/L$ and $\beta>0$ typically requires almost the same number of iterations as HB to achieve the same value of the left-hand-side of (\ref{linerratefunctiondecrease}) and often outperforms GD. \textcolor[rgb]{0.00,0.00,0.00}{Convergence guarantees for CHB under general convex and nonconvex objective function cases are given next. The proofs are provided in Appendix \ref{convexCase} and Appendix \ref{nonConvexCase} respectively.} 
\begin{theorem}(\textbf{convex})\label{convergenceTheorem2}
Under \emph{Assumption \ref{OHBassumption1}} and {\emph{Definition \ref{OHBassumption3}}}, if the constants $\alpha$, $\beta$, and $\varepsilon_1$ are chosen so that (\ref{DDScondition1})--(\ref{etacondition5}) are satisfied with $\sigma_0>0$ in (\ref{DDScondition1}) and $\sigma_1>0$ in (\ref{condition2}), then
\begin{align}
f(\boldsymbol{\theta}^{k})-f\left(\boldsymbol{\theta}^{*}\right)=\mathcal{O}(1 / k).
\end{align}
\end{theorem}

For nonconvex objective functions, the following convergence guarantee can be established.
\begin{theorem}(\textbf{nonconvex})\label{convergenceTheorem3}
Under \emph{Assumption \ref{OHBassumption1}}, if the constants $\alpha$, $\beta$, and $\varepsilon_1$ are chosen so that (\ref{DDScondition1})--(\ref{etacondition5}) are satisfied with $\sigma_0>0$ in (\ref{DDScondition1}) and $\sigma_1>0$ in (\ref{condition2}), then
\begin{align}
\lim_{k\rightarrow\infty}\left\|\nabla f(\boldsymbol{\theta}^{k})\right\|^{2}\rightarrow 0.
\end{align}
\end{theorem}

\begin{figure}
\centering
\begin{subfigure}[b]{.25\textwidth}
  \centering
  \includegraphics[width=\linewidth]{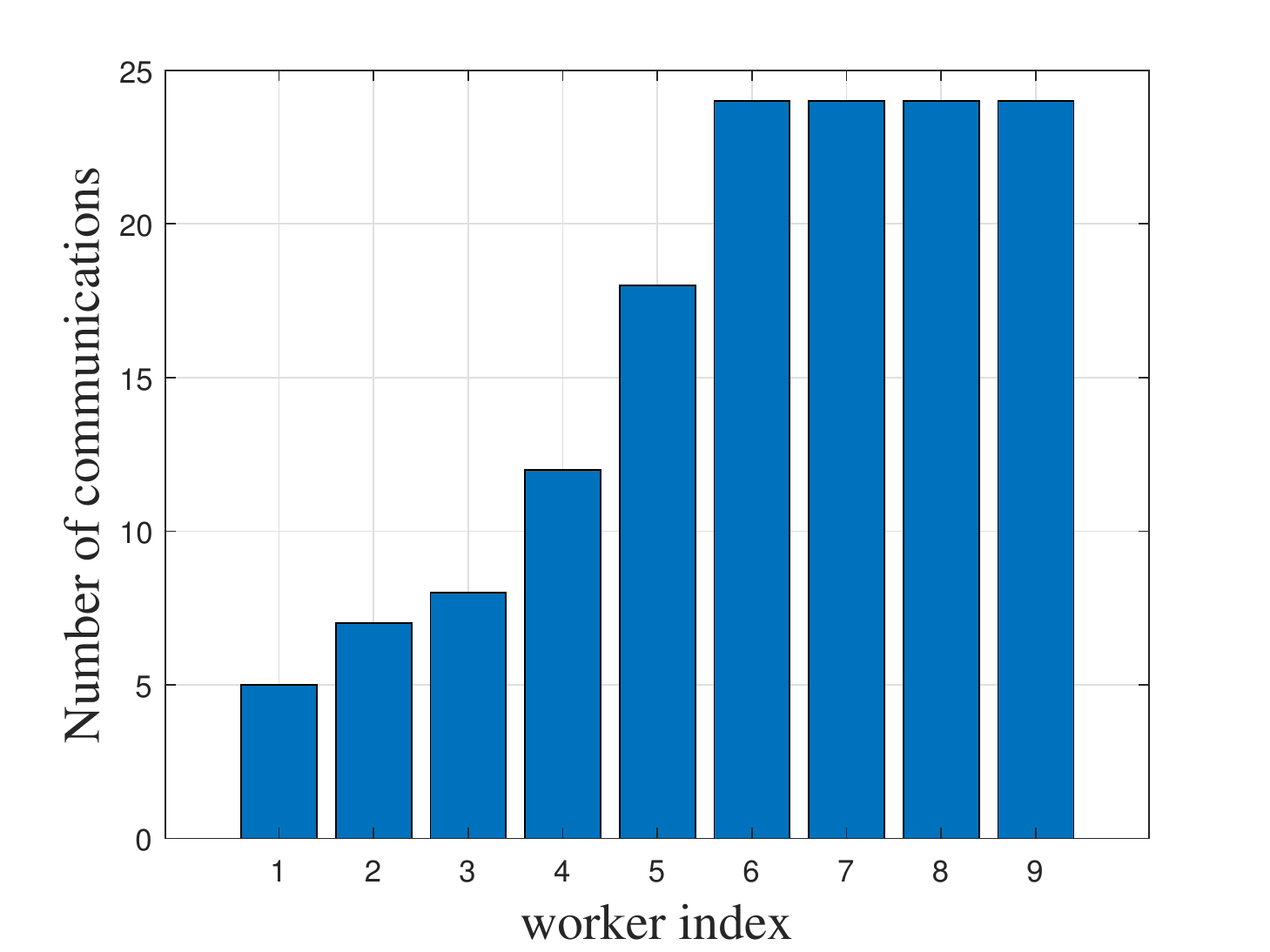}
  \caption{Comm. of $9$ workers in CHB.}
  \label{fig:subdia}
\end{subfigure}%
\begin{subfigure}[b]{.25\textwidth}
  \centering
  \includegraphics[width=\linewidth]{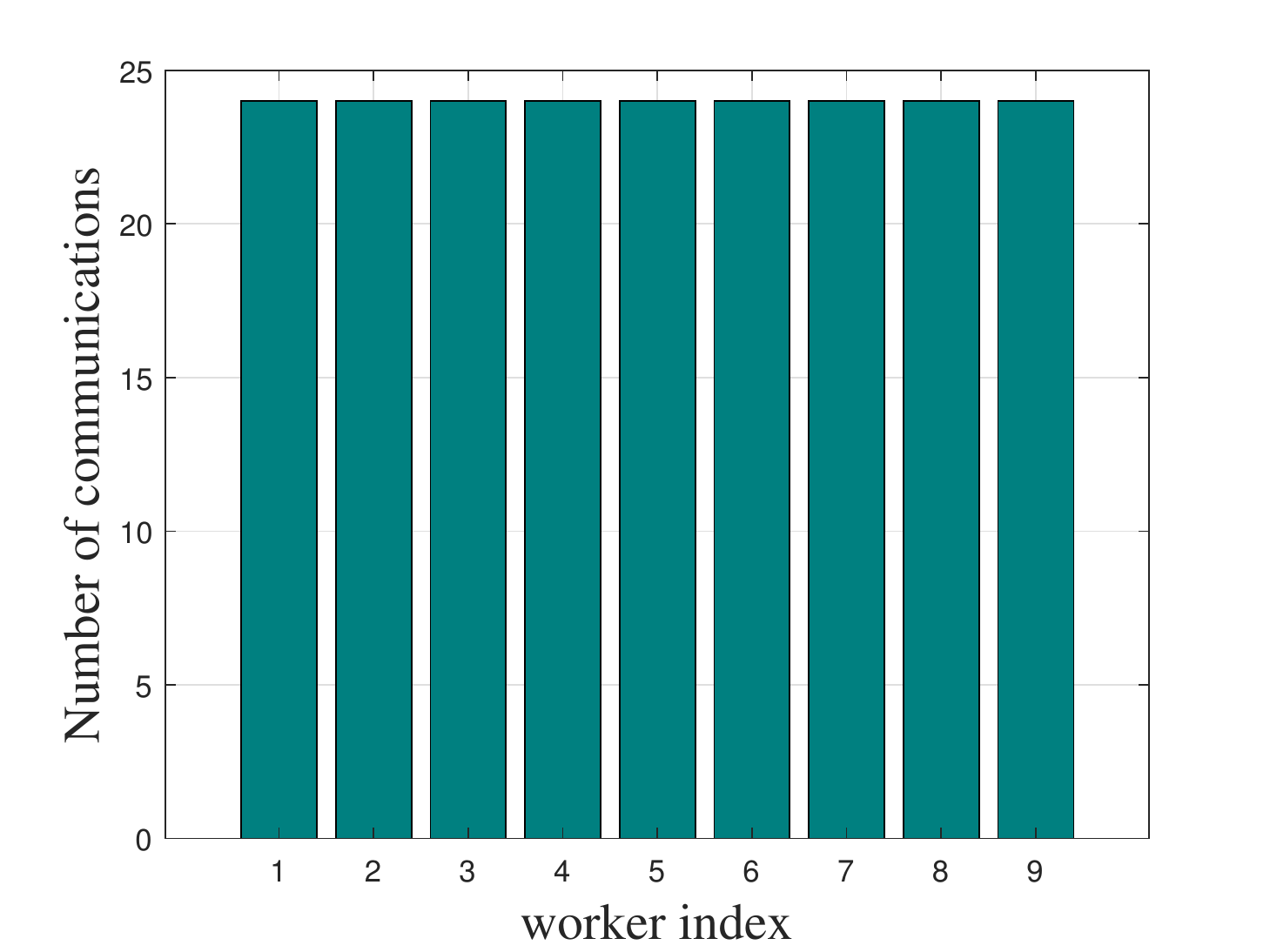}
  \caption{Comm. of $9$ workers in HB.}
  \label{fig:subdia}
\end{subfigure}
\caption{Number of communications of different workers for the first $24$ iterations of CHB and HB in linear regression {with $\alpha=1/L$,  $\varepsilon_1=0.1/(\alpha^2M^2)$ and smoothness constants satisfying $L_1<L_2<...<L_9$ with $L_m=(1.3^{m-1})^2$}. }
\label{fig:realLRdiagram}
\end{figure}

The following lemma illustrates a communication saving bound for the CHB algorithm.
As indicated in Figure \ref{fig:realLRdiagram}, worker $m$ with a small smoothness constant $L_m$ tends to transmit less frequently. The following lemma formally describes the impact of $L_m$ on the communication savings of worker $m$.

\begin{lemma}\label{ONBsavinglemma}
Under \emph{Assumption \ref{OHBassumption4}}, if the smoothness constant $L_m$ of worker $m$ satisfies
\begin{align}\label{Lmstracincondition1}
L_m^2\le\varepsilon_1,
\end{align}
then after iteration $k$ in CHB, the number of 
communications between worker $m$ and the server, denoted as $S_m$, can be upper bounded by
\begin{align}\label{}
S_m\le \frac{k}{2}.
\end{align}
\end{lemma}
\begin{IEEEproof}
Please refer to Appendix \ref{CommSavingLemma}.
\end{IEEEproof}
The result in Lemma \ref{ONBsavinglemma} indicates the smoothness of the local function decides the number of  communications needed by worker $m$. It makes sense that worker $m$ with a small smoothness constant $L_m$ tends to transmit less frequently 
because \textcolor[rgb]{0.00,0.00,0.00}{ $L_m$ indicates the maximum rate of change of worker $m$'s gradient.} It is worth mentioning that if (\ref{Lmstracincondition1}) is true for all workers and the conditions in (\ref{DDScondition1})--(\ref{etacondition5}) are satisfied with $\sigma_0>0$ in (\ref{DDScondition1}) and $\sigma_1>0$ in (\ref{condition2}), then we can  save at least half of the transmissions but we still
achieve the same convergence rate as HB, which is known to be good.

\section{Numerical experiments}\label{numericalresult}
To validate the theoretical results on convergence analysis and communication savings,  the empirical performance of CHB is evaluated for \textcolor[rgb]{0.00,0.00,0.00}{(1) linear regression (convex), (2) regularized logistic regression (strongly convex), (3) lasso regression (nondifferentiable) and (4) training a neural network (nonconvex). The objective error $f(\boldsymbol{\theta}^k)-f(\boldsymbol{\theta}^*)$ evaluates the algorithm progress} for linear regression, regularized logistic regression, and lasso regression. For training a neural network {with one hidden layer with $30$ nodes and the sigmoid activation function}, we use the norm of the gradient $\|\nabla^k\|$ as a figure of merit of the progress. We consider a scenario with one server and nine workers. To benchmark CHB, we compare its performance with GD \cite{nesterov2018lectures}, censoring-based GD (called LAG-WK) \cite{chen2018lag}, and the classical HB method \cite{Bottou:2018} using synthetic datasets and real datasets. Except as stated elsewhere, we set the constant $\beta=0.4$ for HB and CHB and choose the same skip-transmission condition (\ref{CHB-stopping condition}) for CHB and censoring-based GD. Note that hereafter in the paper we refer to regularized logistic regression as logistic regression. {The specific loss function formats of the above machine learning tasks can be found in \cite{geron2019hands}}. {For clarity of presentation, we divide the experiments into two sets.}

\subsection{{Experiment Set 1}}

\begin{figure}[!htb]
\centering
\begin{subfigure}[b]{.25\textwidth}
  \centering
  \includegraphics[width=\linewidth]{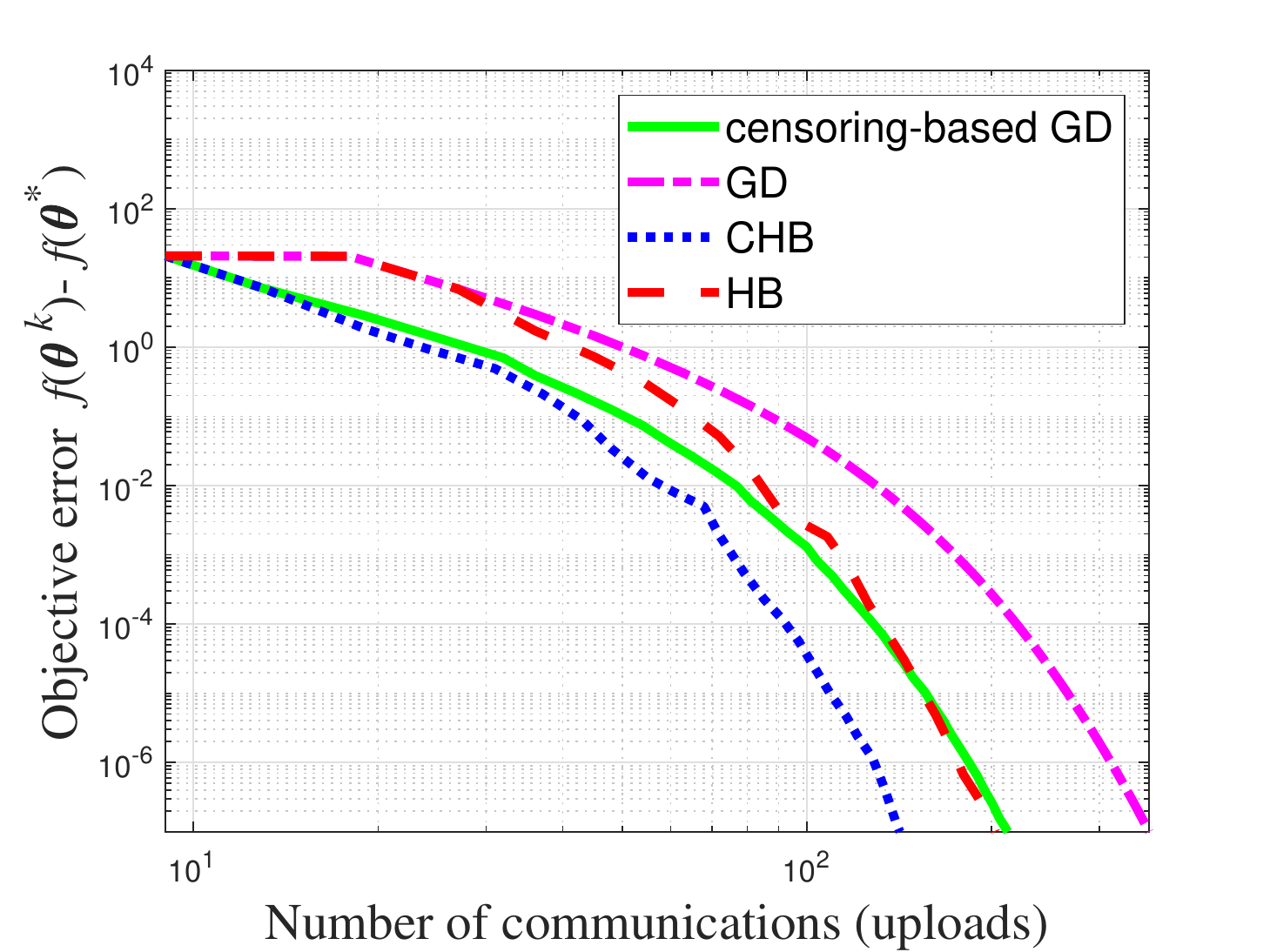}
\end{subfigure}%
\begin{subfigure}[b]{.25\textwidth}
  \centering
  \includegraphics[width=\linewidth]{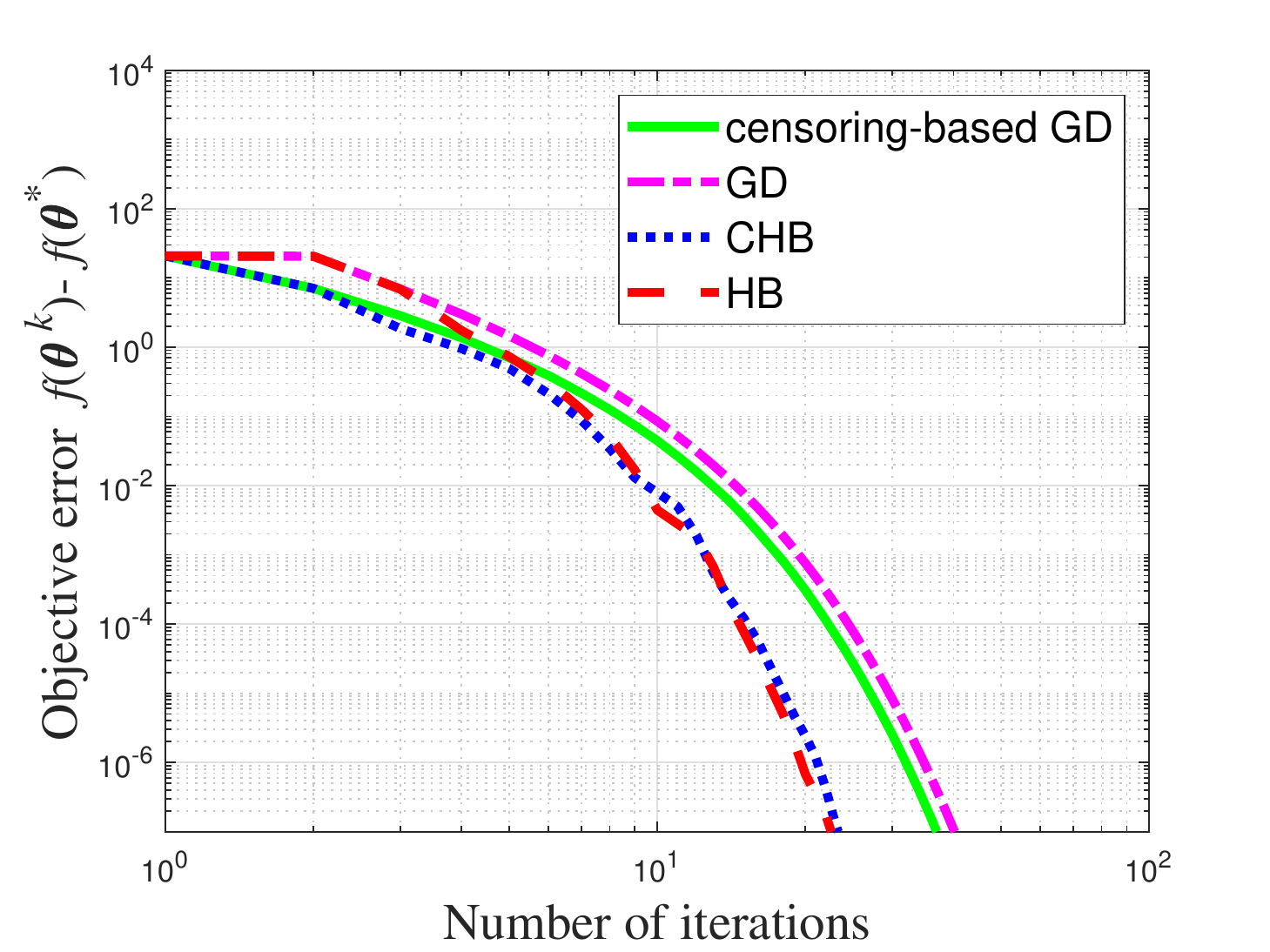}
\end{subfigure}
\caption{{Objective error versus the number of communications and iterations for linear regression with increasing smoothness constant $L_m=(1.3^{m-1})^2$ for $m=1,2,...,9$ in synthetic datasets.}}\label{fig:imageLCallLinear}
\end{figure}

\begin{figure}[!htb]
\centering
\begin{subfigure}[b]{.25\textwidth}
  \centering
  \includegraphics[width=\linewidth]{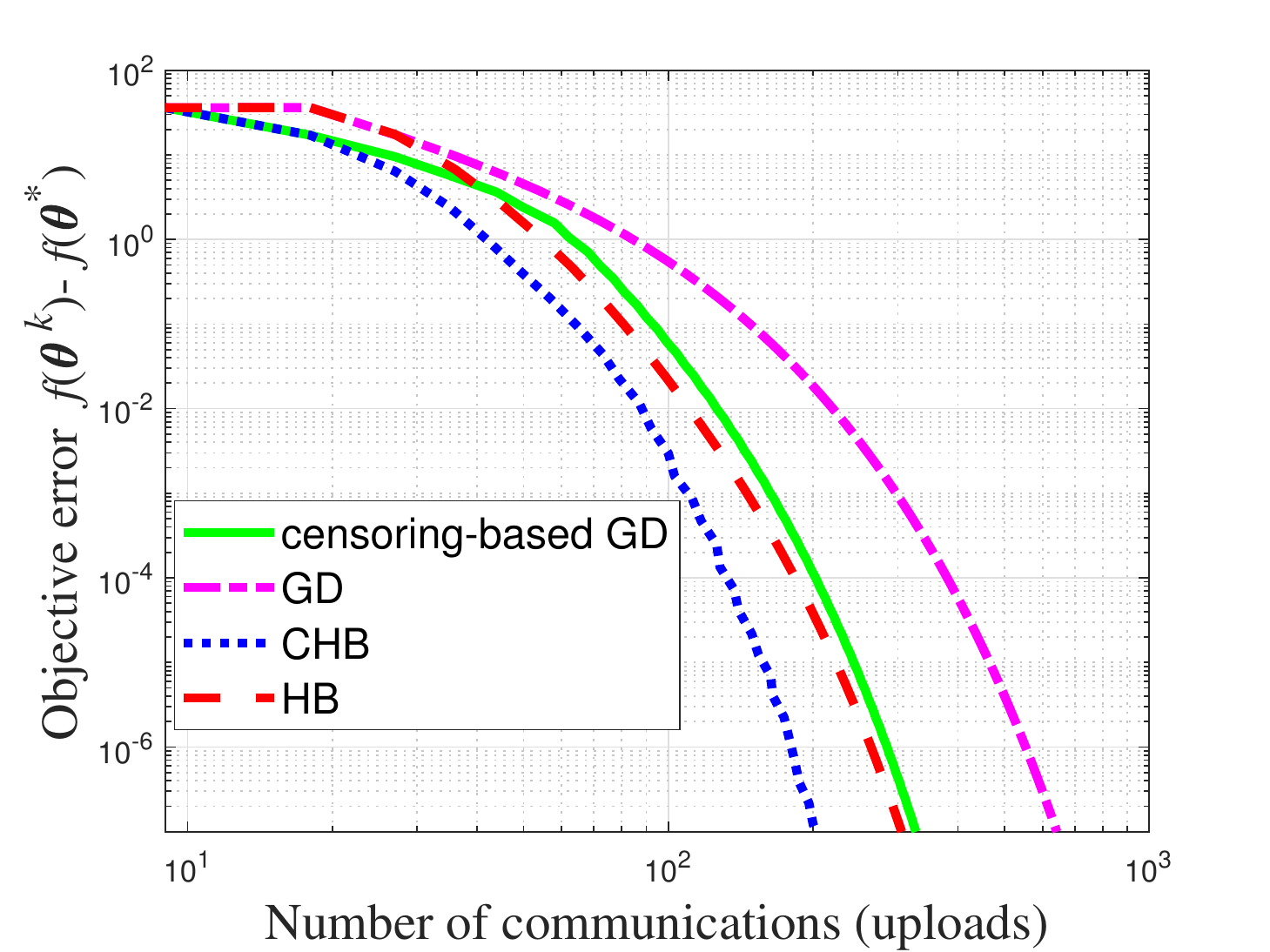}
\end{subfigure}%
\begin{subfigure}[b]{.25\textwidth}
  \centering
  \includegraphics[width=\linewidth]{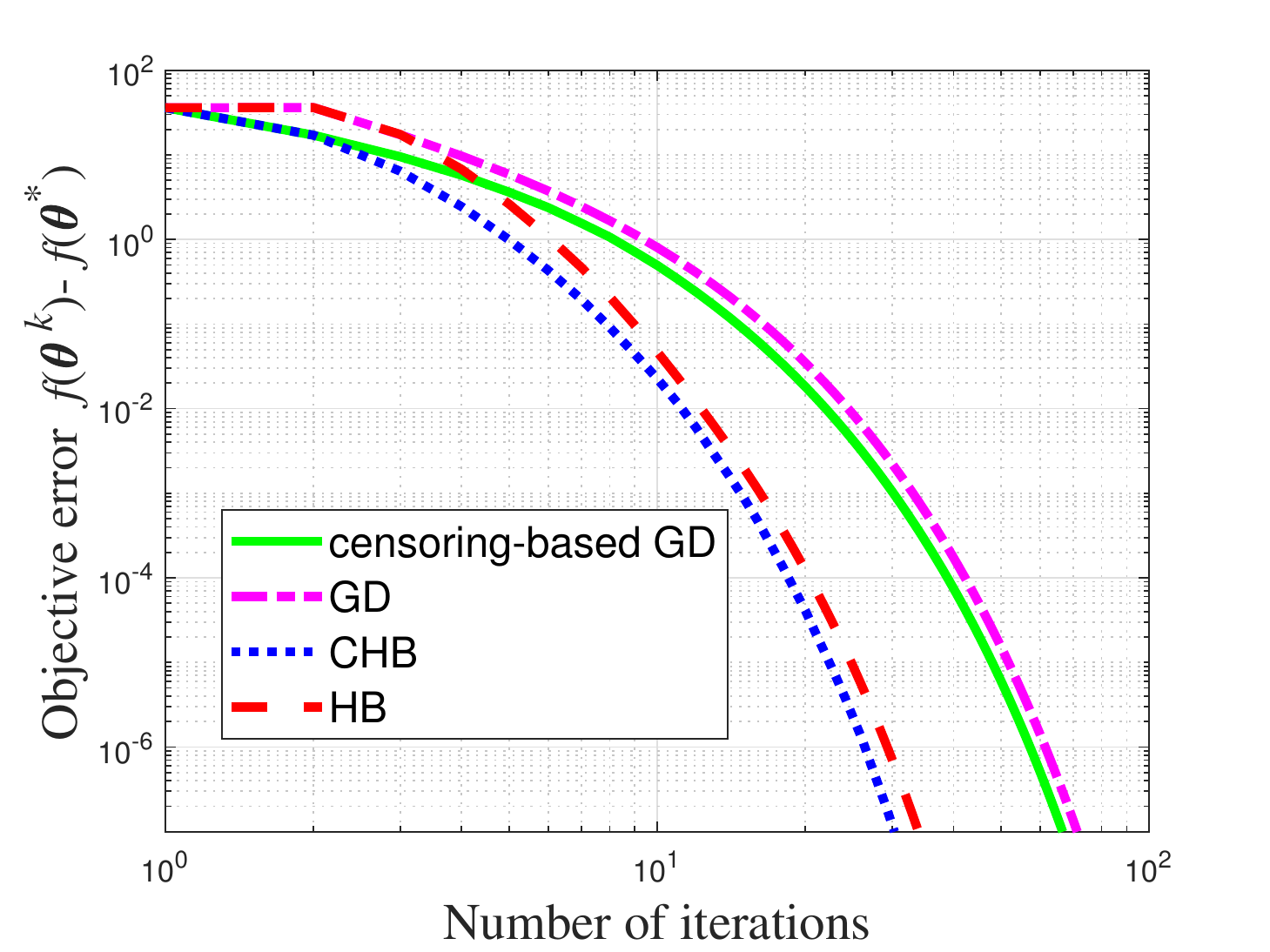}
\end{subfigure}
\caption{{Objective error versus the number of communications and iterations for logistic regression with common smoothness constants
$L_1 = L_2 = ... = L_9 = 4$ in synthetic datasets.}}\label{fig:imageLCall}
\end{figure}

\textbf{Synthetic dataset tests.} \textcolor[rgb]{0.00,0.00,0.00}{Let us initially consider the impact of the smoothness constant on communication savings in synthetic datasets for linear regression} and logistic regression. Specifically, we consider linear regression with increasing smoothness constants $L_m=(1.3^{m-1})^2$ for $m=1,2,...,9$ \textcolor[rgb]{0.00,0.00,0.00}{in Figure \ref{fig:imageLCallLinear} (same setting as Figure \ref{fig:realLRdiagram})} and logistic regression with common smoothness constants $L_1=L_2=...=L_9=4$ \textcolor[rgb]{0.00,0.00,0.00}{in Figure \ref{fig:imageLCall}}. \textcolor[rgb]{0.00,0.00,0.00}{For these two cases, we randomly generate an independent sequence of labels (over $n$), each with equal probability of $y_n=1$ or $y_n=-1$ for each worker $m$. Then we randomly generate 50 independent instances  $\mathbf{x}_{n}\in\mathbb{R}^{50}$ (to pair with the labels) from} a standard normal distribution and use the same approach as \cite{chen2018lag} to rescale the data to change the value of smoothness constants. To enable useful comparison, we set $\alpha=1/L$ for all algorithms. For censoring-based GD and CHB, we choose $\varepsilon_1=0.1/(\alpha^2M^2)$ in (\ref{CHB-stopping condition}). {In Figure \ref{fig:imageLCallLinear}, it is not surprising that CHB requires a fewer number of communications compared to HB since we know from Figure \ref{fig:realLRdiagram} that workers with smaller smoothness constants tend to transmit less frequently such that communications can be saved for these workers. Interestingly, Figure \ref{fig:imageLCall} indicates that even when workers have the same smoothness constants, CHB can still save  communications. Both Figure \ref{fig:imageLCallLinear} and Figure \ref{fig:imageLCall}} indicate CHB outperforms the alternatives in terms of the number of communications saved while still employing nearly the same number of iterations as HB to achieve the same objective error. In these cases, 
CHB requires not only a smaller number of communications but also a smaller number of iterations, compared to {censoring-based GD}, to attain the same objective error, .

\begin{figure}[!htb]
\centering
\begin{subfigure}[b]{.25\textwidth}
  \centering
  \includegraphics[width=\linewidth]{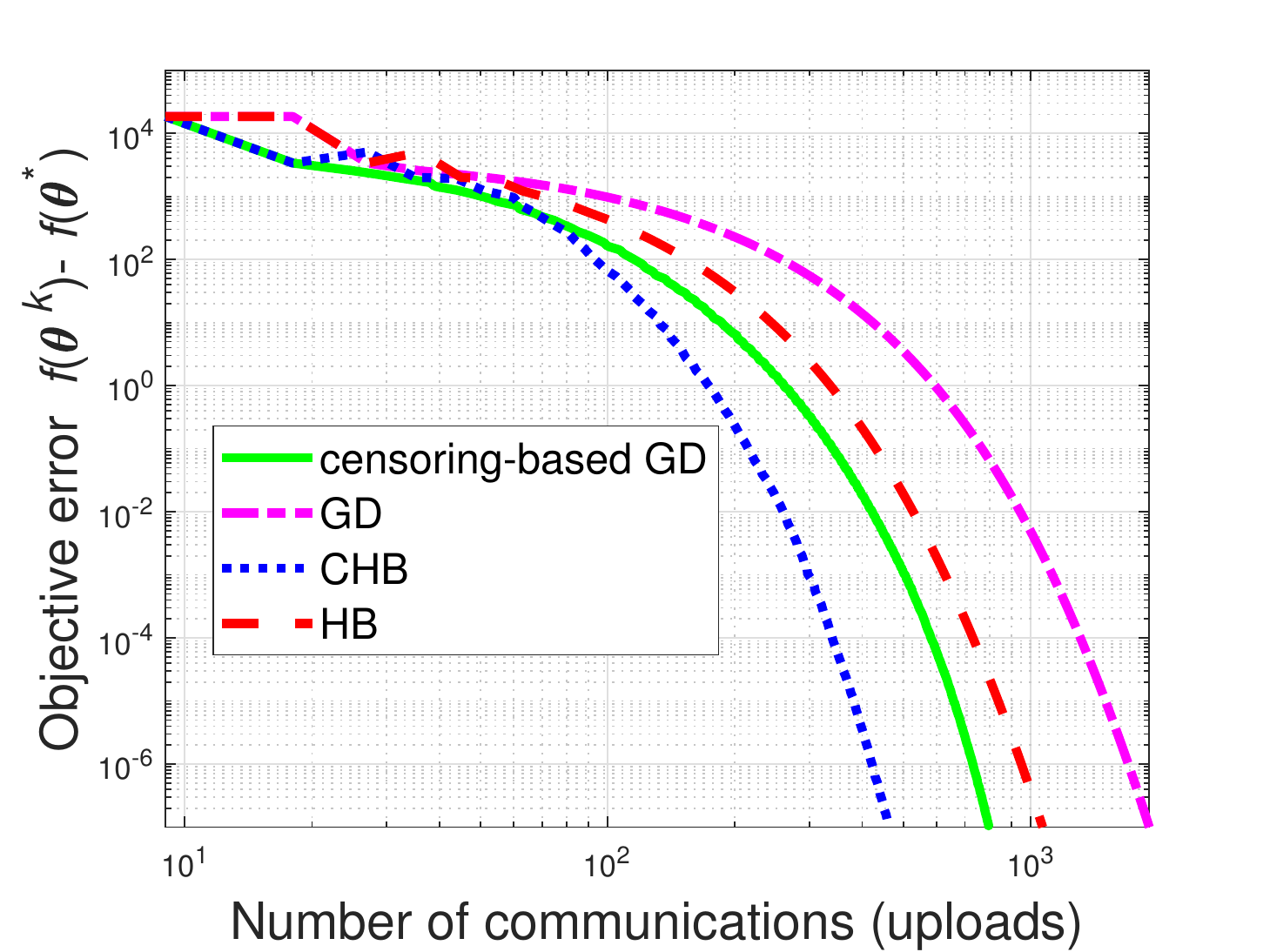}
  \caption*{(a) Linear regression}\label{fig:image1Real}
\end{subfigure}%
\begin{subfigure}[b]{.25\textwidth}
  \centering
  \includegraphics[width=\linewidth]{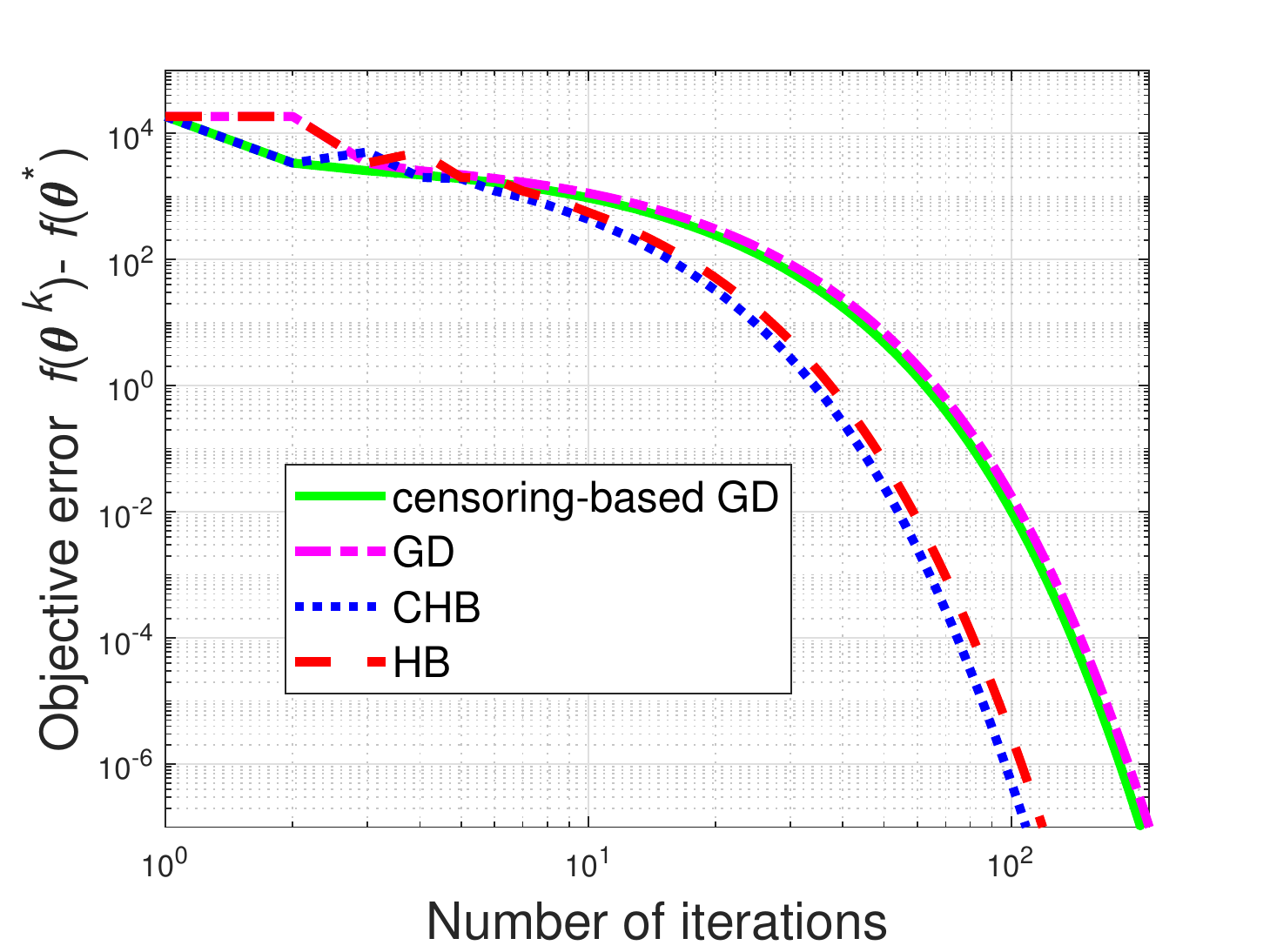}
  \caption*{(b) Linear regression}\label{fig:image2Real}
\end{subfigure}
\begin{subfigure}[b]{.25\textwidth}
  \centering
  \includegraphics[width=\linewidth]{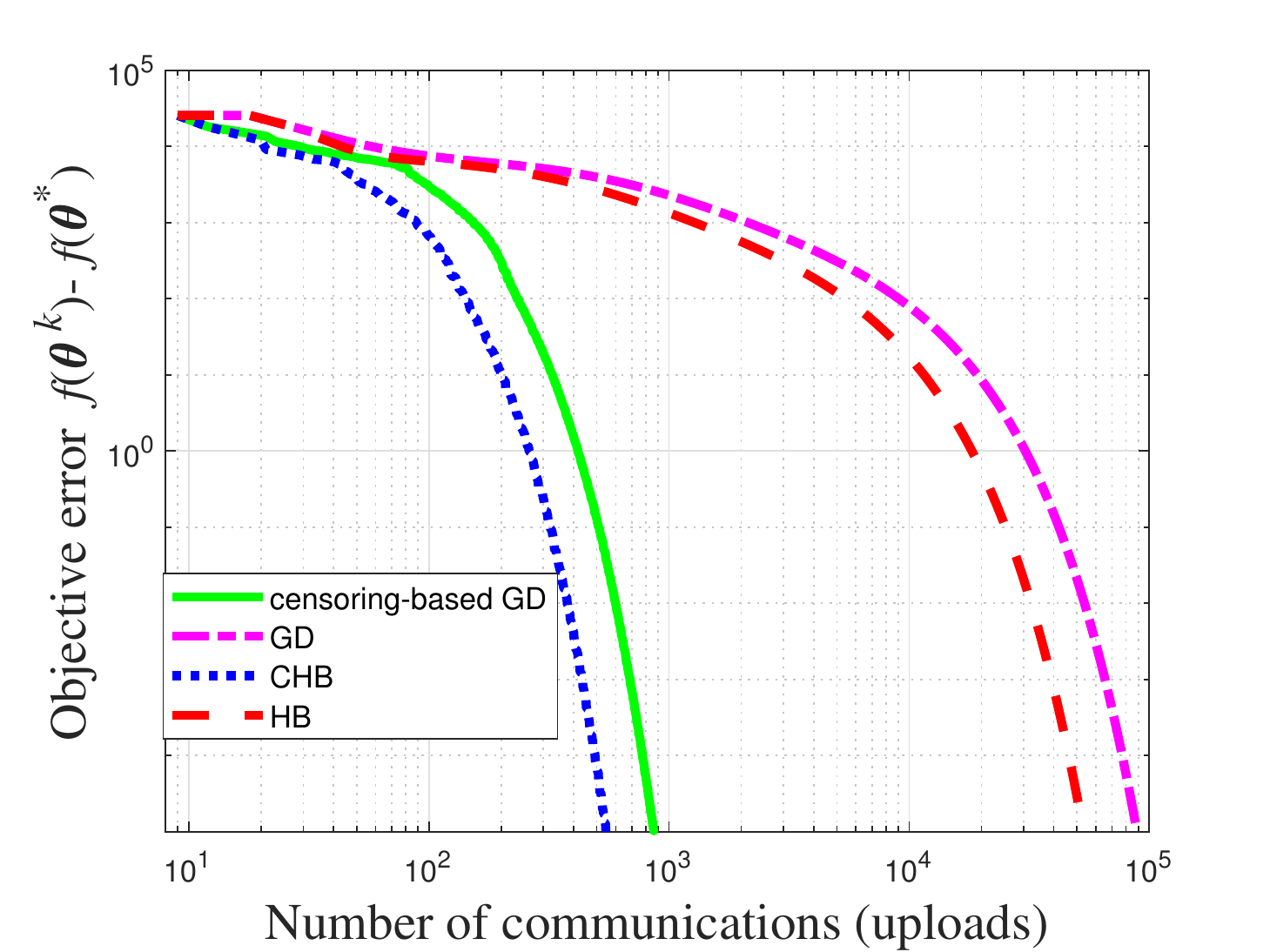}
  \caption*{(c) Logistic regression}\label{fig:image3Real}
\end{subfigure}%
\begin{subfigure}[b]{.25\textwidth}
  \centering
  \includegraphics[width=\linewidth]{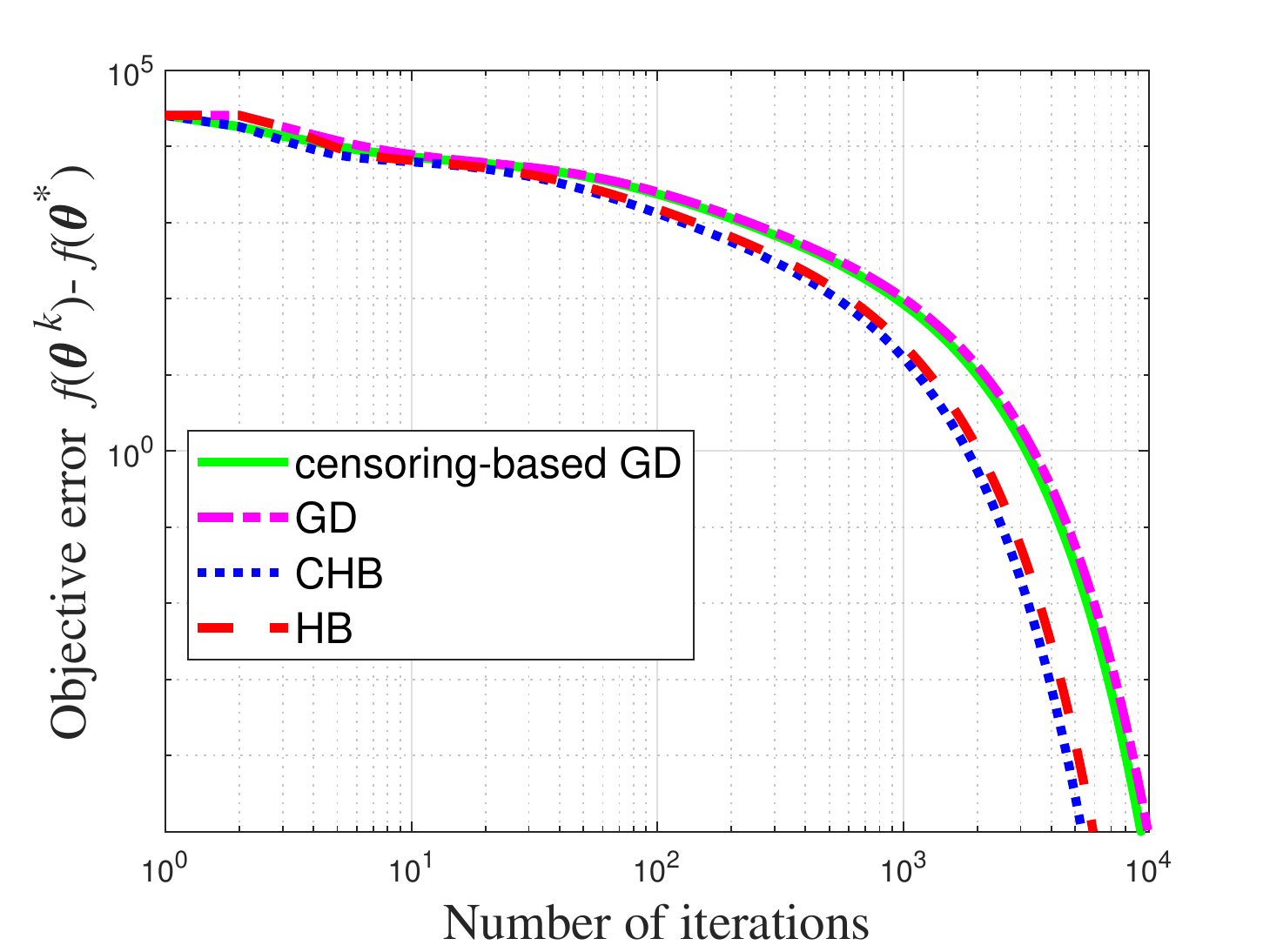}
  \caption*{(d) Logistic regression}\label{fig:image4Real}
\end{subfigure}
\caption{Objective error versus the number of communications and iterations for linear regression and logistic regression in the \emph{ijcnn1} dataset.}\label{fig:imageRealall}
\end{figure}


\begin{figure}[!htb]
\centering
\begin{subfigure}[b]{.25\textwidth}
  \centering
  \includegraphics[width=\linewidth]{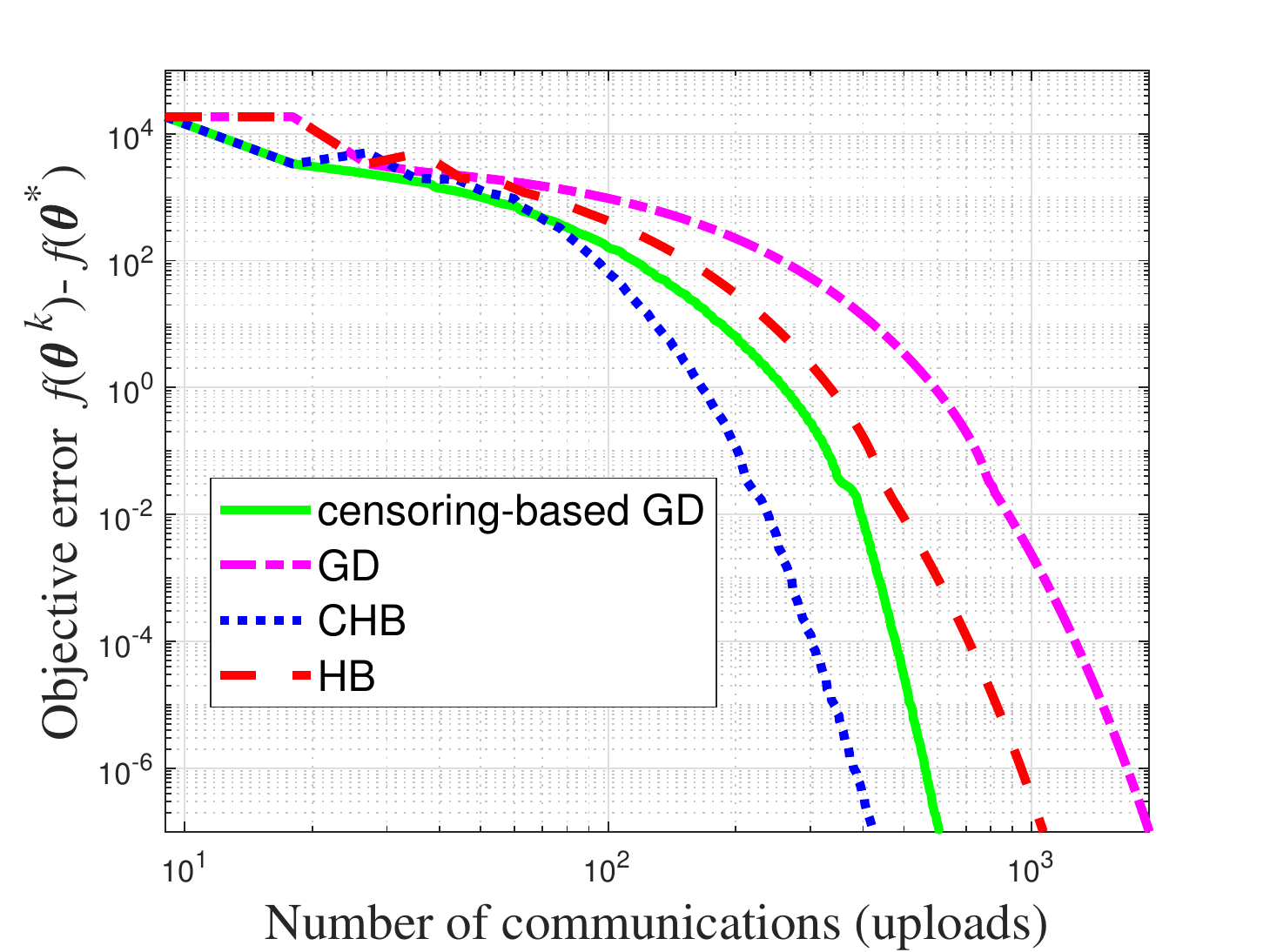}
  \caption*{(a) Lasso regression}\label{fig:image1Lasso}
\end{subfigure}%
\begin{subfigure}[b]{.25\textwidth}
  \centering
  \includegraphics[width=\linewidth]{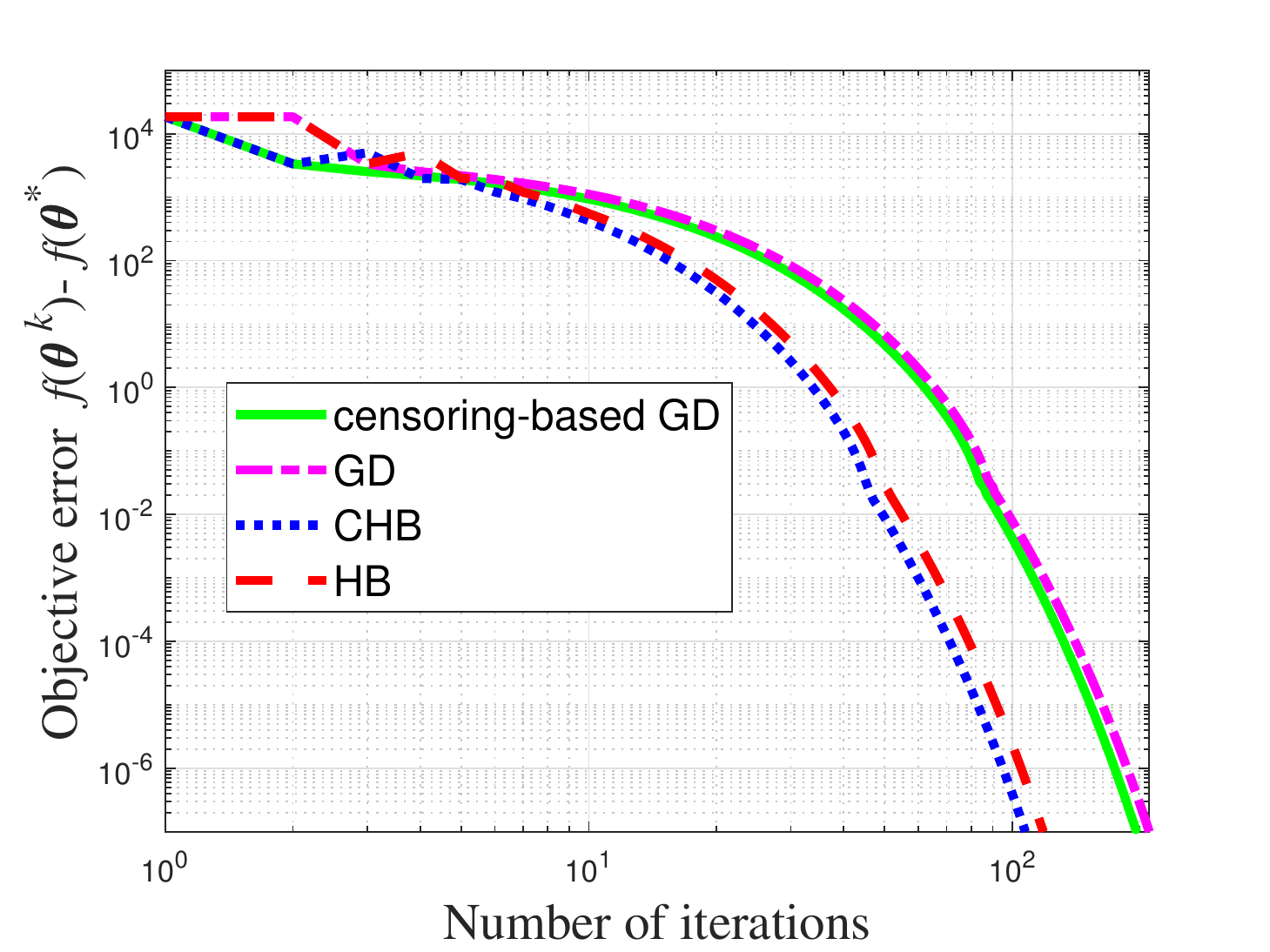}
  \caption*{(b) Lasso regression}\label{fig:image2Lasso}
\end{subfigure}
\begin{subfigure}[b]{.25\textwidth}
  \centering
  \includegraphics[width=\linewidth]{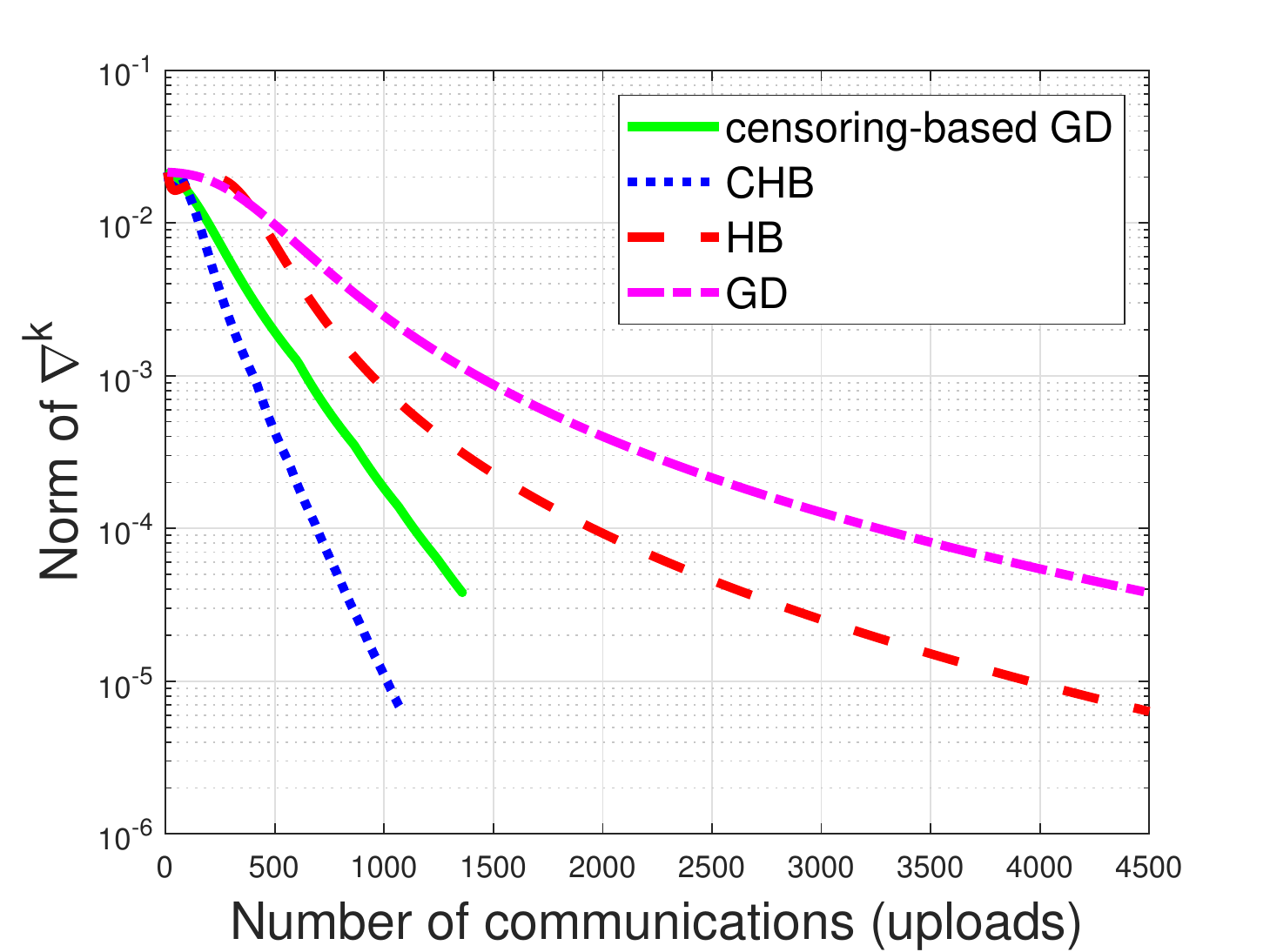}
  \caption*{(c) Neural network}\label{fig:image3NN}
\end{subfigure}%
\begin{subfigure}[b]{.25\textwidth}
  \centering
  \includegraphics[width=\linewidth]{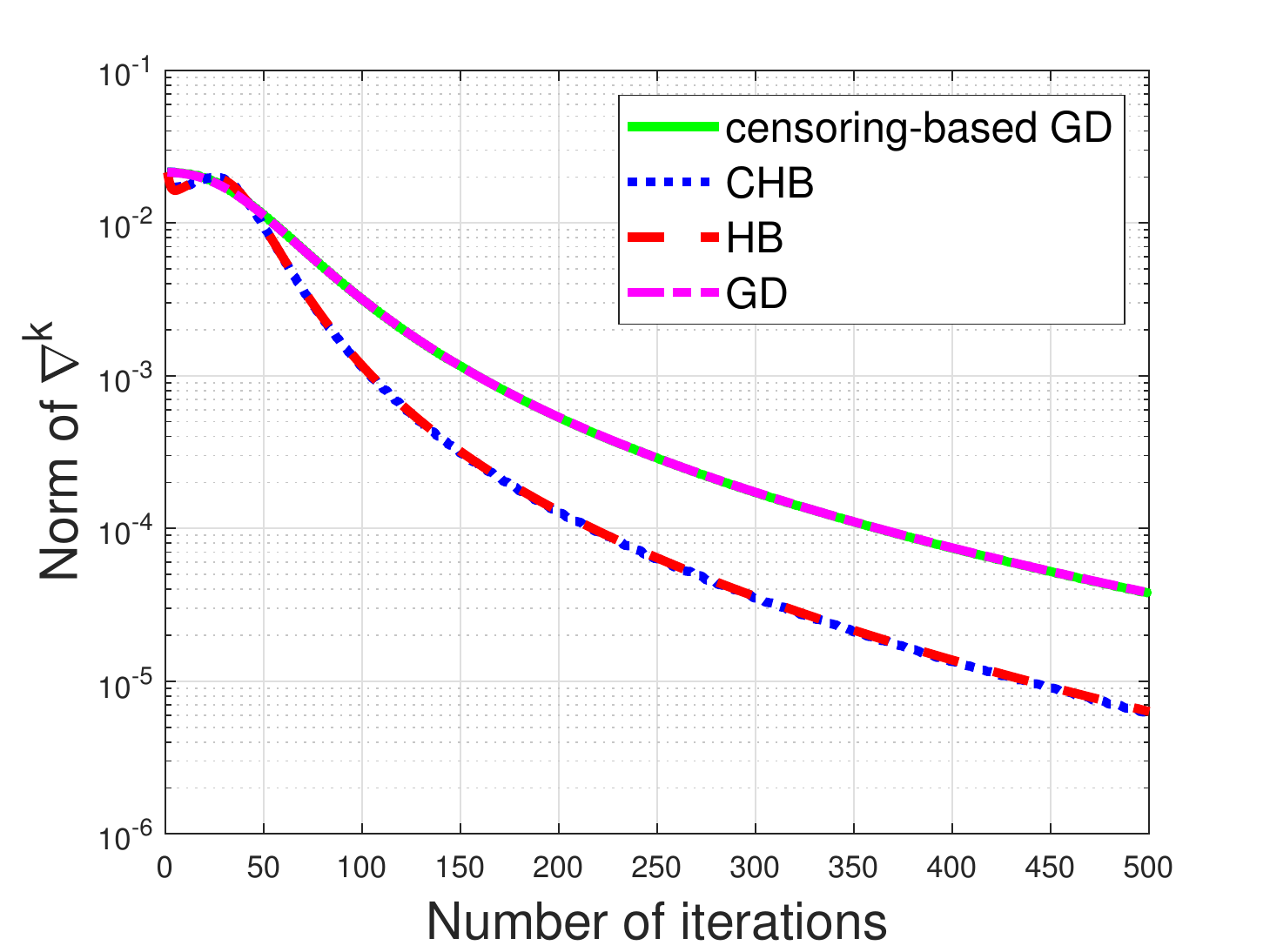}
  \caption*{(d) Neural network}\label{fig:image4NN}
\end{subfigure}
\caption{Objective error versus the number of communications and iterations for lasso regression and training a neural network in the \emph{ijcnn1} dataset.}\label{fig:imageNNall}
\end{figure}

\begin{table*}[htbp]
  \caption{Performance comparison for the \emph{ijcnn1} dataset. The objective error is $10^{-7}$, $10^{-7}$ and $10^{-5}$ for linear regression, lasso regression and logistic regression, respectively. The fixed number of iterations for training a neural network is $500$.}
  \label{GD-table}
  \centering
  \begin{tabular}{lllllllll}
    \toprule
    &\multicolumn{2}{c}{Linear regression}&\multicolumn{2}{c}{Lasso regression}&\multicolumn{2}{c}{Logistic regression} & \multicolumn{2}{c}{Neural network}                \\
    \cmidrule(r){2-3}\cmidrule(r){4-5}\cmidrule(r){6-7}\cmidrule(r){8-9}
    Name     & Comm.      & Iter.   & Comm.   & Iter.   & Comm.   & Iter.  & Comm.    &  Norm square grad.\\
    \midrule
    CHB & $\textbf{465}$  & $\textbf{109}$ & $\textbf{424}$ & $\textbf{108}$ & $\textbf{546}$ & $\textbf{5324}$ & $\textbf{1083}$ & $\mathbf{6.2402\times 10^{-6}}$ \\
    HB  & $1071$  & $119$ & ${1071}$ & ${119}$ & $53244$ & $5916$ & $4500$ & $6.3354\times10^{-6}$   \\
    LAG & $799$  & $203$ & ${608}$ & ${197}$ & $864$ & $9248$ & $1361$ & $3.7549\times10^{-5}$ \\
    GD  & $1917$  & $213$ & ${1899}$ & ${211}$ & $88866$ & $9874$ & $4500$ & $3.7885\times10^{-5}$ \\
    \bottomrule
  \end{tabular}
\end{table*}

\textbf{Real dataset tests.}
We also test the performance on the real dataset \emph{ijcnn1} \cite{prokhorov2001ijcnn} which has 49990 samples and 22 features per sample. All samples are evenly split between nine workers. In addition to linear regression and logistic regression, we also consider lasso regression where we employ a subgradient to replace the gradient. In each of these three regression tasks, we set $\alpha=10^{-4}$ for HB, CHB, GD and censoring-based GD. For CHB and censoring-based GD, we choose $\varepsilon_1=0.1/(\alpha^2M^2)$. We set the regularization
parameter to $\lambda=0.001$ and $\lambda=0.5$ for logistic regression and lasso regression, respectively. We stop the training process when the objective error $f(\boldsymbol{\theta}^k)-f(\boldsymbol{\theta}^*)$ is less than $10^{-7}$, $10^{-5}$, $10^{-7}$ for linear regression, logistic regression, and lasso regression, respectively. Figure \ref{fig:imageRealall} indicates that given a target accuracy, the number of iterations needed for CHB is almost the same as that for HB for linear regression and logistic regression, and it also verifies that CHB requires a smaller number of communications than the other methods. Even for the nondifferentiable lasso regression task, the variant of CHB shows similar performance in Figure \ref{fig:imageNNall}(a) and Figure \ref{fig:imageNNall}(b) as in linear regression and logistic regression. In all regression tasks, CHB reduces the communications compared to HB, by up to several orders of magnitude.

We also evaluate the performance of CHB for training a relatively small neural network with one hidden layer with $30$ nodes. For training the neural network, all algorithms are stopped after $500$ iterations with the step size $\alpha=0.02$ while employing the regularization parameter $\lambda=1/49990$. For CHB and censoring-based GD, we choose $\varepsilon_1=0.01$. Compared to HB, GD, and censoring-based GD, Figure \ref{fig:imageNNall}(c) and Figure \ref{fig:imageNNall}(d) illustrate that CHB has competitive convergence performance for this nonconvex problem with the smallest number of communications. Table \ref{GD-table} summarizes the number of communications and iterations to attain a given objective error which indicates the desirable performance of CHB in terms of communication savings and convergence rate. For linear and lasso regression, CHB reduces the communications to less than half that needed for HB, and for logistic regression the communications are reduced by two orders of magnitude compared to HB. For the neural network training example, CHB and HB reduce the norm square of the gradient by an order of magnitude compared to censoring-based  GD and GD, and CHB uses only $24\%$ of the communications needed for HB. Additional numerical results on other real datasets and discussion of the impact of step size are given {next}.

\subsection{{Experiment Set 2}}

\begin{figure}[!htb]
\centering
\begin{subfigure}[b]{.25\textwidth}
  \centering
  \includegraphics[width=\linewidth]{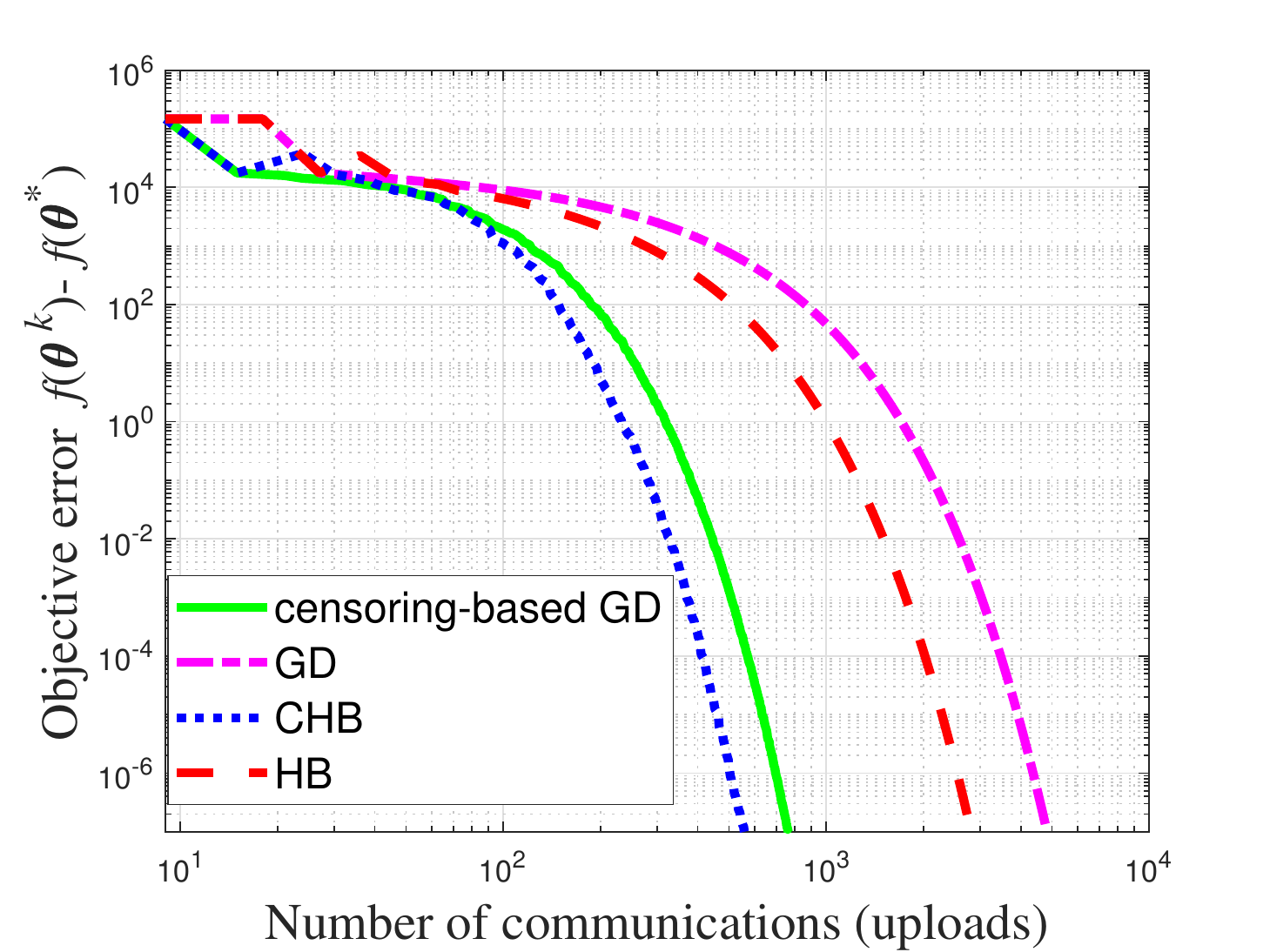}
  \caption*{(a) Linear regression}\label{fig:LinearHousingBodyfatAbalone1}
\end{subfigure}%
\begin{subfigure}[b]{.25\textwidth}
  \centering
  \includegraphics[width=\linewidth]{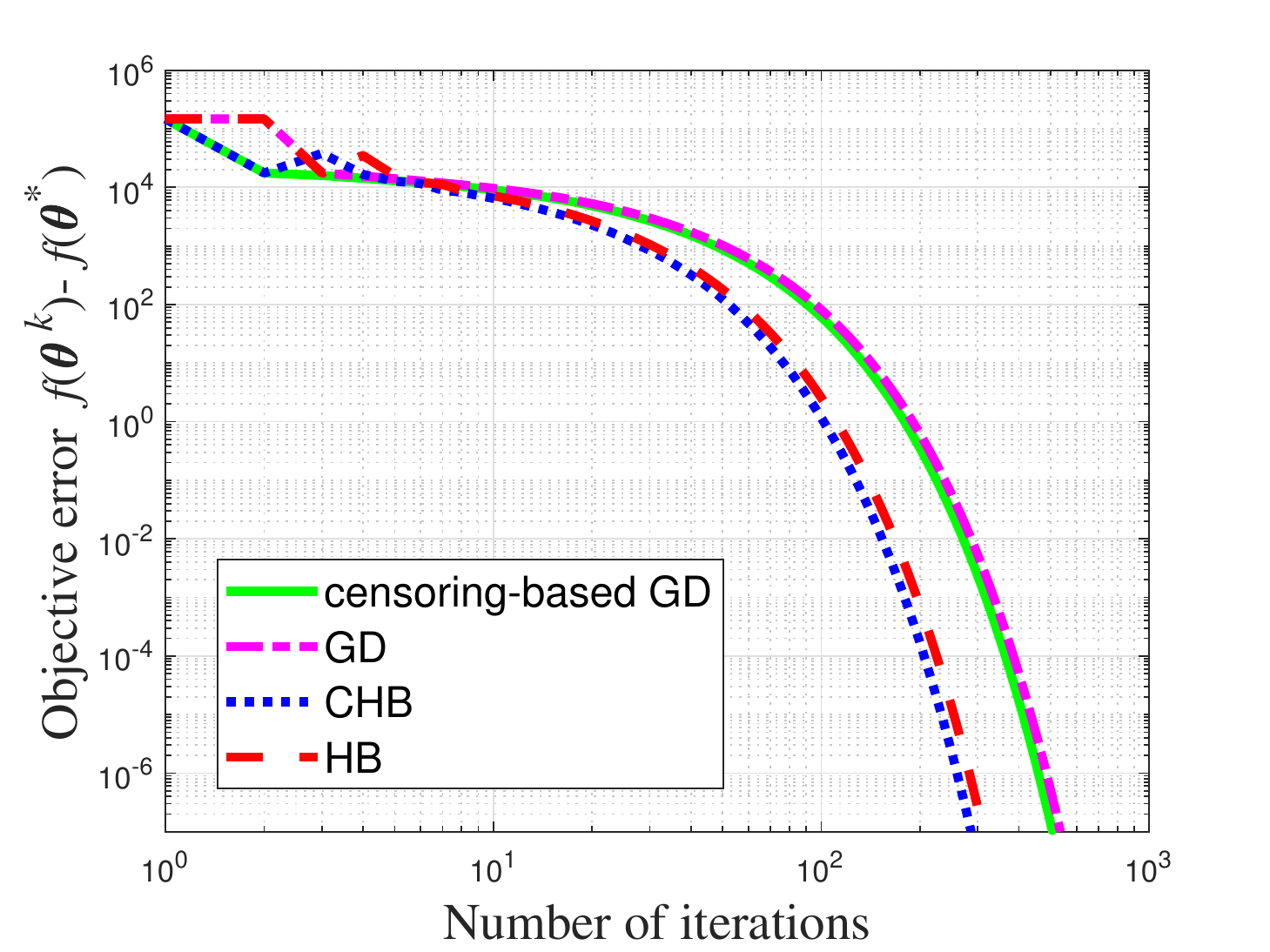}
  \caption*{(b) Linear regression}\label{fig:LinearHousingBodyfatAbalone2}
\end{subfigure}
\begin{subfigure}[b]{.25\textwidth}
  \centering
  \includegraphics[width=\linewidth]{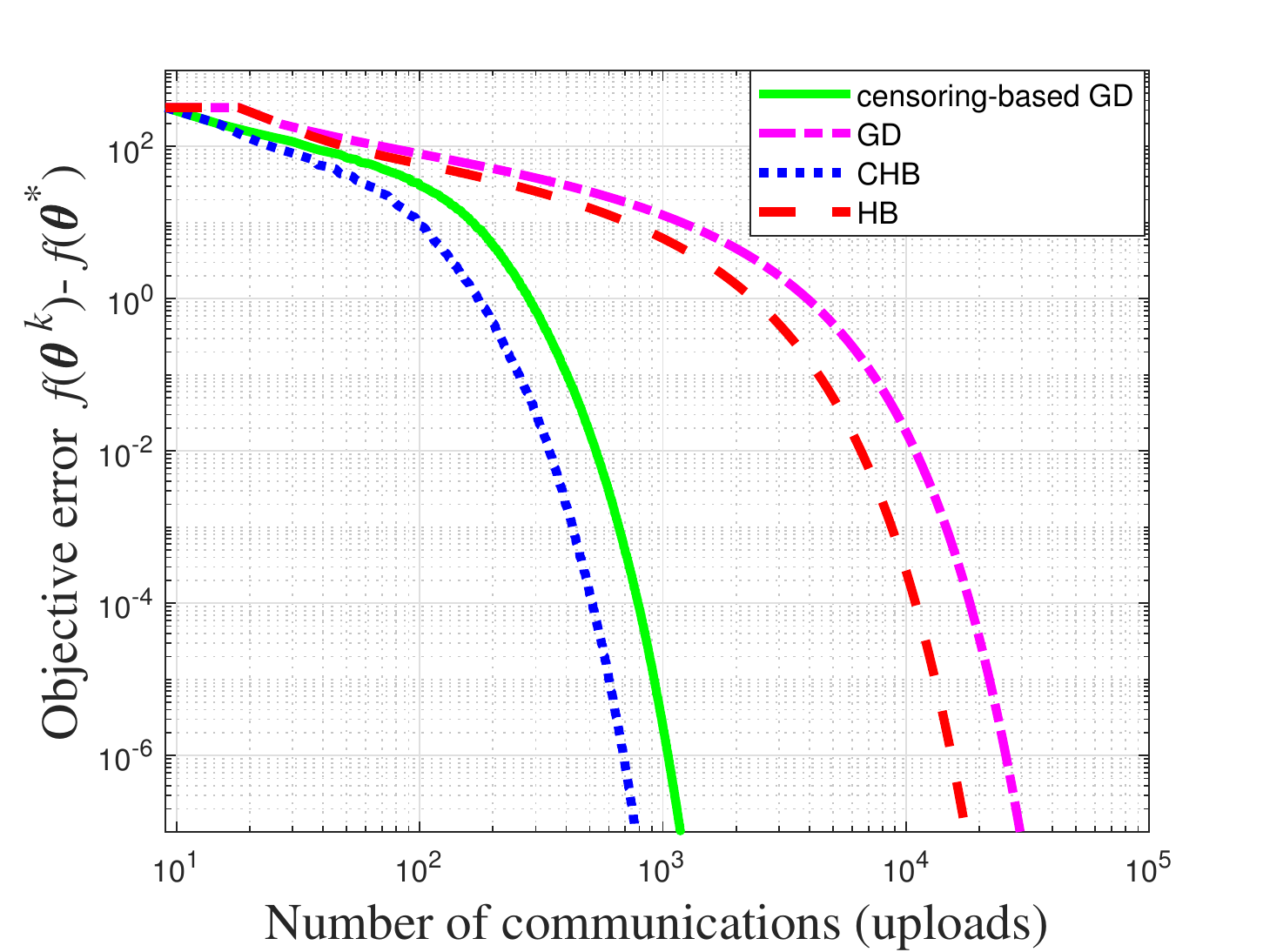}
  \caption*{(c) Logistic regression}\label{fig:logsticIonosphereAdultfatDerm1}
\end{subfigure}%
\begin{subfigure}[b]{.25\textwidth}
  \centering
  \includegraphics[width=\linewidth]{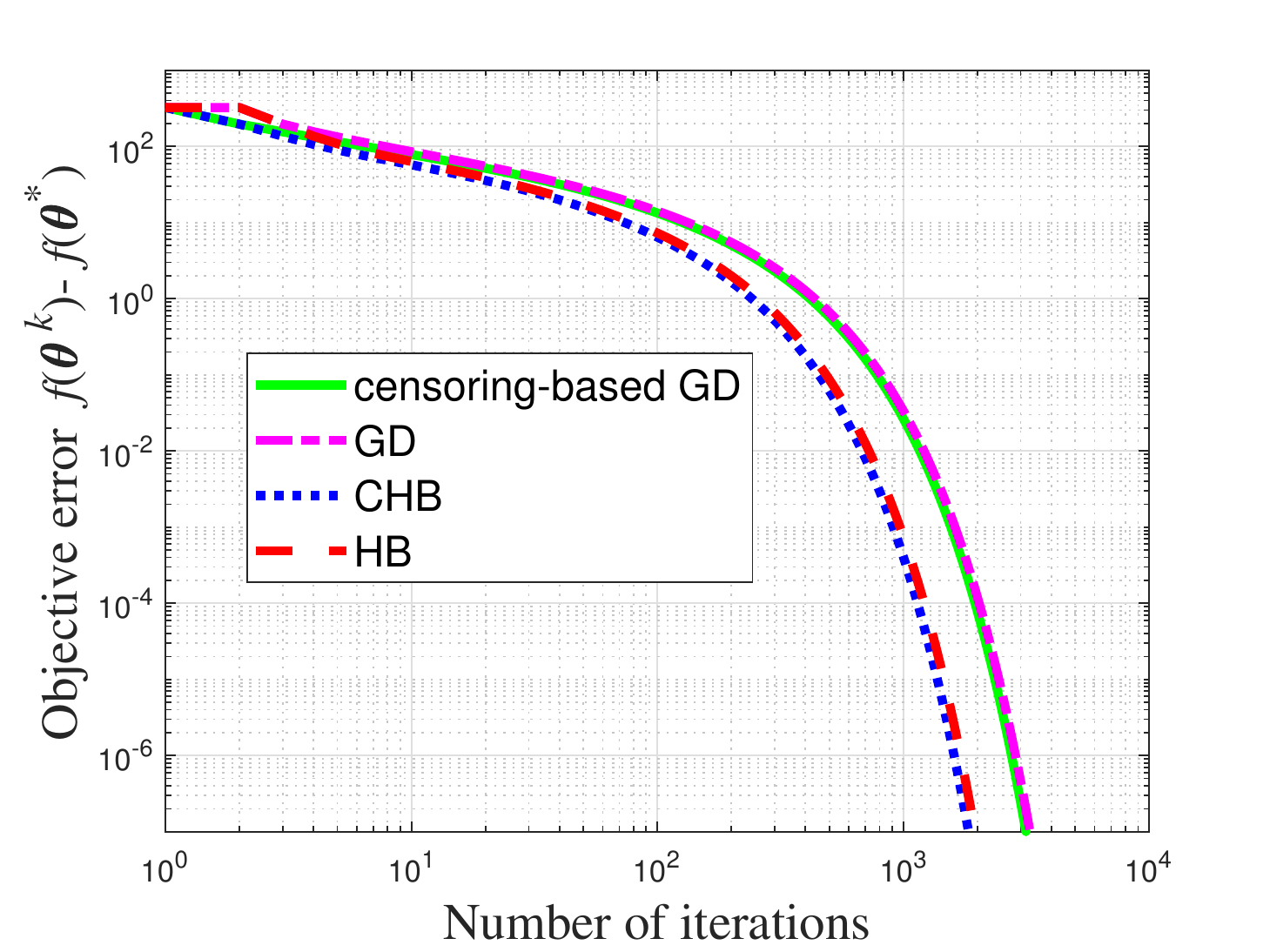}
  \caption*{(d) Logistic regression}\label{fig:logsticIonosphereAdultfatDerm2}
\end{subfigure}
\caption{Objective error for linear regression and logistic regression in the \emph{Housing}, \emph{Body fat}, \emph{Abalone}, \emph{Ionosphere}, \emph{Adult fat}, and \emph{Derm} datasets.}\label{fig:samesetingLAGreal}
\end{figure}

\begin{figure}[!htb]
\centering
\begin{subfigure}[b]{.25\textwidth}
  \centering
  \includegraphics[width=\linewidth]{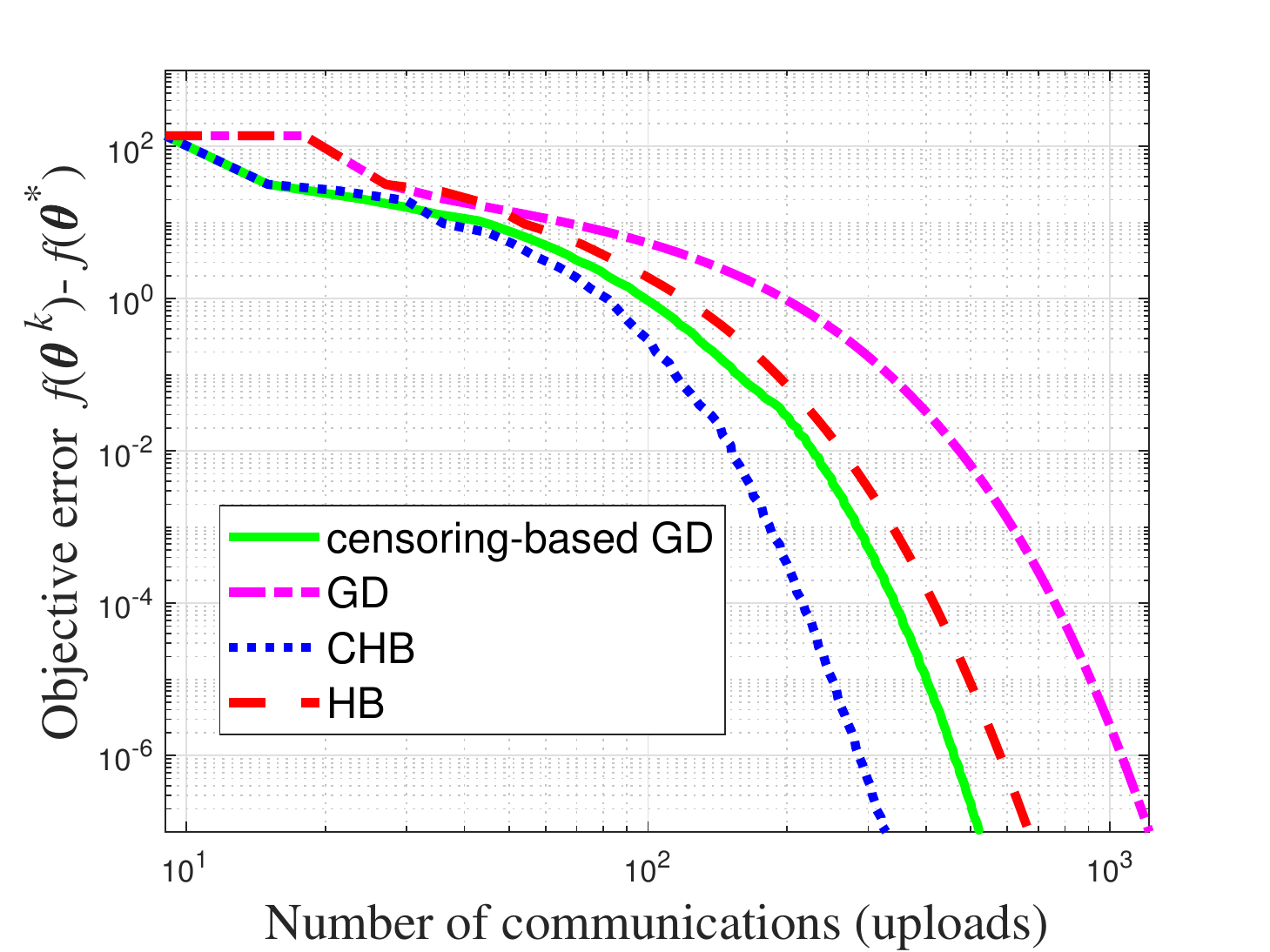}
  \caption*{(a) Lasso regression}\label{fig:LassoIonsphereAdultDerm1}
\end{subfigure}%
\begin{subfigure}[b]{.25\textwidth}
  \centering
  \includegraphics[width=\linewidth]{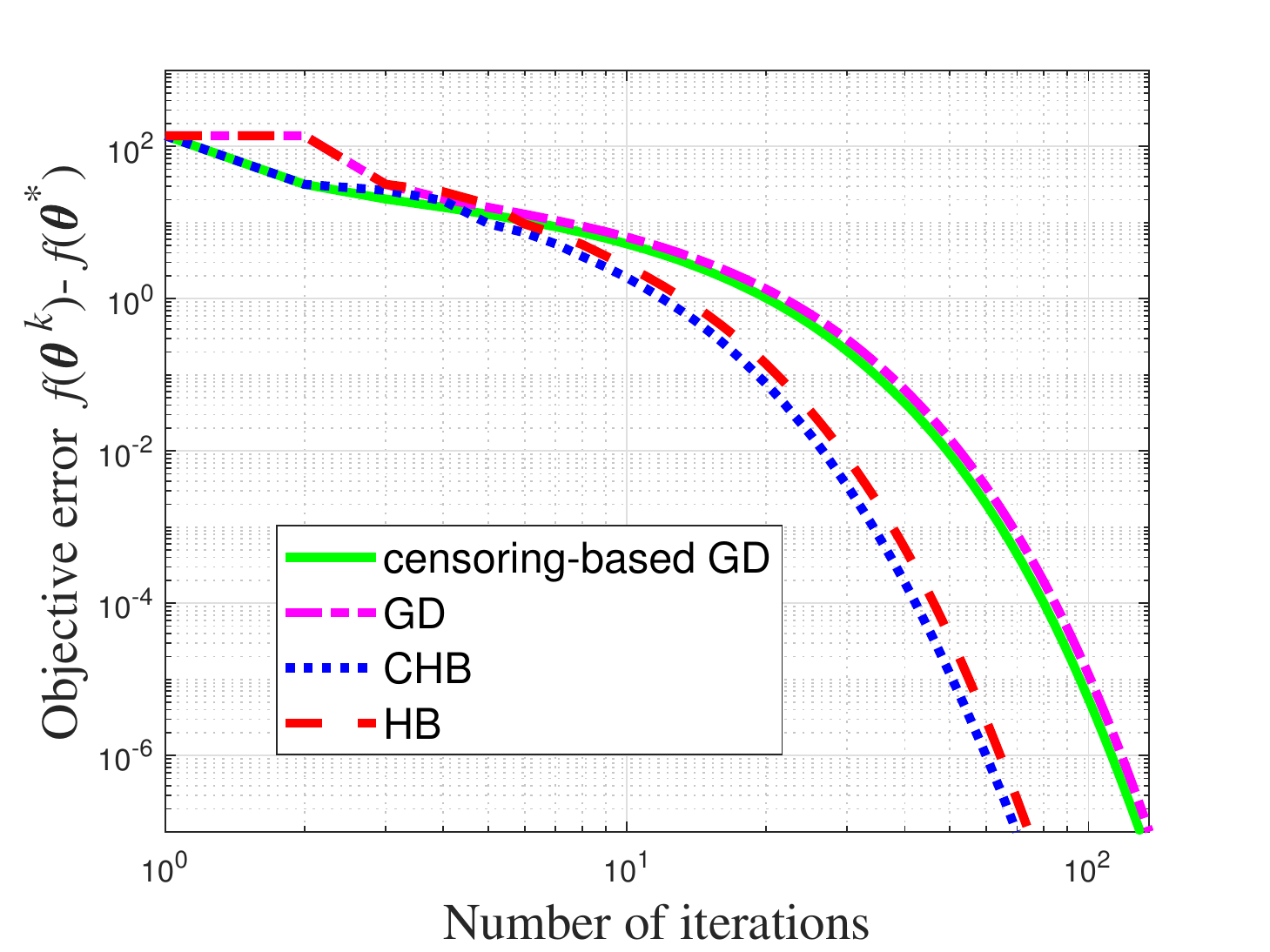}
  \caption*{(b) Lasso regression}\label{fig:LassoIonsphereAdultDerm2}
\end{subfigure}
\begin{subfigure}[b]{.25\textwidth}
  \centering
  \includegraphics[width=\linewidth]{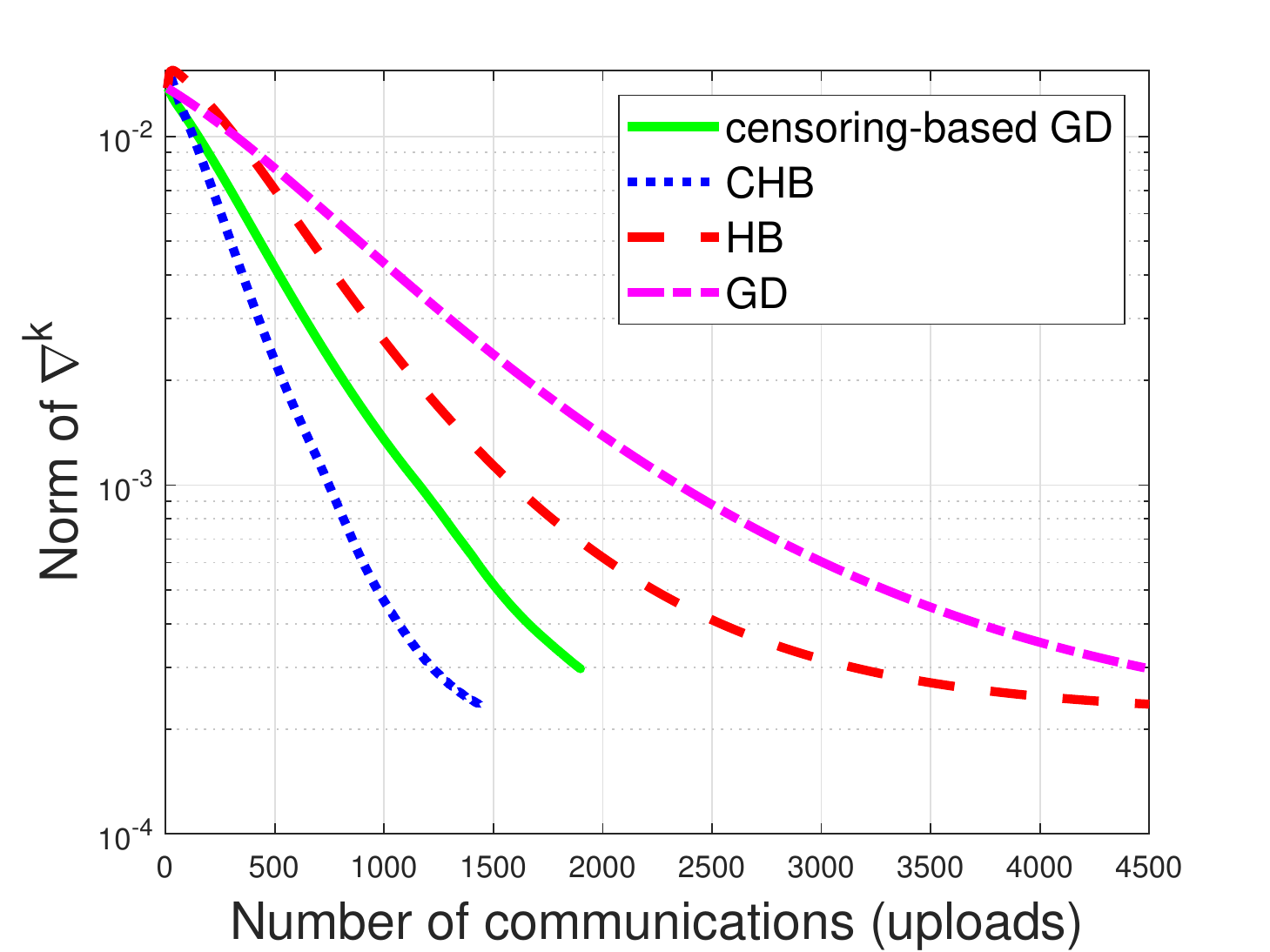}
  \caption*{(c) Neural network}\label{fig:NNadult1}
\end{subfigure}%
\begin{subfigure}[b]{.25\textwidth}
  \centering
  \includegraphics[width=\linewidth]{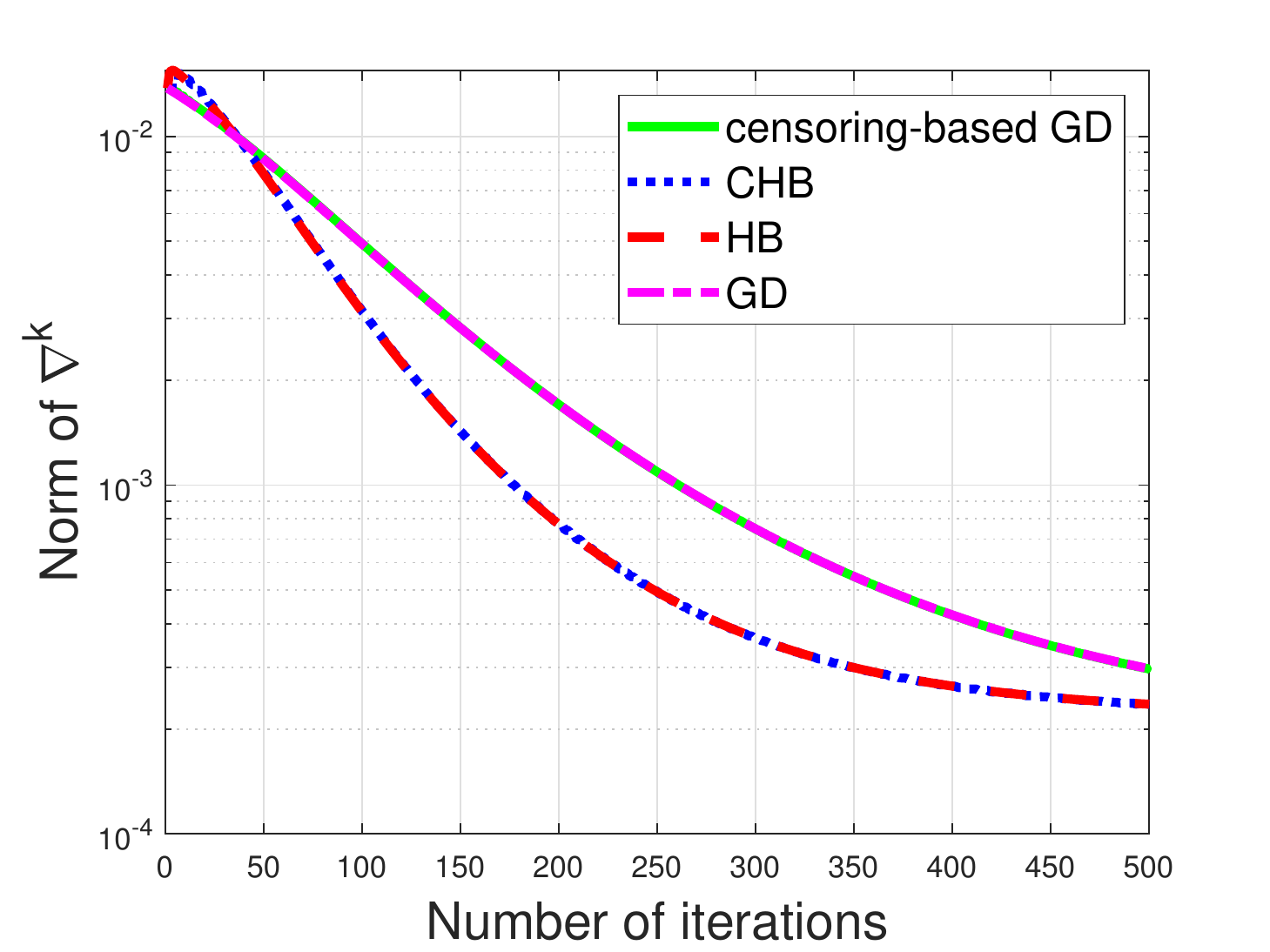}
  \caption*{(d) Neural network}\label{fig:NNadult2}
\end{subfigure}
\caption{Objective error for lasso regression and training a neural network in the \emph{Ionosphere}, \emph{Adult fat}, and \emph{Derm} datasets.}\label{fig:imageNNrealsameaslag}
\end{figure}

To promote comparison, we first use the same real datasets with the same setup as \cite{chen2018lag}. Specifically, we consider linear regression with the \emph{Housing}, \emph{Body fat}, and \emph{Abalone} datasets and logistic regression with the \emph{Ionosphere}, \emph{Adult fat}, and \emph{Derm} datasets. We evenly split each dataset into three workers and employ the minimal number of features among all datasets as the number of features used in the test. Additionally, we also consider lasso regression with the \emph{Ionosphere}, \emph{Adult fat}, and \emph{Derm} datasets. In all these regression tasks, we choose $\alpha=1/L$ for all four algorithms. For CHB and censoring-based GD, we choose $\varepsilon_1=0.1/(\alpha^2M^2)$. For HB and CHB, we choose $\beta=0.4$. The values of the regularization parameter for logistic regression and lasso regression are $\lambda=0.001$ and $\lambda=0.1$, respectively. All regression tasks stop when the objective error is less than $10^{-7}$. Furthermore, we consider training a  neural network with \emph{Adult fat} dataset where we choose $\alpha=0.01$ for all algorithms. For CHB and censoring-based GD, we choose $\varepsilon_1=0.01$. Again, for HB and CHB, we choose $\beta=0.4$. The regularization parameter we choose for the neural network is $\lambda=1/1605$. The optimization process is stopped after $500$ iterations for training the neural network. Figure \ref{fig:samesetingLAGreal} and Figure \ref{fig:imageNNrealsameaslag} indicate the effectiveness of CHB in terms of communication reduction for the four learning tasks considered in this paper. The specific results are summarized in Table \ref{GD-table2realfat}.

\begin{table*}[htbp]
  \caption{Performance comparison in the \emph{Ionosphere, Adult, \emph{and} Derm} dataset. The objective error is $10^{-7}$ for all linear regression tasks. The fixed number of iterations for training the neural network is $500$.}
  \label{GD-table2realfat}
  \centering
  \tabcolsep=0.1cm
  \begin{tabular}{lllllllll}
    \toprule
    &\multicolumn{2}{c}{Linear regression}&\multicolumn{2}{c}{Lasso regression}&\multicolumn{2}{c}{Logistic regression} & \multicolumn{2}{c}{Neural network}                \\
    \cmidrule(r){2-3}\cmidrule(r){4-5}\cmidrule(r){6-7}\cmidrule(r){8-9}
    Name     & Comm.      & Iter.   & Comm.   & Iter.   & Comm.   & Iter.  & Comm.    &  Norm square grad.\\
    \midrule
    CHB & $\textbf{559}$  & $\textbf{287}$ & $\textbf{329}$ & $\textbf{70}$ & $\textbf{772}$ & $\textbf{1842}$ & $\textbf{1436}$ & $\mathbf{2.3566\times 10^{-4}}$ \\
    HB  & $2808$  & $312$ & ${675}$ & ${75}$ & $17595$ & $1955$ & $4500$ & $2.3547\times10^{-4}$   \\
    LAG & $762$  & $507$ & ${522}$ & ${130}$ & $1179$ & $3152$ & $1899$ & $2.9703\times10^{-4}$ \\
    GD  & $4824$  & $536$ & ${1224}$ & ${136}$ & $29439$ & $3271$ & $4500$ & $2.9752\times10^{-4}$ \\
    \bottomrule
  \end{tabular}
\end{table*}

\begin{figure}[!htb]
\centering
\begin{subfigure}[b]{.25\textwidth}
  \centering
  \includegraphics[width=\linewidth]{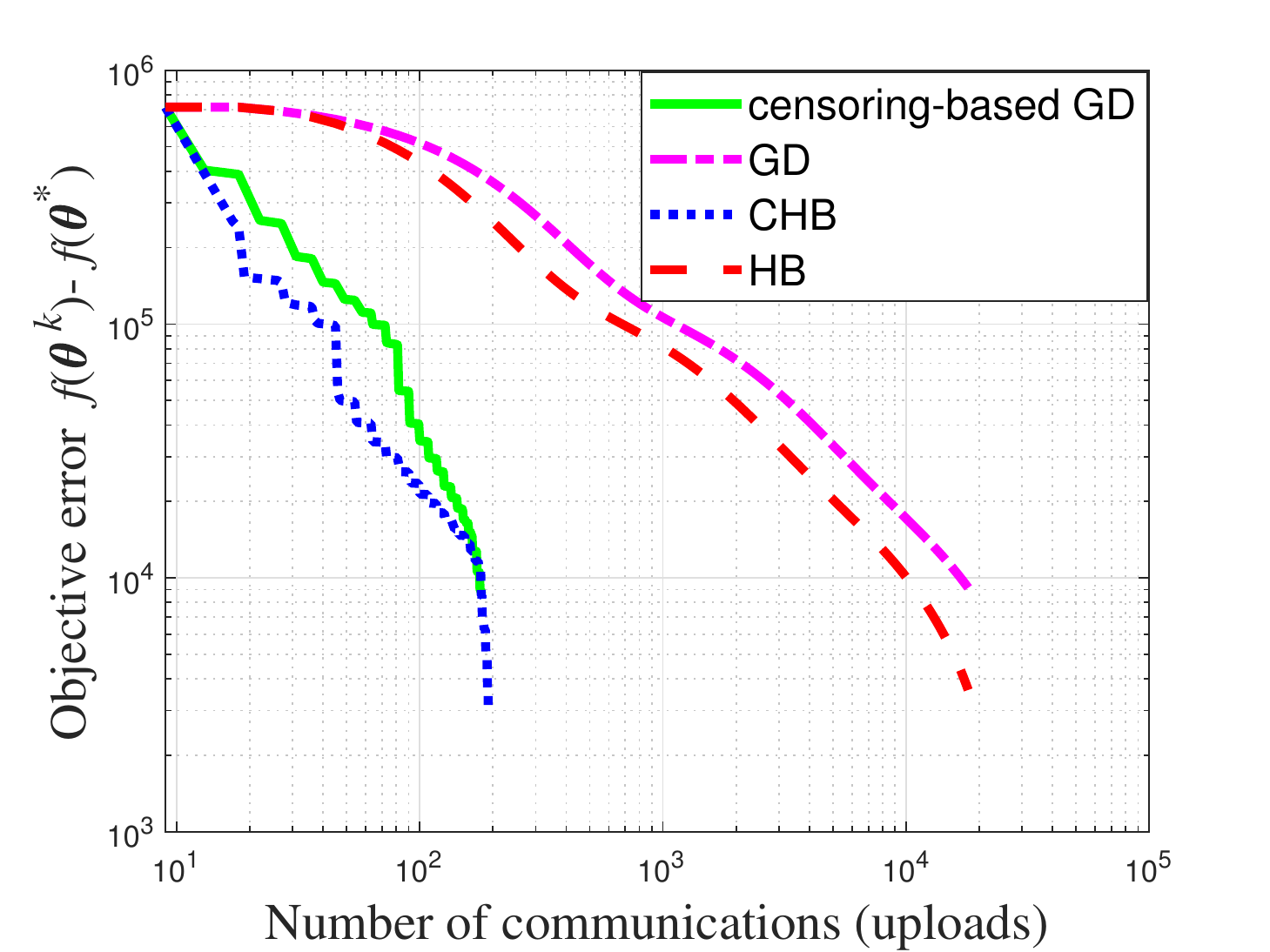}
  \caption*{(a) Linear regression}\label{fig:LinearMNIST1}
\end{subfigure}%
\begin{subfigure}[b]{.25\textwidth}
  \centering
  \includegraphics[width=\linewidth]{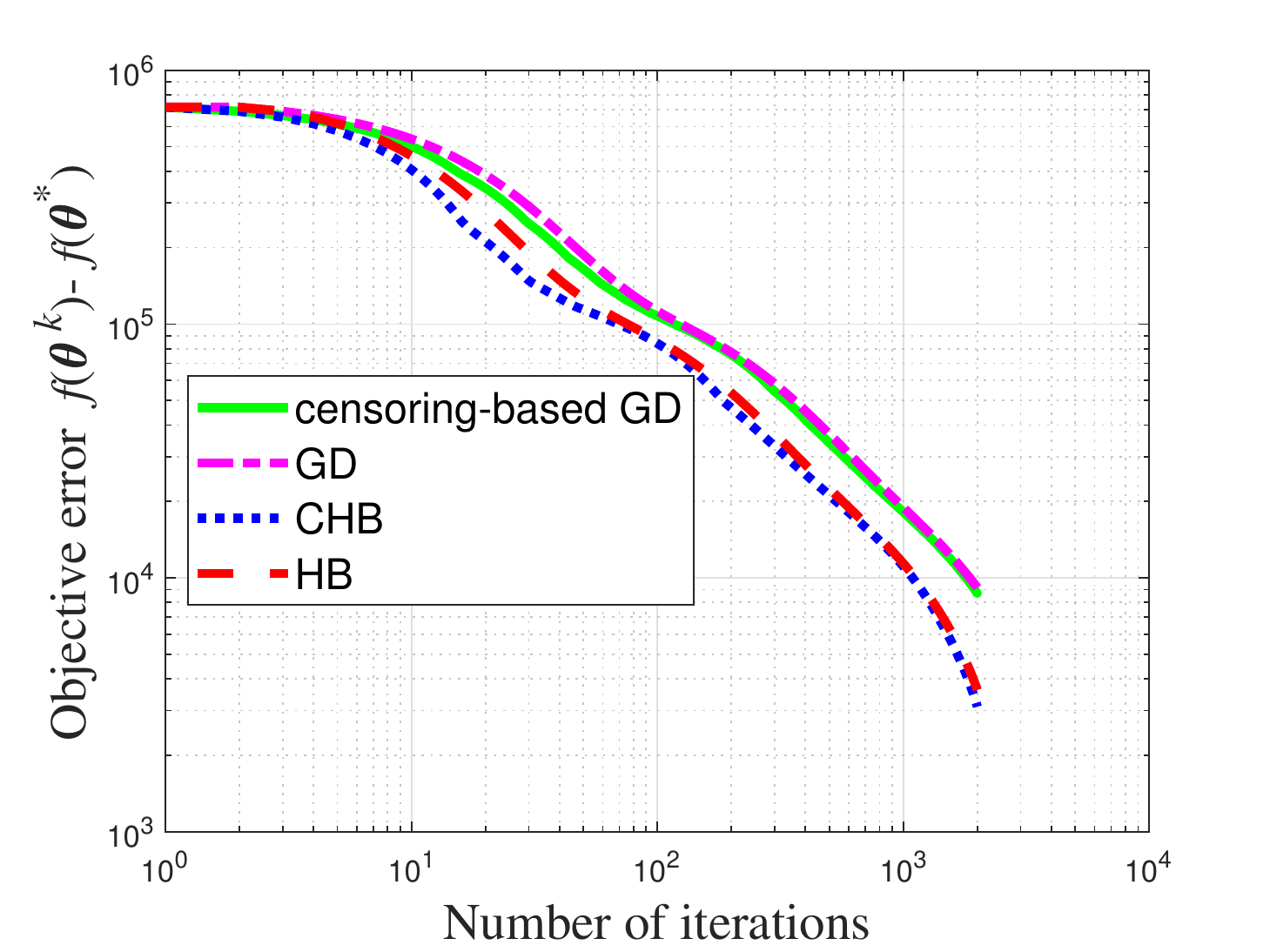}
  \caption*{(b) Linear regression}\label{fig:LinearMNIST2}
\end{subfigure}
\begin{subfigure}[b]{.25\textwidth}
  \centering
  \includegraphics[width=\linewidth]{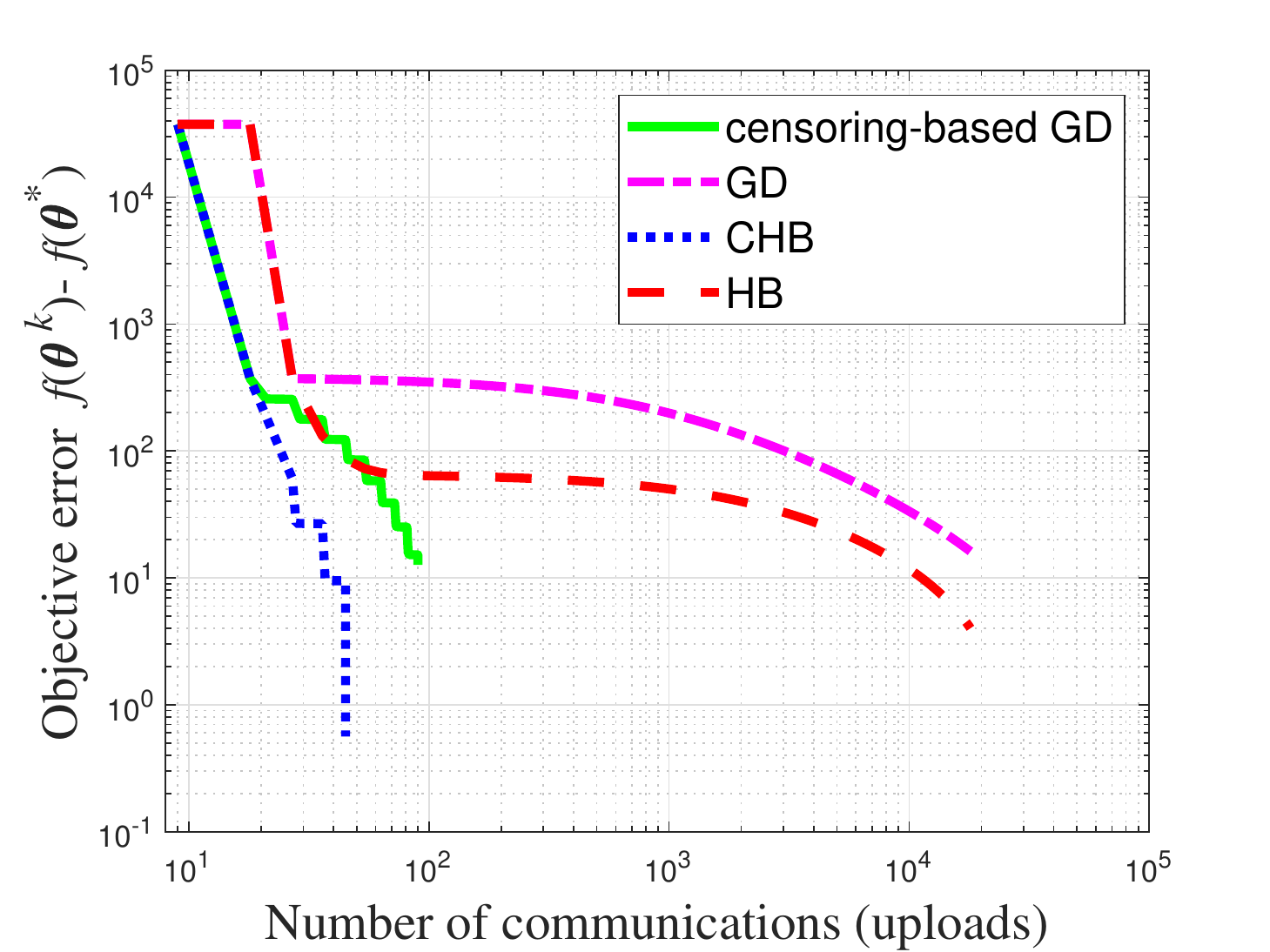}
  \caption*{(c) Logistic regression}\label{fig:logsticMNIST1}
\end{subfigure}%
\begin{subfigure}[b]{.25\textwidth}
  \centering
  \includegraphics[width=\linewidth]{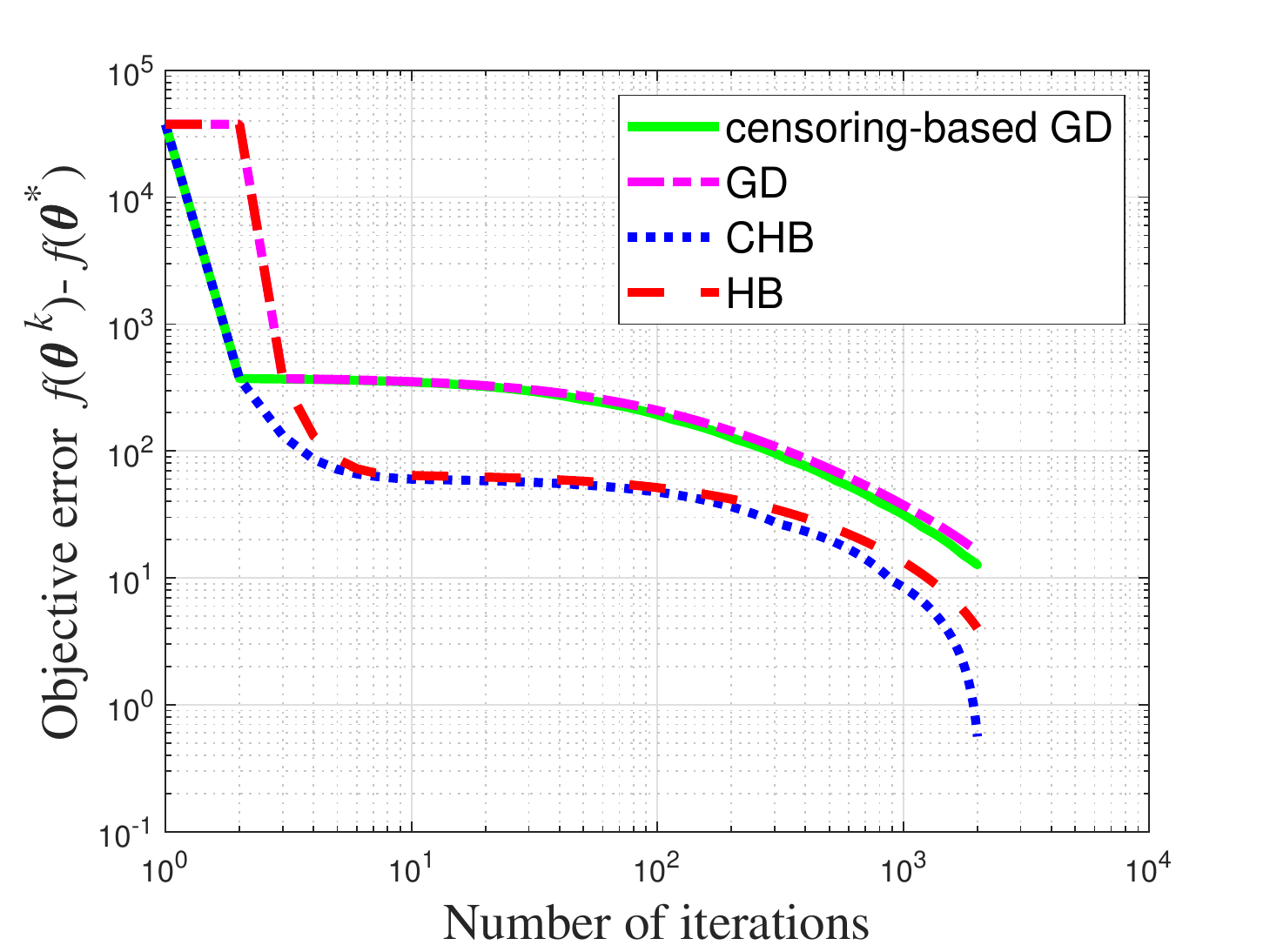}
  \caption*{(d) Logistic regression}\label{fig:logsticMNIST2}
\end{subfigure}
\caption{Objective error for linear regression and logistic regression in the \emph{MNIST} dataset.}\label{fig:MNIST_Linear_Logistic}
\end{figure}

\begin{figure}[!htb]
\centering
\begin{subfigure}[b]{.25\textwidth}
  \centering
  \includegraphics[width=\linewidth]{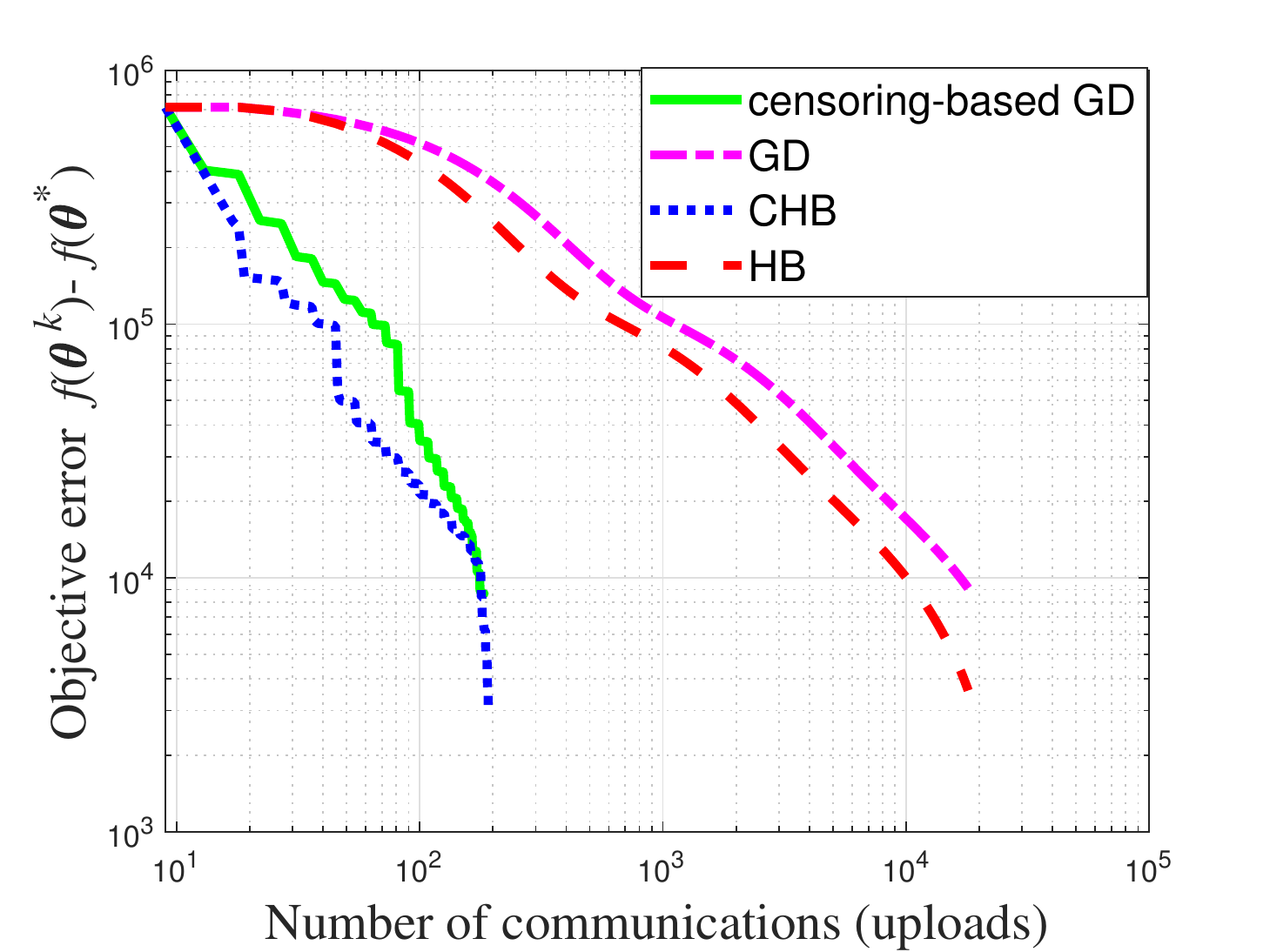}
  \caption*{(a) Lasso regression}\label{fig:LassoMNIST1}
\end{subfigure}%
\begin{subfigure}[b]{.25\textwidth}
  \centering
  \includegraphics[width=\linewidth]{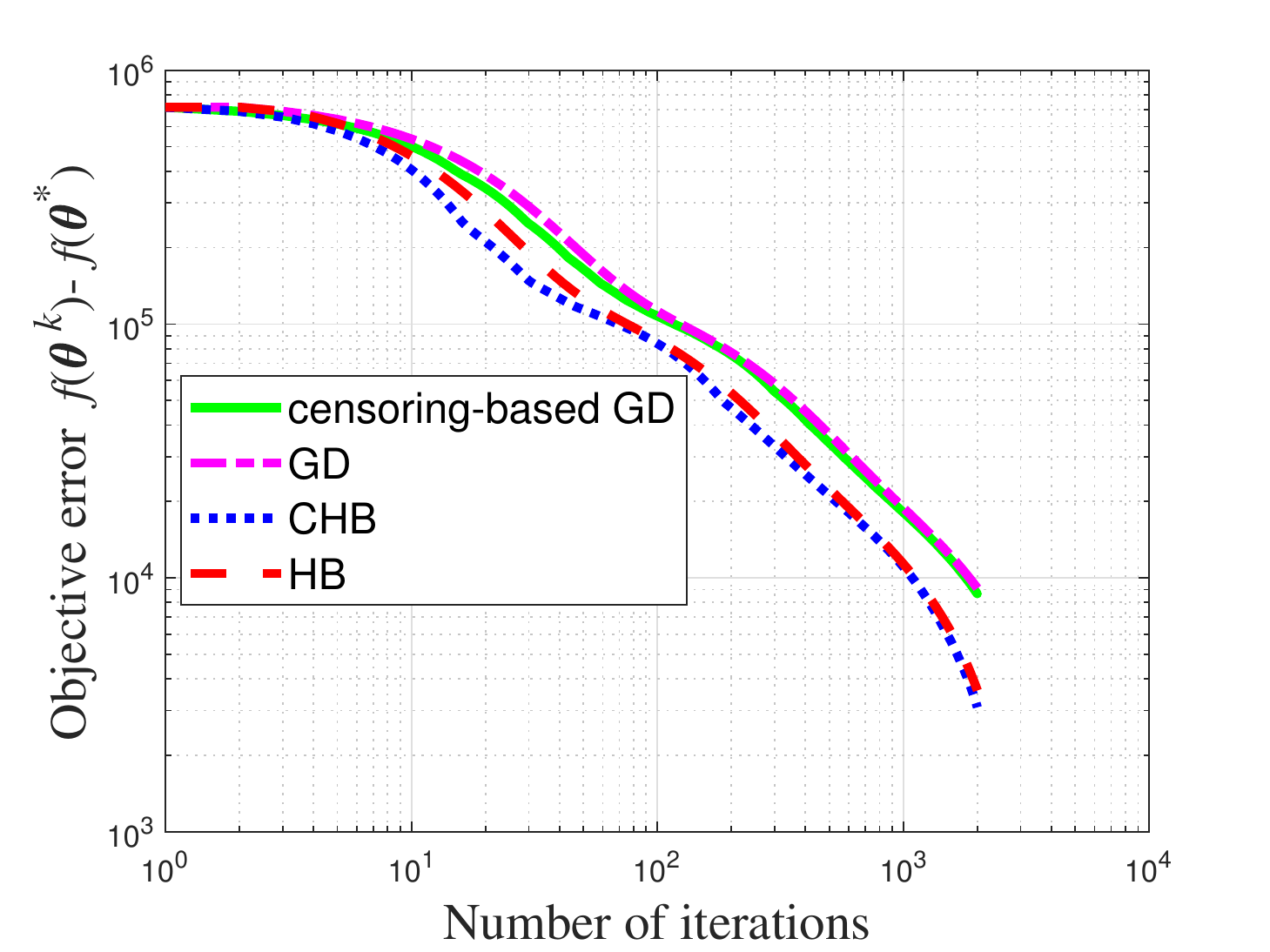}
  \caption*{(b) Lasso regression}\label{fig:LassoMNIST2}
\end{subfigure}
\begin{subfigure}[b]{.25\textwidth}
  \centering
  \includegraphics[width=\linewidth]{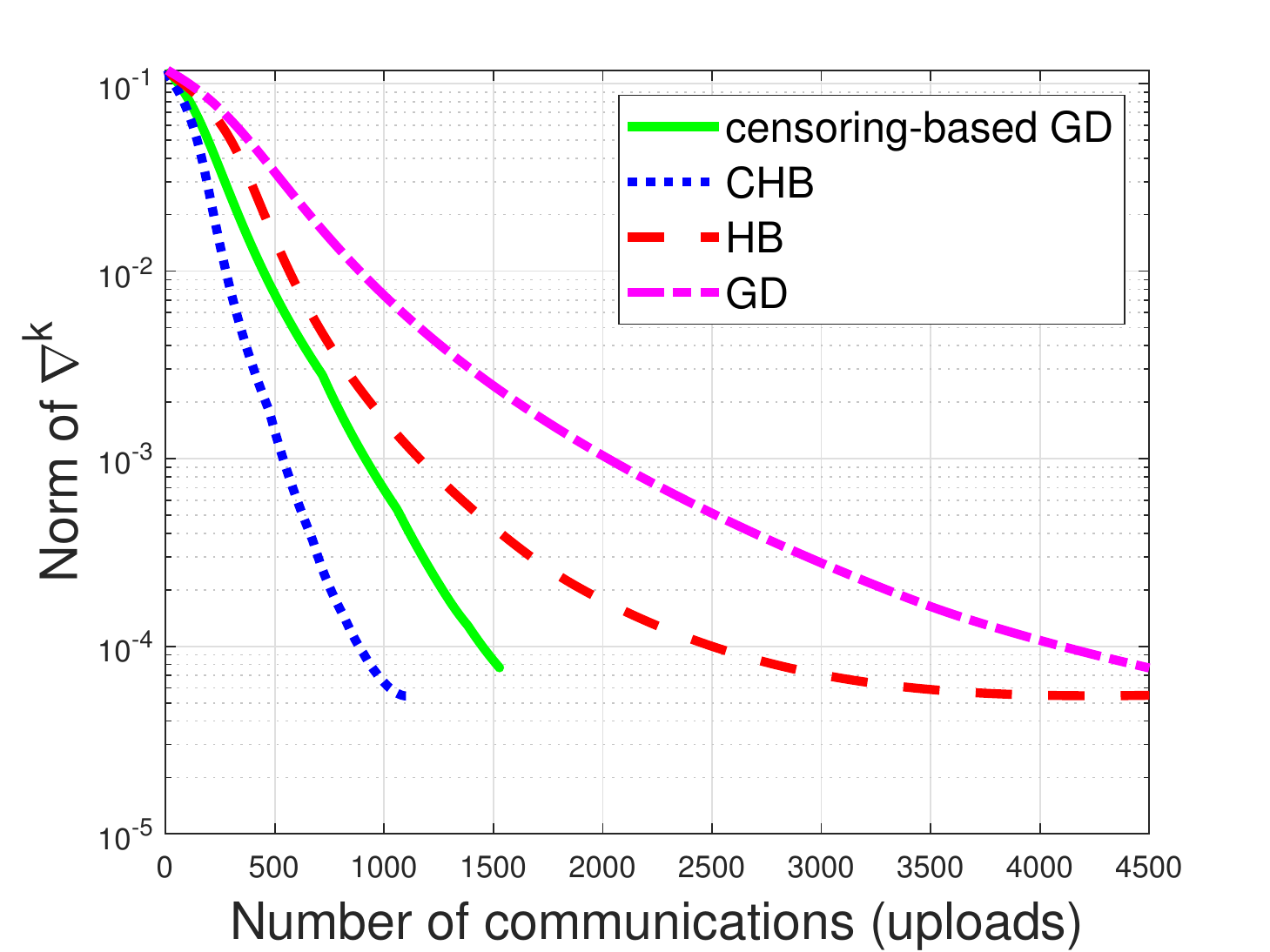}
  \caption*{(c) Neural network}\label{fig:NN_MNIST1}
\end{subfigure}%
\begin{subfigure}[b]{.25\textwidth}
  \centering
  \includegraphics[width=\linewidth]{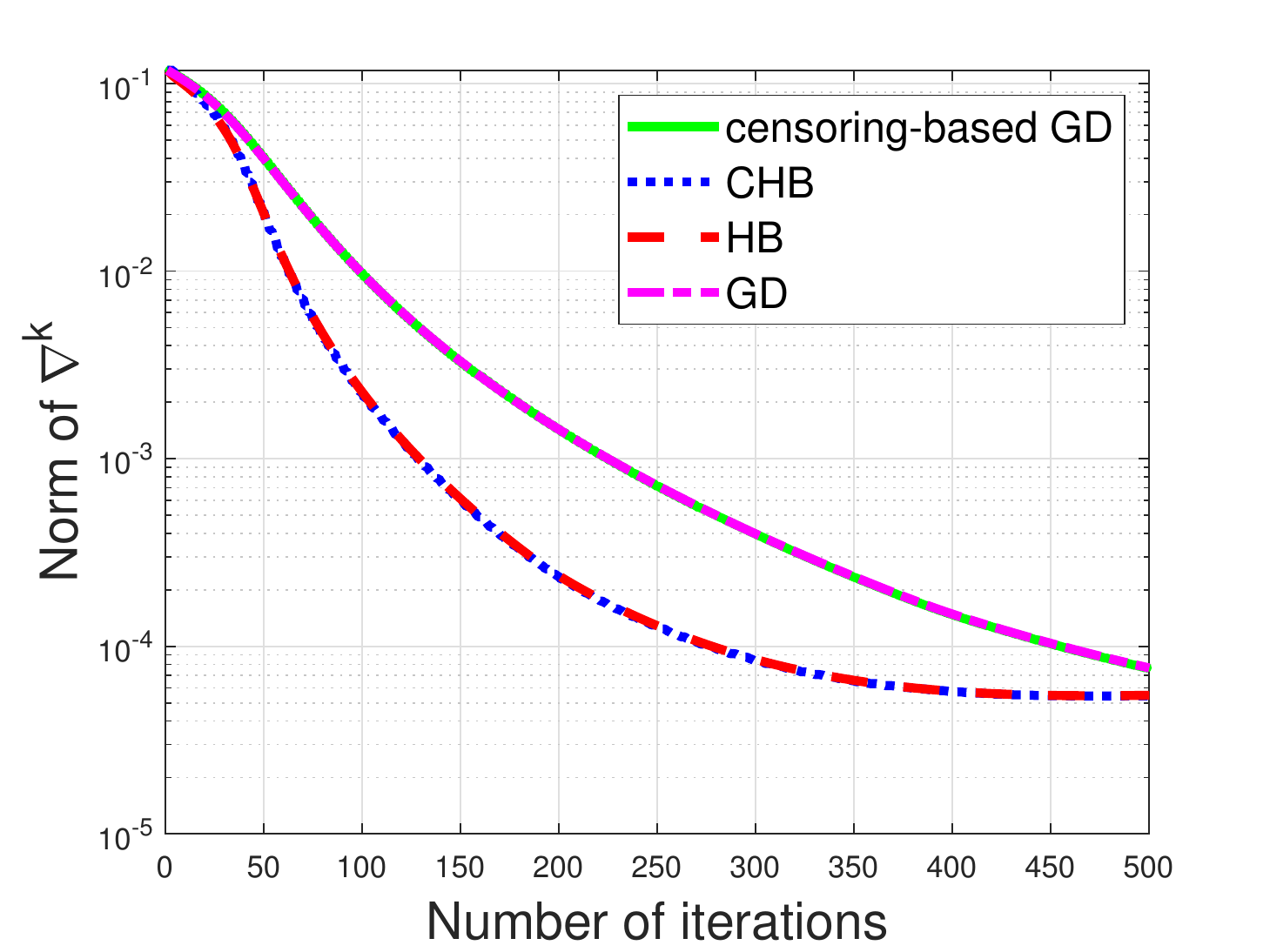}
  \caption*{(d) Neural network}\label{fig:NN_MNIST2}
\end{subfigure}
\caption{Objective error for lasso regression and training a neural network in the \emph{MNIST} dataset.}\label{fig:imageNNLassoMNIST}
\end{figure}

We next test the performance of CHB on a larger dataset \emph{MNIST} \cite{lecun1998gradient}. In all regression tasks, we set $\varepsilon_1=0.1/(\alpha^2M^2)$ for CHB and censoring-based GD. For HB and CHB, we set $\beta=0.4$. The largest number of iterations for each algorithm in each regression task is fixed to be $2000$. For linear regression and lasso regression, we set $\alpha=10^{-8}$. For logistic regression, we use $\alpha=10^{-6}$. The values of regularization parameter for lasso regression and logistic regression are $\lambda=0.5$ and $\lambda=0.001$, respectively. In training a neural network with the regularization parameter $\lambda=1/60000$, we choose $\alpha=0.02$ for all algorithms. For CHB and censoring-based GD, we choose $\varepsilon_1=0.01$. For HB and CHB, we choose the constant $\beta=0.4$. The largest number of iterations is set to be $500$ in training a neural network. In Figure \ref{fig:MNIST_Linear_Logistic} and Figure \ref{fig:imageNNLassoMNIST}, for the fixed total number of iterations, CHB shows a competitive communication saving performance while maintaining a relatively smaller objective error compared to GD and censoring-based GD. Specific results are summarized in Table \ref{GD-tableMNIST}.

\begin{figure}[!htb]
\centering
\begin{subfigure}[b]{.25\textwidth}
  \centering
  \includegraphics[width=\linewidth]{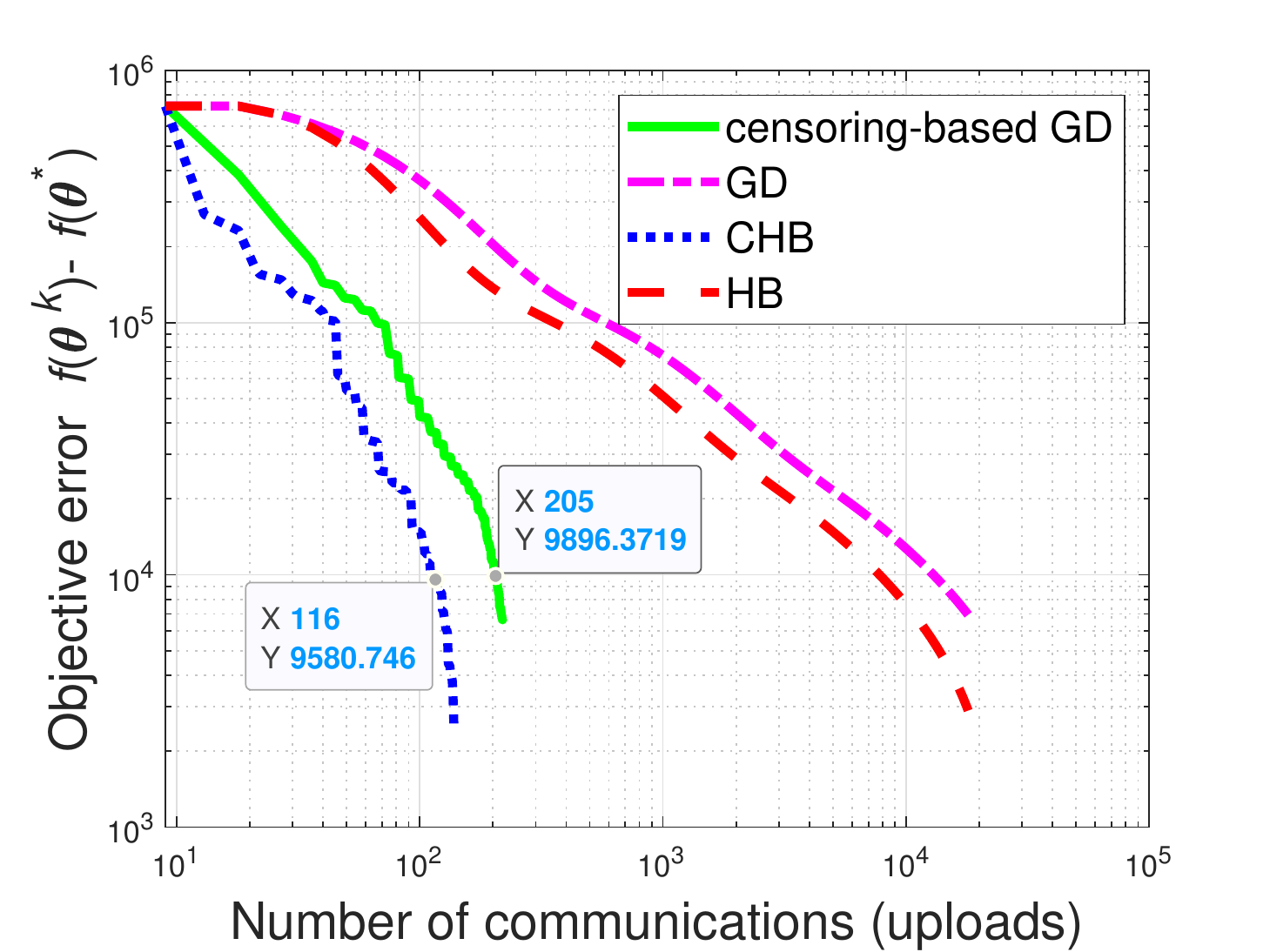}
  \caption*{(a) $\alpha=2.2\times10^{-8}$}\label{fig:stepsize228}
\end{subfigure}%
\begin{subfigure}[b]{.25\textwidth}
  \centering
  \includegraphics[width=\linewidth]{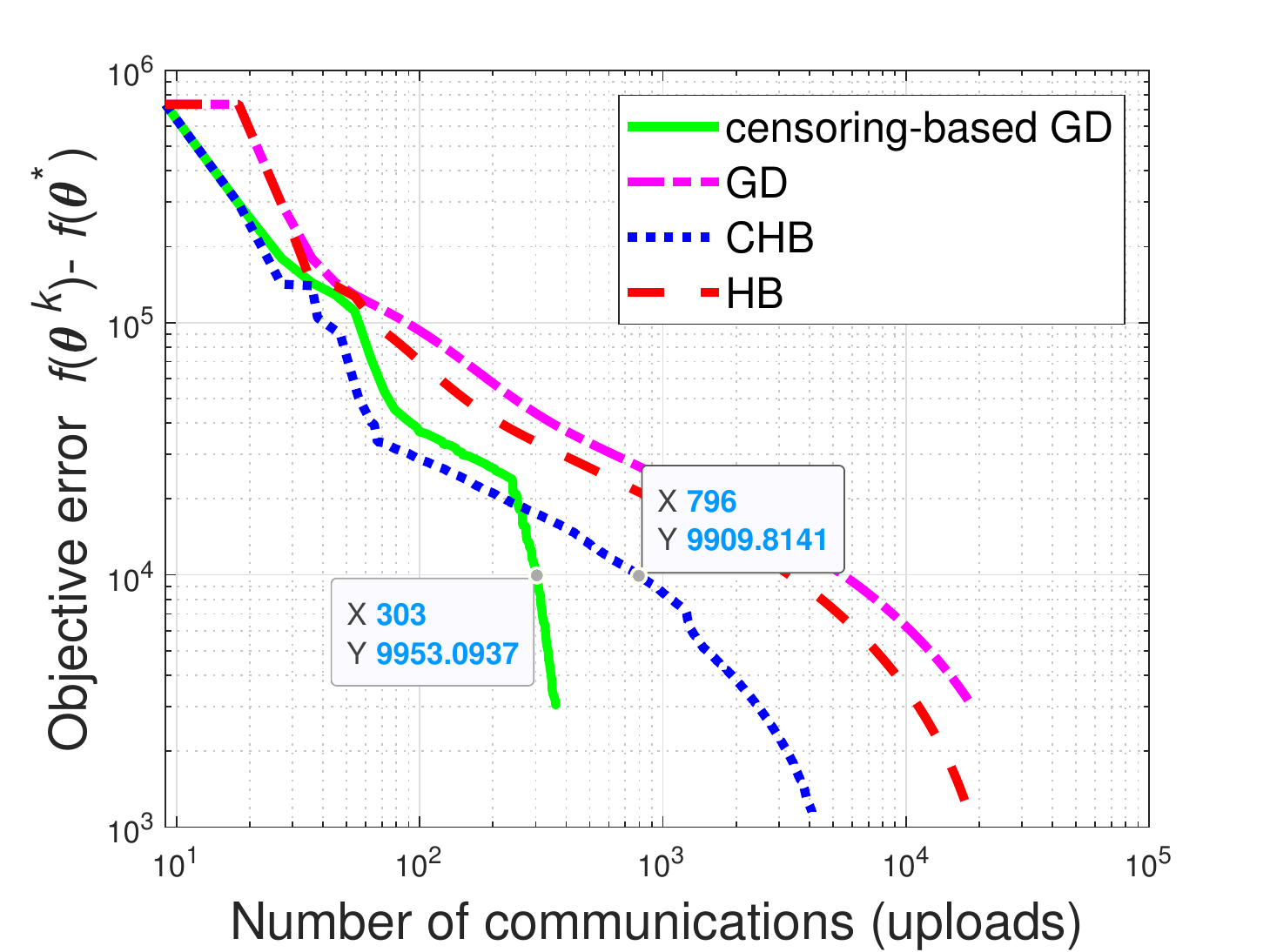}
  \caption*{(b) $\alpha=2.2\times10^{-7}$}\label{fig:stepsize227}
\end{subfigure}
\begin{subfigure}[b]{.25\textwidth}
  \centering
  \includegraphics[width=\linewidth]{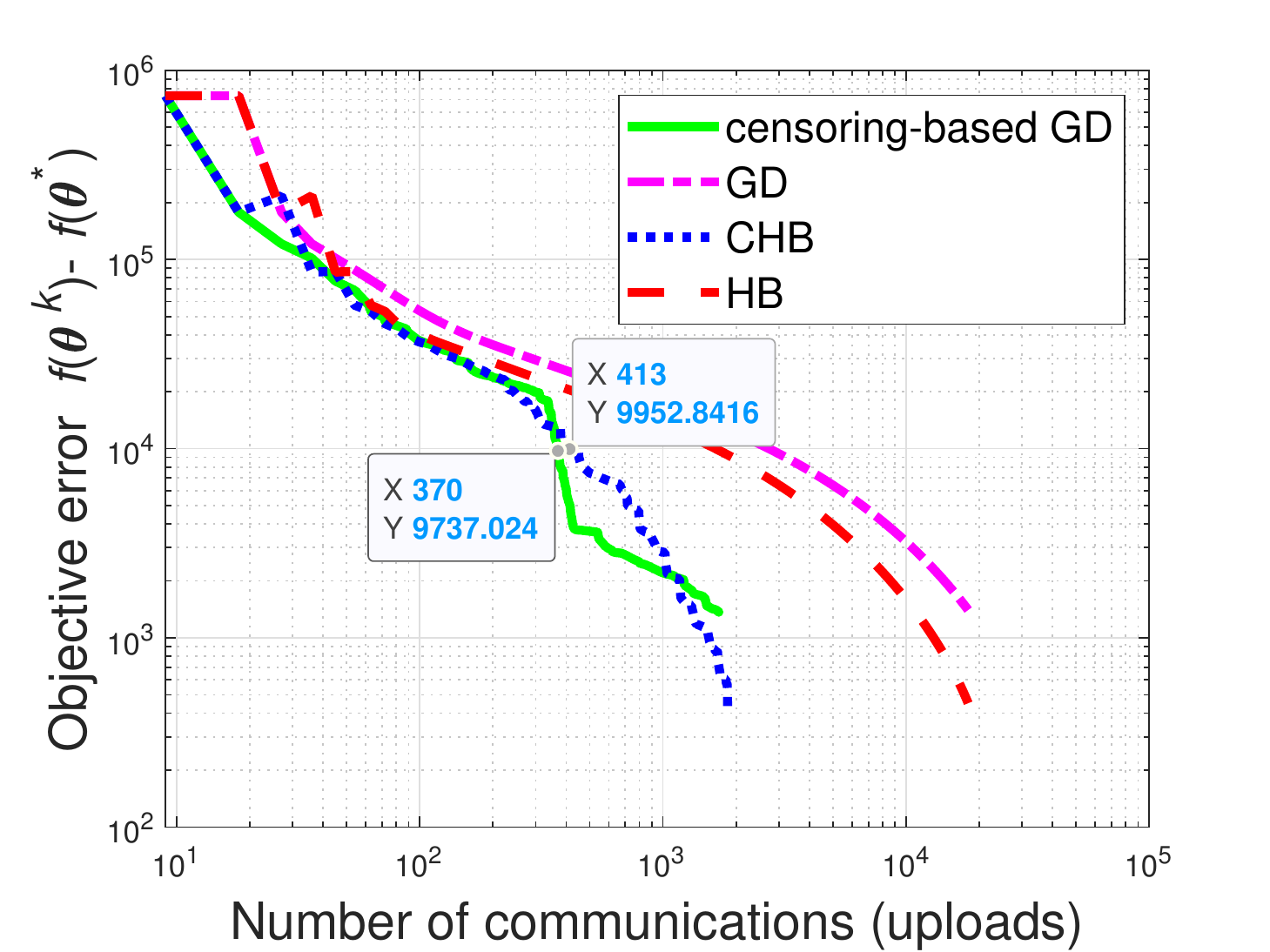}
  \caption*{(c) $\alpha=5.5\times10^{-7}$}\label{fig:stepsize557}
\end{subfigure}%
\begin{subfigure}[b]{.25\textwidth}
  \centering
  \includegraphics[width=\linewidth]{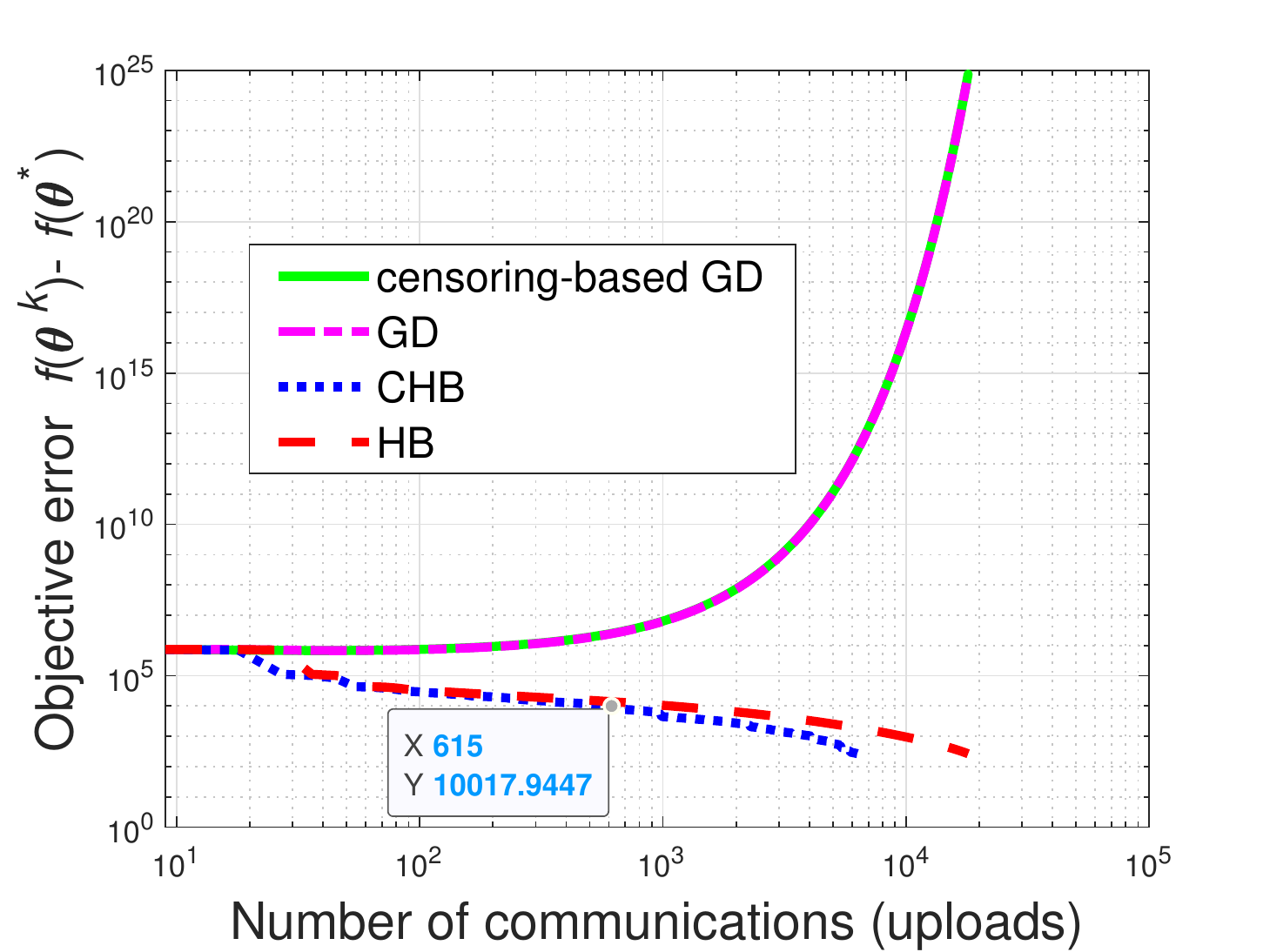}
  \caption*{(d) $\alpha=8.8\times10^{-7}$}\label{fig:stepsize887}
\end{subfigure}
\caption{Objective error for linear regression with different step sizes in the \emph{MNIST} dataset.}\label{fig:lineardiffstep}
\end{figure}

Furthermore, with the same parameter setting in the linear regression task on the MNIST dataset except for the step size, we consider the impact of the step size $\alpha$ on the number of communications needed to attain a target objective error. Figure \ref{fig:lineardiffstep}(a) and Figure \ref{fig:lineardiffstep}(b) indicates that choosing a large step size might speed up the optimization process (compare $\alpha=2.2\times10^{-7}$ and $\alpha=2.2\times10^{-8}$ at iteration $2000$ which corresponds to the end point of each curve), but choosing a large step size might not save the largest number of communications for CHB and censoring-based GD, e.g., to attain the fixed objective error $\approx10^{4}$,  the number of communications of censoring-based GD for $\alpha=2.2\times10^{-7}$ is $303$ which is larger than that for $\alpha=2.2\times10^{-8}$ with its number of communications being $205$.
Note that to attain the objective error $\approx10^4$, even through censoring-based GD can save more communications than CHB for the case when $\alpha=2.2\times10^{-7}$, this step size is not a good choice in term of communication savings. It is clear that censoring-based GD can save more communications if we  choose $\alpha=2.2\times10^{-8}$ and in this case CHB still outperforms censoring-based GD.
In order to reduce the total number of communications, a relatively small step size is preferable for CHB and censoring-based GD. This illustrates the basic trade-off between the number of iterations and the number of communication in some cases, which is interesting. Communications can be saved using a relatively small step size which results in more iterations to attain a target accuracy.
Additionally, Figure \ref{fig:lineardiffstep}(d) illustrates that compared to GD and censoring-based GD, the momentum term can help CHB converge for a relatively large step size case.

\begin{figure}[!htb]
\centering
\begin{subfigure}[b]{.25\textwidth}
  \centering
  \includegraphics[width=\linewidth]{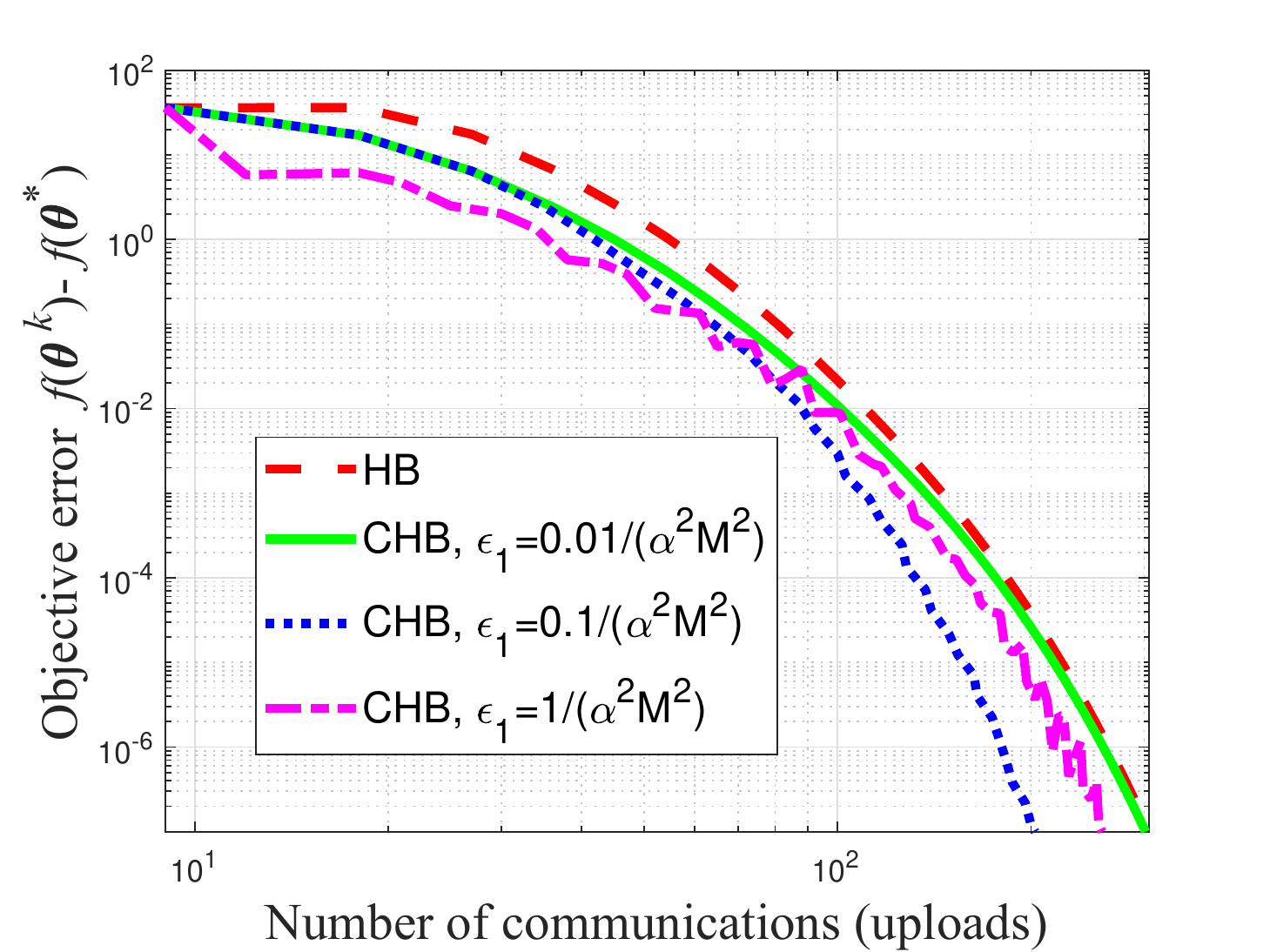}
\end{subfigure}%
\begin{subfigure}[b]{.25\textwidth}
  \centering
  \includegraphics[width=\linewidth]{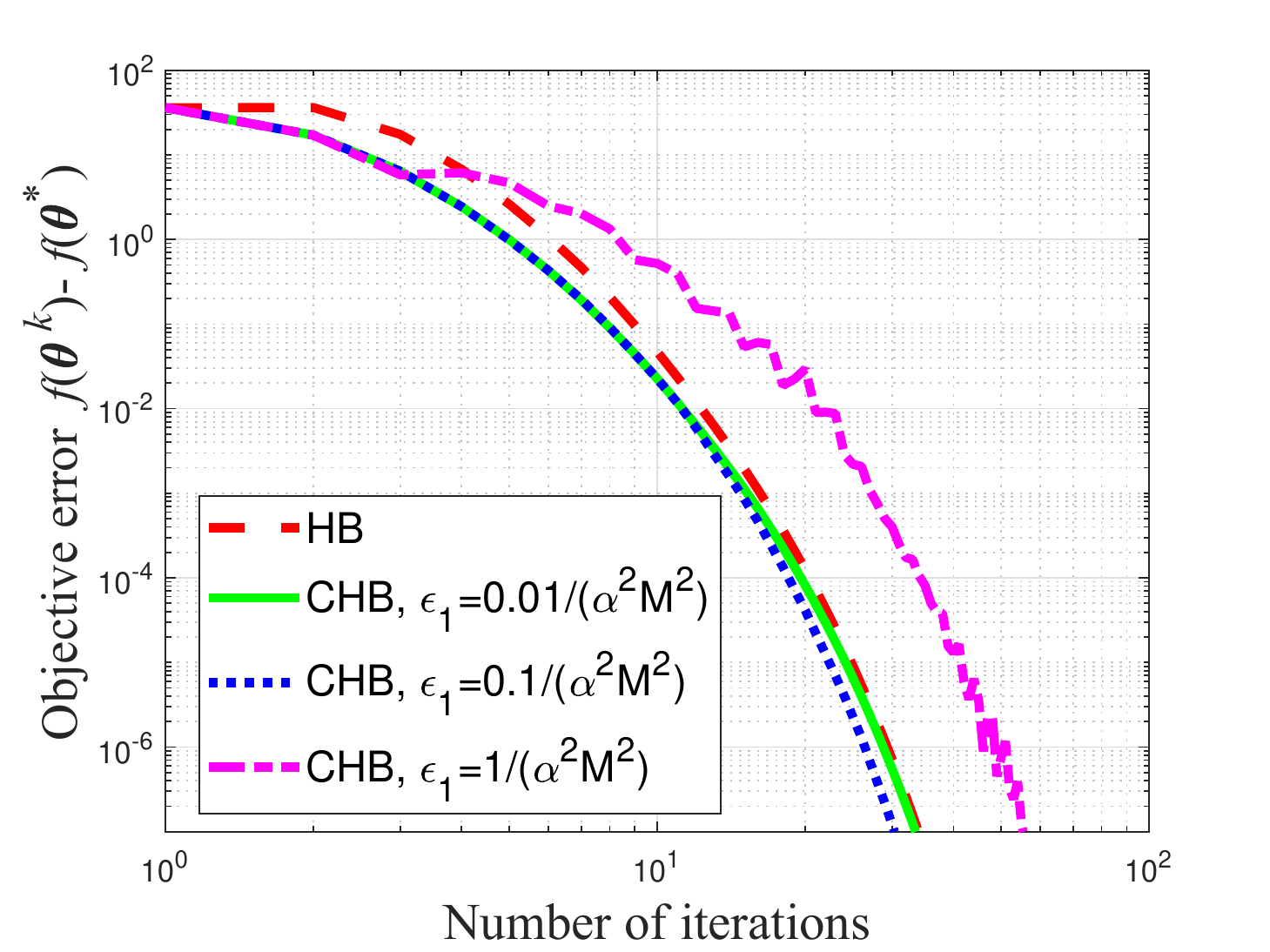}
\end{subfigure}
\caption{{Objective error versus the number of communications and iterations for logistic regression with different $\varepsilon_1$ in synthetic datasets.}}\label{fig:diffThres}
\end{figure}

{Now we consider the impact of $\varepsilon_1$ on the communication savings and convergence of CHB. Specifically, with the same setting as Figure \ref{fig:imageLCall} except $\varepsilon_1$, Figure \ref{fig:diffThres} indicates that CHB with $\varepsilon_1=0.01/(\alpha^2M^2)$ performs similarly to  the classical HB in terms of the number of iterations while saving some communications for a given objective error. This is reasonable since smaller $\varepsilon_1$ indicates less communication censoring. When we increase $\varepsilon_1$ to $0.1/(\alpha^2M^2)$, we can save more communications without too much impact on convergence. However, when $\varepsilon_1$ is too large, e.g., $\varepsilon_1=1/(\alpha^2M^2)$, increased communication censoring will occur such that more iterations are required to attain a given objective error.  Figure \ref{fig:diffThres} also indicates that CHB is able to achieve a good communication-iteration trade-off by tuning $\varepsilon_1$.  }
 
\begin{figure}[!htb]
\centering
\includegraphics[width=\linewidth]{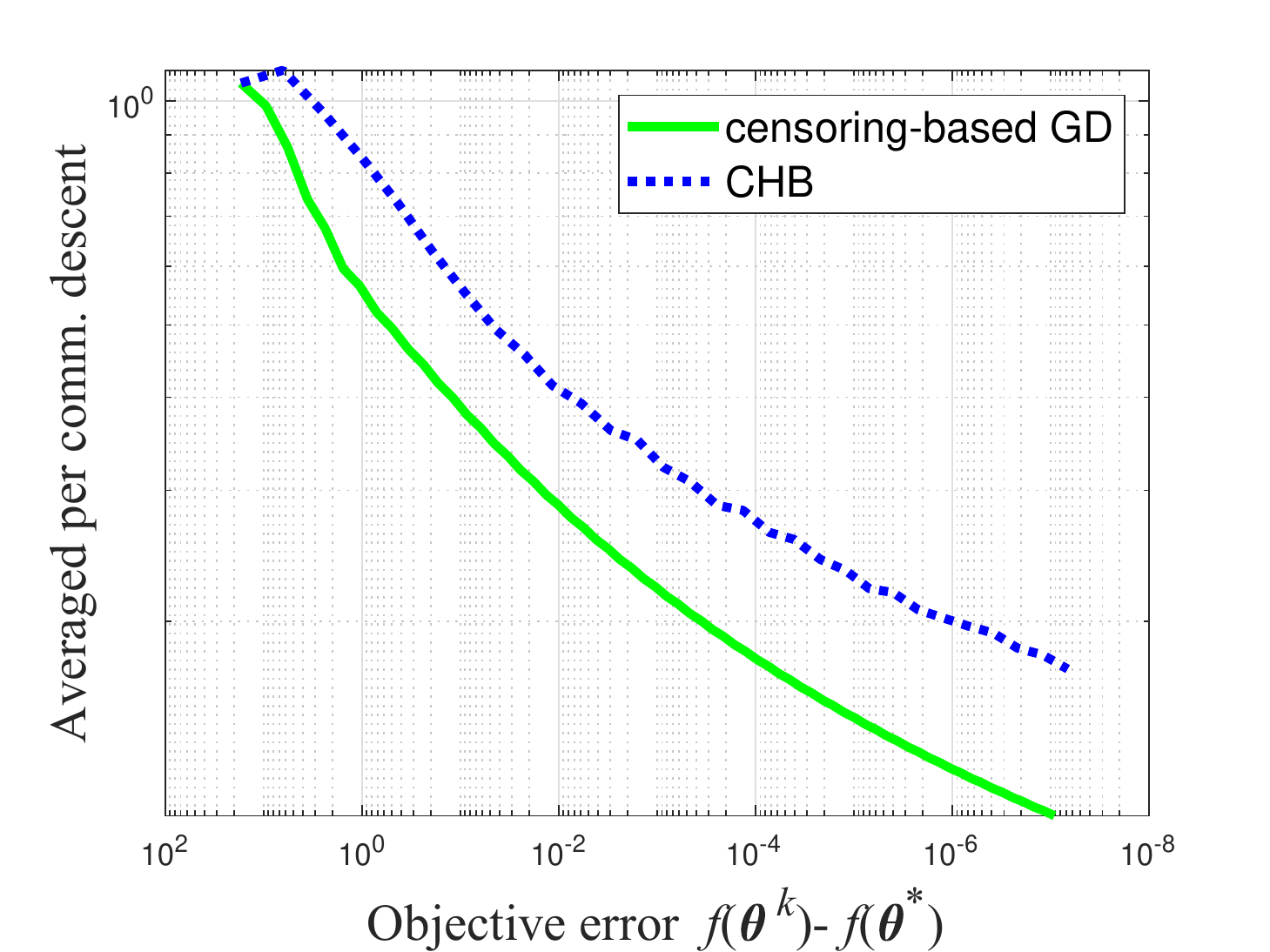}
\caption{{The averaged per-communication descent versus objective error for logistic regression with exactly the same setting as Figure \ref{fig:imageLCall}.}}
\label{ExplainMomentumTerm}
\end{figure}

{From the previous numerical results where we plot the objective error versus the number of iterations for CHB and censoring-based GD, e.g., Figure \ref{fig:imageLCall}, we can see that CHB provides faster convergence in these cases compared to censoring-based GD due to the momentum term in the update rule.  The averaged per-communication descent is defined by $(f(\boldsymbol{\theta}^{0})-f(\boldsymbol{\theta}^{k}))/(\text{the total number of communications})$. Figure \ref{ExplainMomentumTerm} shows that CHB has a larger averaged per-communication descent than  censoring-based GD for a given objective error in this case. The results in Figure \ref{ExplainMomentumTerm} also indicate that as the objective error becomes smaller (iteration $k$ becomes larger), the averaged per-communication descent for both CHB and censoring-based GD becomes smaller for both algorithms. This is reasonable since the descent typically becomes smaller for larger $k$ for both algorithms.   }

\begin{table*}[htbp]
  \caption{Performance comparison in the \emph{MNIST} dataset at the accomplishment of the largest number of iterations.}
  \label{GD-tableMNIST}
  \centering
  \tabcolsep=0.1cm
  \begin{tabular}{lllllllll}
    \toprule
    &\multicolumn{2}{c}{Linear regression}&\multicolumn{2}{c}{Lasso regression}&\multicolumn{2}{c}{Logistic regression} & \multicolumn{2}{c}{Neural network}                \\
    \cmidrule(r){2-3}\cmidrule(r){4-5}\cmidrule(r){6-7}\cmidrule(r){8-9}
    Name     & Comm.      & Obj. error   & Comm.   & Obj. error   & Comm.   & Obj. error  & Comm.    & Norm square grad.\\
    \midrule
    CHB & $\textbf{191}$  & $\mathbf{3.1077\times 10^{3}}$ & $\textbf{191}$ & $\mathbf{3.0901\times 10^{3}}$ & $\textbf{45}$ & $\textbf{0.5621}$ & $\textbf{1123}$ & $\mathbf{5.4545\times 10^{-5}}$ \\
    HB  & $18000$  & $3.6170\times 10^{3}$ & ${18000}$ & ${3.5995\times 10^{3}}$ & $18000$ & $4.0102$ & $4500$ & $5.4850\times10^{-5}$   \\
    LAG & $179$  & $8.6826\times 10^{3}$ & ${184}$ & ${8.6483\times 10^{3}}$ & $90$ & $12.6540$ & $1532$ & $7.5920\times10^{-5}$ \\
    GD  & $18000$  & $9.1365\times 10^{3}$ & ${18000}$ & ${9.1007\times 10^{3}}$ & $18000$ & $16.1719$ & $4500$ & $7.6681\times10^{-5}$ \\
    \bottomrule
  \end{tabular}
\end{table*}

\section{Conclusion}
This paper proposed a censoring-based heavy ball (CHB) method to optimize an objective function which is a sum of local functions in a communication-efficient manner. The key to save communications is to employ a censoring strategy where a worker is not allowed to transmit if its gradient is not sufficiently different from its previously transmitted one.  
{The HB method provides smoothing and so it is well matched with gradient censoring. }
Compared to the classical HB method, CHB is able to significantly reduce the number of communications while maintaining a fast convergence rate for strongly convex, convex, and nonconvex objective functions. 
{As we have shown, CHB can provide significant communications savings with very little impact on convergence rate, and CHB can also be tuned to trade-off communications and convergence rate if desired. }
In all cases numerical results validate the communication efficiency of CHB on both synthetic and real datasets. 
{The CHB approach provides a new tool for reducing communications, and so it can potentially be applied along with other complementary techniques such as quantization, compression, and gradient sparsification,}
to make CHB more efficient in terms of bandwidth per communication as well as the number of communications. {Additionally, finding an optimal approach to tune the parameters of CHB, e.g., $\varepsilon_1$, is also very interesting}. Finally, we note that worker privacy and security are important considerations, and this should be combined with CHB.

\appendices
\section{Proof of a Useful Lemma}
\begin{lemma}\label{CHBdescentlemma}
Suppose $f(\boldsymbol{\theta})$ is $L$-smooth and let ${\boldsymbol{\theta}}^{k+1}$ be generated by (\ref{commuinicationHBupdate}), then the CHB algorithm yields the following descent
\begin{align}\label{CHBupdateparanew1}
&f(\boldsymbol{\theta}^{k+1}) - f(\boldsymbol{\theta}^{k})\le -\frac{\alpha}{2}\left\|\nabla f\left(\boldsymbol{\theta}^{k}\right)\right\|^{2}-\frac{1-\alpha L}{2\alpha}\left\|\boldsymbol{\theta}^{k+1}-\boldsymbol{\theta}^{k}\right\|^{2}\notag\\ &\qquad+ \Big\| \sqrt{\frac{\alpha}{2}}\sum_{m\in\mathcal{M}_c^{k}} \delta \nabla_{m}^{k} + \frac{\beta}{\sqrt{2\alpha}}(\boldsymbol{\theta}^{k}-\boldsymbol{\theta}^{k-1}) \Big\|^{2}.
\end{align}
\begin{IEEEproof}
\textcolor[rgb]{0.00,0.00,0.00}{From  Assumption \ref{OHBassumption1} we know \cite{nesterov2018lectures}}
\begin{align}\label{smoothproperty}
&f\left(\boldsymbol{\theta}^{k+1}\right)-f\left(\boldsymbol{\theta}^{k}\right) \notag\\
&\leq\left\langle\nabla f\left(\boldsymbol{\theta}^{k}\right), \boldsymbol{\theta}^{k+1}-\boldsymbol{\theta}^{k}\right\rangle+\frac{L}{2}\left\|\boldsymbol{\theta}^{k+1}-\boldsymbol{\theta}^{k}\right\|^{2}.
\end{align}
\textcolor[rgb]{0.00,0.00,0.00}{Using (\ref{CHBupdatestep0}) in  $\left\langle\nabla f\left(\boldsymbol{\theta}^{k}\right), \boldsymbol{\theta}^{k+1}-\boldsymbol{\theta}^{k}\right\rangle$ yields}
\begin{align}
&\left\langle\nabla f\left(\boldsymbol{\theta}^{k}\right), \boldsymbol{\theta}^{k+1}-\boldsymbol{\theta}^{k}\right\rangle\notag\\
&=-\alpha\Big\langle \nabla f\left(\boldsymbol{\theta}^{k}\right), \nabla f\left(\boldsymbol{\theta}^{k}\right)- \sum_{m\in\mathcal{M}_c^{k}} \delta \nabla_{m}^{k}\Big\rangle \notag\\
&\qquad\qquad+ \beta\left\langle \nabla f\left(\boldsymbol{\theta}^{k}\right), \boldsymbol{\theta}^{k}-\boldsymbol{\theta}^{k-1}\right\rangle\\
&=-\alpha\left\|\nabla f(\boldsymbol{\theta}^{k})\right\|^2 + \Big\langle \sqrt{\alpha}\nabla f(\boldsymbol{\theta}^{k}),\sqrt{\alpha} \sum_{m\in\mathcal{M}_c^{k}} \delta \nabla_{m}^{k} \Big\rangle \notag\\
&\qquad\qquad+ \beta\left\langle \nabla f\left(\boldsymbol{\theta}^{k}\right), \boldsymbol{\theta}^{k}-\boldsymbol{\theta}^{k-1}\right\rangle\label{innerprodemix0}\\
&=-\alpha\left\|\nabla f(\boldsymbol{\theta}^{k})\right\|^2 + \frac{\alpha}{2}\left\|\nabla f(\boldsymbol{\theta}^{k})\right\|^2 + \frac{\alpha}{2}\Big\| \sum_{m\in\mathcal{M}_c^{k}} \delta \nabla_{m}^{k} \Big\|^2\notag\\
&\quad-\frac{\alpha}{2}\Big\| \nabla f(\boldsymbol{\theta}^{k})-\!\! \sum_{m\in\mathcal{M}_c^{k}} \delta \nabla_{m}^{k} \Big\|^2+ \beta\big\langle \nabla f\left(\boldsymbol{\theta}^{k}\right), \boldsymbol{\theta}^{k}-\boldsymbol{\theta}^{k-1}\big\rangle.\label{innerprodemix}
\end{align}
\textcolor[rgb]{0.00,0.00,0.00}{To go from (\ref{innerprodemix0}) to (\ref{innerprodemix}), the quality $\left\langle \boldsymbol x, \boldsymbol y\right\rangle=\frac{1}{2}\|\boldsymbol x\|^2+\frac{1}{2}\|\boldsymbol y\|^2-\frac{1}{2}\|\boldsymbol{x-y}\|^2$ is employed for the term $\langle \sqrt{\alpha}\nabla f(\boldsymbol{\theta}^{k}),\sqrt{\alpha} \sum_{m\in\mathcal{M}_c^{k}} \delta \nabla_{m}^{k} \rangle$.}

From (\ref{CHBupdatestep0}), we obtain that
\begin{align}\label{decomposingsqure}
&\left\|\boldsymbol{\theta}^{k+1}-\boldsymbol{\theta}^{k}\right\|^{2} \notag\\
&= \alpha^2\Big\|\nabla f\left(\boldsymbol{\theta}^{k}\right) - \sum_{m\in\mathcal{M}_c^{k}} \delta \nabla_{m}^{k} \Big\|^{2} + \beta^2 \left\| \boldsymbol{\theta}^{k}-\boldsymbol{\theta}^{k-1}\right\|^{2}\notag\\
&\qquad-2\alpha\beta\Big\langle\nabla f\left(\boldsymbol{\theta}^{k}\right), \boldsymbol{\theta}^{k}-\boldsymbol{\theta}^{k-1}\Big\rangle \notag\\
&\qquad + 2\alpha\beta\Big\langle \sum_{m\in\mathcal{M}_c^{k}} \delta \nabla_{m}^{k}, \boldsymbol{\theta}^{k}-\boldsymbol{\theta}^{k-1}  \Big\rangle.
\end{align}

Plugging (\ref{innerprodemix}) and (\ref{decomposingsqure}) into (\ref{smoothproperty}), we obtain
\begin{align}\label{communicationHB11}
&f\left(\boldsymbol{\theta}^{k+1}\right)-f\left(\boldsymbol{\theta}^{k}\right)\notag\\
& \leq  - \frac{\alpha}{2}\left\|\nabla f\left(\boldsymbol{\theta}^{k}\right)\right\|^{2}
-\frac{\alpha(1-\alpha L)}{2}\Big\| \nabla f\left(\boldsymbol{\theta}^{k}\right) - \sum_{m\in\mathcal{M}_c^{k}} \delta \nabla_{m}^{k} \Big\|^{2}\notag\\
&\quad + \frac{\alpha}{2}\Big\| \sum_{m\in\mathcal{M}_c^{k}} \delta \nabla_{m}^{k} \Big\|^{2} + \frac{L\beta^2}{2}\left\|\boldsymbol{\theta}^{k}-\boldsymbol{\theta}^{k-1}\right\|^{2} \notag\\
&\quad+ \beta(1-\alpha L)\Big\langle\nabla f\left(\boldsymbol{\theta}^{k}\right), \boldsymbol{\theta}^{k}-\boldsymbol{\theta}^{k-1} \Big\rangle\notag\\
&\quad + \alpha\beta L\Big\langle \sum_{m\in\mathcal{M}_c^{k}} \delta \nabla_{m}^{k}, \boldsymbol{\theta}^{k}-\boldsymbol{\theta}^{k-1}\Big\rangle.
\end{align}

Multiplying both sides of (\ref{decomposingsqure}) by a constant $(1-\alpha L)/(2\alpha)$ implies
\begin{align}\label{CHBupdatepara}
&\frac{1-\alpha L}{2\alpha}\left\|\boldsymbol{\theta}^{k+1}-\boldsymbol{\theta}^{k}\right\|^{2} \notag\\
&= \frac{\alpha(1-\alpha L)}{2}\Big\|\nabla f\left(\boldsymbol{\theta}^{k}\right) \!\!- \!\!\!\! \sum_{m\in\mathcal{M}_c^{k}} \delta \nabla_{m}^{k}\Big\|^{2} \notag\\
&\quad\qquad+ \frac{(1-\alpha L)\beta^2}{2\alpha} \left\| \boldsymbol{\theta}^{k}-\boldsymbol{\theta}^{k-1}\right\|^{2}\notag\\
&\quad\qquad-\beta(1-\alpha L)\Big\langle\nabla f\left(\boldsymbol{\theta}^{k}\right), \boldsymbol{\theta}^{k}-\boldsymbol{\theta}^{k-1}\Big\rangle\notag\\ &\quad\qquad+ (1-\alpha L)\beta\Big\langle \sum_{m\in\mathcal{M}_c^{k}} \delta \nabla_{m}^{k}, \boldsymbol{\theta}^{k}-\boldsymbol{\theta}^{k-1}  \Big\rangle.
\end{align}
Adding (\ref{CHBupdatepara}) to (\ref{communicationHB11}) yields
\begin{align}
&f\left(\boldsymbol{\theta}^{k+1}\right)\! +\! \frac{1-\alpha L}{2\alpha}\left\|\boldsymbol{\theta}^{k+1}-\boldsymbol{\theta}^{k}\right\|^{2}\notag\\
& \leq  f\left(\boldsymbol{\theta}^{k}\right)- \frac{\alpha}{2}\left\|\nabla f\left(\boldsymbol{\theta}^{k}\right)\right\|^{2} + \frac{\beta^2}{2\alpha}\left\| \boldsymbol{\theta}^{k}-\boldsymbol{\theta}^{k-1}\right\|^{2} \notag\\
&\quad+ \frac{\alpha}{2}\Big\| \sum_{m\in\mathcal{M}_c^{k}} \delta \nabla_{m}^{k} \Big\|^{2}+\beta\Big\langle \sum_{m\in\mathcal{M}_c^{k}} \delta \nabla_{m}^{k}, \boldsymbol{\theta}^{k}-\boldsymbol{\theta}^{k-1}\Big\rangle\\
&\!\leq f\left(\boldsymbol{\theta}^{k}\right)- \!\frac{\alpha}{2}\left\|\nabla f\left(\boldsymbol{\theta}^{k}\right)\right\|^{2} \notag\\
&\quad+ \Big\| \sqrt{\frac{\alpha}{2}}\sum_{m\in\mathcal{M}_c^{k}}\delta \nabla_{m}^{k}+\frac{\beta}{\sqrt{2\alpha}}(\boldsymbol{\theta}^{k}-\boldsymbol{\theta}^{k-1}) \Big\|^{2}.\label{CHBupdateparanew}
\end{align}
The result in (\ref{CHBupdateparanew1}) follows after rearranging terms in (\ref{CHBupdateparanew}).
\end{IEEEproof}
\end{lemma}

\section{Proof of Lemma \ref{descentLyapunovLemma}}\label{descentProof}

\textcolor[rgb]{0.00,0.00,0.00}{With  $\mathbb{L}(\boldsymbol{\theta}^{k})$ in (\ref{definationLyapunov}), it can be shown that}
\begin{align}\label{diffLyapunov}
&\mathbb{L}(\boldsymbol{\theta}^{k+1})-\mathbb{L}(\boldsymbol{\theta}^{k})\notag\\
&= f(\boldsymbol{\theta}^{k+1})-f(\boldsymbol{\theta}^{k})+\eta_1\left\|\boldsymbol{\theta}^{k+1}
-\boldsymbol{\theta}^{k}\right\|^{2} - \eta_1\left\|\boldsymbol{\theta}^{k}
-\boldsymbol{\theta}^{k-1}\right\|^{2}.
\end{align}

Plugging (\ref{CHBupdateparanew1}) into (\ref{diffLyapunov}) yields
\begin{align}
&\mathbb{L}(\boldsymbol{\theta}^{k+1})-\mathbb{L}(\boldsymbol{\theta}^{k})\notag\\
&\le -\frac{\alpha}{2}\left\|\nabla f\left(\boldsymbol{\theta}^{k}\right)\right\|^{2}+\left(\eta_1-\frac{1-\alpha L}{2\alpha}\right)\left\|\boldsymbol{\theta}^{k+1}-\boldsymbol{\theta}^{k}\right\|^{2} \notag\\
&\qquad+ \Big\| \sqrt{\frac{\alpha}{2}}\sum_{m\in\mathcal{M}_c^{k}} \delta \nabla_{m}^{k} + \frac{\beta}{\sqrt{2\alpha}}(\boldsymbol{\theta}^{k}-\boldsymbol{\theta}^{k-1}) \Big\|^{2}\notag\\
&\qquad-\eta_1 \left\| \boldsymbol{\theta}^{k}
-\boldsymbol{\theta}^{k-1} \right\|^{2}\label{simplifydiffLyapunov1}\\
&\le -\frac{\alpha}{2}\left\|\nabla f\left(\boldsymbol{\theta}^{k}\right)\right\|^{2}+\left(\eta_1-\frac{1-\alpha L}{2\alpha}\right)\left\|\boldsymbol{\theta}^{k+1}-\boldsymbol{\theta}^{k}\right\|^{2} \notag\\
&\qquad+ \frac{\alpha}{2}(1+\rho_3)\Big\| \sum_{m\in\mathcal{M}_c^{k}} \delta \nabla_{m}^{k} \Big\|^{2}\notag\\
&\qquad+ \frac{\beta^2}{{2\alpha}}(1+\rho^{-1}_3)\left\| \boldsymbol{\theta}^{k}-\boldsymbol{\theta}^{k-1} \right\|^{2}
-\eta_1 \left\| \boldsymbol{\theta}^{k}
-\boldsymbol{\theta}^{k-1} \right\|^{2}\label{simplifydiffLyapunov2}
\end{align}
where $\rho_3$ is any positive number. In going from (\ref{simplifydiffLyapunov1}) to (\ref{simplifydiffLyapunov2}), we employ Young's inequality: $\|\boldsymbol{x+y}\|^2\le(1+\rho)\|\boldsymbol{x}\|^2+(1+\rho^{-1})\|\boldsymbol{y}\|^2, \forall\rho>0$. From (\ref{CHBupdatestep0}), we have
\begin{align}
&\left\|\boldsymbol{\theta}^{k+1}
-\boldsymbol{\theta}^{k}\right\|^{2}\notag\\
&=\Big\| -\alpha\Big(\nabla f(\boldsymbol{\theta}^{k}) - \!\! \sum_{m\in\mathcal{M}_c^{k}} \delta \nabla_{m}^{k}\Big)+\beta\Big( \boldsymbol{\theta}^{k}-\boldsymbol{\theta}^{k-1}\Big) \Big\|^{2}\\
&\le\alpha^2(1+\rho_1)\Bigg((1+\rho_2)\Big\| \nabla f(\boldsymbol{\theta}^{k})  \Big\|^{2}\notag\\
&\qquad\qquad +  (1+\rho_2^{-1})\Big\|\sum_{m\in\mathcal{M}_c^{k}} \delta \nabla_{m}^{k}\Big\|^2\Bigg) \notag\\
&\qquad+\beta^2(1+\rho_1^{-1})\left\| \boldsymbol{\theta}^{k}-\boldsymbol{\theta}^{k-1} \right\|^{2}\label{extenddiff}
\end{align}
where $\rho_1$ and $\rho_2$ are any positive numbers, and the inequality in (\ref{extenddiff}) is obtained using Young's inequality. \textcolor[rgb]{0.00,0.00,0.00}{Using  (\ref{extenddiff}) in  (\ref{simplifydiffLyapunov2}) along with $\eta_1-\frac{1-\alpha L}{2\alpha}\ge0$,} yields 
\begin{align}\label{diffLayfunc}
&\mathbb{L}(\boldsymbol{\theta}^{k+1})-\mathbb{L}(\boldsymbol{\theta}^{k})\notag\\
&\le \left(  -\frac{\alpha}{2} + \Big(\eta_1-\frac{1-\alpha L}{2\alpha}\Big)\alpha^2(1+\rho_1)(1+\rho_2)  \right)\left\|\nabla f\left(\boldsymbol{\theta}^{k}\right)\right\|^{2}\notag\\
&\quad + \Bigg( \Big(\eta_1-\frac{1-\alpha L}{2\alpha}\Big)\alpha^2(1+\rho_1)(1+\rho_2^{-1})  \notag\\ &\qquad\qquad+\frac{\alpha}{2}(1+\rho_3)  \Bigg)\left\| \sum_{m\in\mathcal{M}_c^{k}} \delta \nabla_{m}^{k} \right\|^{2}\notag\\
&\quad + \Bigg( \frac{\beta^2}{2\alpha}(1+\rho_3^{-1}) -\eta_1 \notag\\
&\qquad+ \Big(\eta_1-\frac{1-\alpha L}{2\alpha}\Big)\beta^2(1+\rho_1^{-1}) \Bigg)\left\| \boldsymbol{\theta}^{k}-\boldsymbol{\theta}^{k-1} \right\|^{2}.
\end{align}
Employing the inequality $\left\|\sum_{n=1}^N \boldsymbol x_n\right\|^{2}\le N\sum_{n=1}^N\left\|\boldsymbol x_n\right\|^{2}$, $\left\| \sum_{m\in\mathcal{M}_c^{k}} \delta \nabla_{m}^{k} \right\|^{2}$ can be bounded by
\begin{align}
\Big\| \sum_{m\in\mathcal{M}_c^{k}} \delta \nabla_{m}^{k} \Big\|^{2}&\le |\mathcal{M}_c^{k}|\sum_{m\in\mathcal{M}_c^{k}}\Big\|  \delta \nabla_{m}^{k} \Big\|^{2}\notag\\
&\le |\mathcal{M}_c^{k}|^2\varepsilon_1\|\boldsymbol{\theta}^{k}-\boldsymbol{\theta}^{k-1}\|^2 \label{boundoncensoring}
\end{align}
where (\ref{boundoncensoring}) is obtained using the CHB-skip-transmission condition in (\ref{CHB-stopping condition}).

Plugging (\ref{boundoncensoring}) into (\ref{diffLayfunc}) with $\gamma\buildrel \Delta \over= \frac{\alpha}{2}(1+\rho_3) +\Big(\eta_1-\frac{1-\alpha L}{2\alpha}\Big)\alpha^2(1+\rho_1)(1+\rho_2^{-1})$ and $\eta_1-\frac{1-\alpha L}{2\alpha}\ge0$, we obtain
\begin{align}\label{generaldiffLayfunc}
&\mathbb{L}(\boldsymbol{\theta}^{k+1})-\mathbb{L}(\boldsymbol{\theta}^{k})\notag\\
&\le \left(  -\frac{\alpha}{2} + \Big(\eta_1-\frac{1-\alpha L}{2\alpha}\Big)\alpha^2(1+\rho_1)(1+\rho_2)  \right)\left\|\nabla f\left(\boldsymbol{\theta}^{k}\right)\right\|^{2}\notag\\
& \qquad+ \Bigg( \Big(\eta_1-\frac{1-\alpha L}{2\alpha}\Big)\beta^2(1+\rho_1^{-1}) + |\mathcal{M}_c^{k}|^2\varepsilon_1 \gamma \notag\\
& \qquad+ \frac{\beta^2}{2\alpha}(1+\rho_3^{-1}) -\eta_1  \Bigg)\left\| \boldsymbol{\theta}^{k}-\boldsymbol{\theta}^{k-1} \right\|^{2}.
\end{align}
Note that $\eta_1-\frac{1-\alpha L}{2\alpha}\ge0$ implies $\gamma\ge\frac{\alpha}{2}(1+\rho_3)>0$.
The proof is complete after defining the non-negative constants $\sigma_0$ and $\sigma_1$ as shown in (\ref{DDScondition1}) and (\ref{condition2}), respectively. Since the expressions of $\sigma_0$ and $\sigma_1$ are complicated, we provide several choices of parameters in the following.

\textcolor[rgb]{0.00,0.00,0.00}{\paragraph{ Parameter Choices} First note that  (\ref{DDScondition1})--(\ref{etacondition5}) are equivalent to}
\begin{align}
&\frac{1}{2\eta_1+L}\le\alpha\le\frac{1+(1+\rho_1)(1+\rho_2)}{(2\eta_1+L)(1+\rho_1)(1+\rho_2)}, \label{equalconditionsdic1}\\
& \beta\le \sqrt{\frac{2\alpha\eta_1}{(1+\rho_3^{-1})+(1+\rho_1^{-1})(2\alpha\eta_1-1+\alpha L)}}\label{equalconditionsdic2},\\
&\varepsilon_1\le\frac{\eta_1-\beta^2\big(\frac{1+\rho_3^{-1}}{2\alpha} + (\eta_1-\frac{1-\alpha L}{2\alpha})(1+\rho_1^{-1})\big)}{|\mathcal{M}_c^{k}|^2\gamma}\label{equalconditionsdic3}.
\end{align}
$\bullet$ \textcolor[rgb]{0.00,0.00,0.00}{Setting $\eta_1=\frac{1-\alpha L}{2\alpha}$, (\ref{equalconditionsdic1})--(\ref{equalconditionsdic3}) are equivalent to}
\begin{align}
\alpha&\le\frac{1}{L}, \quad \beta\le\sqrt{\frac{1-\alpha L}{1+\rho_3^{-1}}},\quad \varepsilon_1\le\frac{(1-\alpha L)-\beta^2(1+\rho_3^{-1})}{\alpha^2(1+\rho_3)|\mathcal{M}_c^{k}|^2}.
\end{align}

$\bullet$ \textcolor[rgb]{0.00,0.00,0.00}{Imposing $\eta_1 = 0$, $\rho_1 = 0$, and $\rho_2 = 0$ shows that  (\ref{equalconditionsdic1})--(\ref{equalconditionsdic3}) degenerate to $1/L\le\alpha\le 2/L$, $\beta=0$, and $\varepsilon_1=0$.}

$\bullet$ \textcolor[rgb]{0.00,0.00,0.00}{For $\alpha=1/L$ and $\eta_1>0$, we find  (\ref{equalconditionsdic1})--(\ref{equalconditionsdic3}) are equivalent to}
\begin{align}
0&\le\eta_1\le\frac{L}{2(1+\rho_1)(1+\rho_2)},\notag\\
\beta^2&\le\frac{2\eta_1}{2\eta_1(1+\rho_1^{-1})+L(1+\rho_3^{-1})},\notag\\
\varepsilon_1&\le\frac{L^2\Big(2\eta_1-\beta^2\big({L}(1+\rho_3^{-1})+2\eta_1(1+\rho_1^{-1})\big)\Big)}{|\mathcal{M}_c^{k}|^2\Big(L(1+\rho_3)+2\eta_1(1+\rho_1)(1+\rho_2^{-1})\Big)}.
\end{align}

\section{Proof of Theorem \ref{convergenceTheorem}}\label{theorem1Proof}
\textcolor[rgb]{0.00,0.00,0.00}{From strong convexity from \textcolor[rgb]{0.00,0.00,0.00}{Definition \ref{OHBassumption2}}, $f(\boldsymbol{\theta})$ satisfies \cite{nesterov2018lectures}}
\begin{align}\label{stronglyconvexpro}
2\mu\left(f\left(\boldsymbol{\theta}^{k}\right)-f\left(\boldsymbol{\theta}^{*}\right)\right) \leq\left\|\nabla f\left(\boldsymbol{\theta}^{k}\right)\right\|^{2}
\end{align}
\textcolor[rgb]{0.00,0.00,0.00}{where $\boldsymbol{\theta}^{*}$ denotes the minimizer of (\ref{basicprob}). Using (\ref{stronglyconvexpro}) in  (\ref{generaldiffLayfunc}) with $\sigma_0\ge0$ in (\ref{DDScondition1}) and $\eta_1\ne0$ yields }
\begin{align}\label{}
&\mathbb{L}(\boldsymbol{\theta}^{k+1})-\mathbb{L}(\boldsymbol{\theta}^{k})\notag\\
&\le \Bigg(   2\mu \Big(\eta_1-\frac{1-\alpha L}{2\alpha}\Big)\alpha^2(1+\rho_1)(1+\rho_2)  \notag\\
&\qquad\qquad-\alpha\mu\Bigg)\left(f(\boldsymbol{\theta}^{k})-f(\boldsymbol{\theta}^{*})\right)\notag\\
& \qquad+ \Bigg( \frac{\beta^2}{2\alpha}(1+\rho_3^{-1}) -\eta_1 + \Big(\eta_1-\frac{1-\alpha L}{2\alpha}\Big)\beta^2(1+\rho_1^{-1})\notag\\
&\qquad\qquad + |\mathcal{M}_c^{k}|^2\varepsilon_1 \gamma \Bigg)\left\| \boldsymbol{\theta}^{k}-\boldsymbol{\theta}^{k-1} \right\|^{2}\\
&=-\Bigg( - \Big(2\eta_1-\frac{1-\alpha L}{\alpha}\Big)\alpha^2\mu(1+\rho_1)(1+\rho_2)  \notag\\
&\qquad\qquad+\alpha\mu\Bigg)\left(f(\boldsymbol{\theta}^{k})-f(\boldsymbol{\theta}^{*})\right)\notag\\
&\qquad-\Bigg(1-  \frac{\beta^2}{2\alpha\eta_1}(1+\rho_3^{-1}) - \Big(1-\frac{1-\alpha L}{2\alpha\eta_1}\Big)\beta^2(1+\rho_1^{-1}) \notag\\
&\qquad\qquad-\frac{|\mathcal{M}_c^{k}|^2\varepsilon_1 \gamma}{\eta_1} \Bigg)\eta_1\left\| \boldsymbol{\theta}^{k}-\boldsymbol{\theta}^{k-1} \right\|^{2} \label{blumadd}
\end{align}
{Define 
\begin{align}
&c\left(\alpha,\beta,\varepsilon_1\right)\notag\\
&\buildrel \Delta \over=\min\Bigg\{ {\alpha}\mu - \Big(2\eta_1-\frac{1-\alpha L}{\alpha}\Big)\alpha^2\mu(1+\rho_1)(1+\rho_2), \notag\\
&\qquad \min_k\big\{1-\frac{\beta^2}{2\alpha\eta_1}(1+\rho_3^{-1})  -  \Big(1-\frac{1-\alpha L}{2\alpha\eta_1}\Big)\beta^2(1+\rho_1^{-1}) \notag\\
&\qquad\qquad\qquad-  \frac{|\mathcal{M}_c^{k}|^2\varepsilon_1 \gamma}{\eta_1}\big\}\Bigg\}\label{cgammaeta}\\
& = \min\big\{ 2\sigma_0\mu,\min_k\{\sigma_1\}/\eta_1 \big\}.\label{simplifycgammaeta}
\end{align}}
\textcolor[rgb]{0.00,0.00,0.00}{The result in (\ref{simplifycgammaeta}) follows from  the expressions for $\sigma_0$ and $\sigma_1$  in (\ref{DDScondition1}) and (\ref{condition2}), respectively.
Using (\ref{blumadd}) and  (\ref{cgammaeta}) yields }
\begin{align}\label{}
&\mathbb{L}(\boldsymbol{\theta}^{k+1})-\mathbb{L}(\boldsymbol{\theta}^{k})\notag\\
&\le - c\left(\alpha,\beta,\varepsilon_1\right)\left(f(\boldsymbol{\theta}^{k})-f\left(\boldsymbol{\theta}^{*}
\right) +\eta_{1}\left\|\boldsymbol{\theta}^{k}-\boldsymbol{\theta}^{k-1}\right\|^{2}\right)\\
&= - c\left(\alpha,\beta,\varepsilon_1\right)\mathbb{L}(\boldsymbol{\theta}^{k})
\end{align}
which implies
\begin{align}\label{Qlinearconverge}
\mathbb{L}(\boldsymbol{\theta}^{k+1})\le\left(1-c\left(\alpha,\beta,\varepsilon_1\right)\right)\mathbb{L}(\boldsymbol{\theta}^{k}).
\end{align}
It is clear that when (\ref{DDScondition1})--(\ref{condition2}) are satisfied with $\sigma_0>0$ in (\ref{DDScondition1}) and $\sigma_1>0$ in (\ref{condition2}), we have $c\left(\alpha,\beta,\varepsilon_1\right)>0$. If (\ref{etacondition5}) is satisfied, then we can obtain $c\left(\alpha,\beta,\varepsilon_1\right)<1$ from (\ref{cgammaeta}). Thus, we have $c\left(\alpha,\beta,\varepsilon_1\right)\in(0,1)$. The result in (\ref{Qlinearconverge}) indicates the Q-linear convergence of $\mathbb{L}(\boldsymbol{\theta}^{k})$. Since $\eta_1$ is non-negative in the definition of $\mathbb{L}(\boldsymbol{\theta}^{k})$, the result in (\ref{Qlinearconverge}) implies that
\begin{align}\label{RlinearConvergence}
f\left(\boldsymbol{\theta}^{k}\right)-f\left(\boldsymbol{\theta}^{*}\right)\le\left(1-c\left(\alpha,\beta,\varepsilon_1\right)\right)^k\mathbb{L}(\boldsymbol{\theta}^{0}), 
\end{align}
\textcolor[rgb]{0.00,0.00,0.00}{which  completes the proof. Consider a case with $\eta_1-\frac{1-\alpha L}{2\alpha}=0$ with $\eta_1\ne0$. Then (\ref{etacondition5}) yields  $\gamma=\alpha(1+\rho_3)/2$, and (\ref{cgammaeta}) yields }
\begin{align}\label{constantcpara}
&c\left(\alpha,\beta,\varepsilon_1\right)\notag\\
&=\min_k\Bigg\{ {\alpha}\mu,
1-\frac{\beta^2}{1-\alpha L}(1+\rho_3^{-1}) -  \frac{\alpha^2(1+\rho_3)|\mathcal{M}_c^{k}|^2\varepsilon_1 }{1-\alpha L}\Bigg\}.
\end{align}
\textcolor[rgb]{0.00,0.00,0.00}{Setting  $\rho_3=1$, $\delta\in(0,1)$, and  $\alpha=\frac{1-\delta}{L}$, gives }
\begin{align}
&\eta_1=\frac{1-\alpha L}{2\alpha}, \varepsilon_1=\frac{(1-\alpha L)(1-\alpha \mu)}{4\alpha^2M^2}, \notag\\
&\qquad\mbox{and}\  \beta=\frac{1}{2}\sqrt{(1-\alpha L)(1-\alpha\mu)},
\end{align}
\textcolor[rgb]{0.00,0.00,0.00}{and  (\ref{constantcpara}) yields }
\begin{align}
c\left(\alpha,\beta,\varepsilon_1\right)&=\min_k\Bigg\{ \alpha\mu, 1-\frac{1-\alpha\mu}{2}\left(1+\frac{|\mathcal{M}_c^{k}|^2}{M^2}\right)\Bigg\}\notag\\
&=\alpha\mu.
\end{align}
\textcolor[rgb]{0.00,0.00,0.00}{It follows that }
\begin{align}\label{iterationcomp}
\frac{\mathbb{L}(\boldsymbol{\theta}^{k+1})}{\mathbb{L}(\boldsymbol{\theta}^{1})}\le\left(1- \frac{1-\delta}{L/\mu}\right)^k\le\epsilon.
\end{align}
\textcolor[rgb]{0.00,0.00,0.00}{Rearranging terms in (\ref{iterationcomp}), yields }
\begin{align}\label{}
\log(\frac{1}{\epsilon})\le k\log\left(1+\frac{1}{\frac{L/\mu}{1-\delta}-1}\right)\le  \frac{k}{\frac{L/\mu}{1-\delta}-1}
\end{align}
\textcolor[rgb]{0.00,0.00,0.00}{such that the iteration complexity $\mathbb{I}_{CHB}(\epsilon)$ is}
\begin{align}\label{}
\mathbb{I}_{CHB}(\epsilon)= \frac{L/\mu}{1-\delta}\log(\frac{1}{\epsilon}).
\end{align}

\section{Proof of Theorem \ref{convergenceTheorem2}}\label{convexCase}
\textcolor[rgb]{0.00,0.00,0.00}{The following is inspired by the previous work in \cite{chen2018lag} which can be regarded as a special case of CHB without the momentum term (with $\beta=0$). We start with a lemma that will be helpful later.}
\begin{lemma}\label{}
Under \textcolor[rgb]{0.00,0.00,0.00}{Definition \ref{OHBassumption3}}, the Lyapunov function $\mathbb{L}(\boldsymbol{\theta}^{k})$ satisfies
\begin{align}
&\mathbb{L}(\boldsymbol{\theta}^{k})\le \sqrt{\Big(\|\nabla f(\boldsymbol{\theta}^k)\|^2 + \eta_{1}\|\boldsymbol{\theta}^{k}-\boldsymbol{\theta}^{k-1}\|^{2} \Big)}\notag\\
&\qquad\qquad\cdot\sqrt{\Big(\|\boldsymbol{\theta}^k-\boldsymbol{\theta}^*\|^2 + \eta_{1}\|\boldsymbol{\theta}^{k}-\boldsymbol{\theta}^{k-1}\|^{2}\Big)}.
\end{align}
\end{lemma}
\begin{IEEEproof}
\textcolor[rgb]{0.00,0.00,0.00}{Using  \textcolor[rgb]{0.00,0.00,0.00}{Definition \ref{OHBassumption3}} (convexity) }
\begin{align}\label{conveproer12}
f(\boldsymbol{\theta}^k)-f(\boldsymbol{\theta}^*)\le\left\langle\nabla f(\boldsymbol{\theta}^{k}), \boldsymbol{\theta}^{k}-\boldsymbol{\theta}^{*}\right\rangle.
\end{align}
\textcolor[rgb]{0.00,0.00,0.00}{Thus}
\begin{align}
&\mathbb{L}(\boldsymbol{\theta}^{k})\notag\\
&=f(\boldsymbol{\theta}^{k})-f\left(\boldsymbol{\theta}^{*}\right)+ \eta_{1}\|\boldsymbol{\theta}^{k}-\boldsymbol{\theta}^{k-1}\|^{2}\notag\\
&\le \left\langle\nabla f(\boldsymbol{\theta}^{k}), \boldsymbol{\theta}^{k}-\boldsymbol{\theta}^{*}\right\rangle \notag\\
&\qquad + \left\langle \sqrt{\eta_1}\|\boldsymbol{\theta}^{k}-\boldsymbol{\theta}^{k-1}\|, \sqrt{\eta_1}\|\boldsymbol{\theta}^{k}-\boldsymbol{\theta}^{k-1}\| \right\rangle\label{convexproperty1}\\
&=\Big\langle \left[\nabla f(\boldsymbol{\theta}^{k}),\sqrt{\eta_1}\|\boldsymbol{\theta}^{k}-\boldsymbol{\theta}^{k-1}\|\right]^{\top}, \notag\\
&\qquad \qquad \left[\boldsymbol{\theta}^{k}-\boldsymbol{\theta}^{*}, \sqrt{\eta_1}\|\boldsymbol{\theta}^{k}-\boldsymbol{\theta}^{k-1}\| \right]^{\top}\Big\rangle\label{innerproduct}\\
&\le\sqrt{\Big(\|\nabla f(\boldsymbol{\theta}^k)\|^2 + \eta_{1}\|\boldsymbol{\theta}^{k}-\boldsymbol{\theta}^{k-1}\|^{2} \Big)}\notag\\
&\qquad \qquad \cdot\sqrt{\Big(\|\boldsymbol{\theta}^k-\boldsymbol{\theta}^*\|^2 + \eta_{1}\|\boldsymbol{\theta}^{k}-\boldsymbol{\theta}^{k-1}\|^{2}\Big)}\label{finalpropert12}.
\end{align}
\textcolor[rgb]{0.00,0.00,0.00}{We obtained  (\ref{convexproperty1}) by using (\ref{conveproer12}). We use (\ref{innerproduct}) and the inequality $\langle\boldsymbol{x},\boldsymbol{y}\rangle\le\|\boldsymbol{x}\|\|\boldsymbol{y}\|$ to obtain (\ref{finalpropert12}). This completes the proof.}
\end{IEEEproof}
\textcolor[rgb]{0.00,0.00,0.00}{Given {Assumption \ref{OHBassumption1}} and assuming the constants $\alpha$, $\beta$, and $\varepsilon_1$ satisfy  (\ref{DDScondition1})--(\ref{etacondition5}) while   $\sigma_0>0$ and $\sigma_1>0$, then  (\ref{descentLyapunovequation}) of Lemma \ref{descentLyapunovLemma} suggests that}
\begin{align}
&\mathbb{L}(\boldsymbol{\theta}^{k+1}) -  \mathbb{L}(\boldsymbol{\theta}^{k}) \notag\\
&\le  -\sigma_0\left\|\nabla f\left(\boldsymbol{\theta}^{k}\right)\right\|^{2} - \sigma_1\left\| \boldsymbol{\theta}^{k}-\boldsymbol{\theta}^{k-1} \right\|^{2}\\
&\le -\min_k\{\sigma_0,\frac{\sigma_1}{\eta_1}\}\left(\left\|\nabla f\left(\boldsymbol{\theta}^{k}\right)\right\|^{2} + \eta_1\left\| \boldsymbol{\theta}^{k}-\boldsymbol{\theta}^{k-1} \right\|^{2}\right). \label{decreasLay112}
\end{align}
\textcolor[rgb]{0.00,0.00,0.00}{Further,  $\min_k\{\sigma_0,\frac{\sigma_1}{\eta_1}\}>0$
with $\sigma_0$ and $\sigma_1$ provided in (\ref{DDScondition1}) and (\ref{condition2}). Note that (\ref{decreasLay112})
yields }
\begin{align}\label{boudnfunctiondiff}
&\left\|\nabla f\left(\boldsymbol{\theta}^{k}\right)\right\|^{2} + \eta_1\left\| \boldsymbol{\theta}^{k}-\boldsymbol{\theta}^{k-1} \right\|^{2}\notag\\
&\le\frac{1}{\min_k\{\sigma_0,\sigma_1/\eta_1\}}\left( \mathbb{L}(\boldsymbol{\theta}^{k})-\mathbb{L}(\boldsymbol{\theta}^{k+1}) \right).
\end{align}
\textcolor[rgb]{0.00,0.00,0.00}{Assumption \ref{OHBassumption1} implies that 
$\left\|\boldsymbol{\theta}^{k}-\boldsymbol{\theta}^{*}\right\|<+\infty$ and $\left\|\boldsymbol{\theta}^{k}-\boldsymbol{\theta}^{k-1}\right\|<+\infty$. It follows that}
\begin{align}
\|\boldsymbol{\theta}^k-\boldsymbol{\theta}^*\|^2 + \eta_{1}\|\boldsymbol{\theta}^{k}-\boldsymbol{\theta}^{k-1}\|^{2}\le N. \label{boundfinite}
\end{align}
\textcolor[rgb]{0.00,0.00,0.00}{for finite $N$. We can use  (\ref{boudnfunctiondiff}) and (\ref{boundfinite}) in  (\ref{finalpropert12}) to show that }
\begin{align}
&(\mathbb{L}(\boldsymbol{\theta}^{k}))^2\notag\\
&\le \Big(\|\nabla f(\boldsymbol{\theta}^k)\|^2 + \eta_{1}\|\boldsymbol{\theta}^{k}-\boldsymbol{\theta}^{k-1}\|^{2} \Big)\notag\\
&\qquad\quad\Big(\|\boldsymbol{\theta}^k-\boldsymbol{\theta}^*\|^2 + \eta_{1}\|\boldsymbol{\theta}^{k}-\boldsymbol{\theta}^{k-1}\|^{2}\Big)\\
&\le\frac{N}{\min_k\{\sigma_0,\sigma_1/\eta_1\}}\left( \mathbb{L}(\boldsymbol{\theta}^{k})-\mathbb{L}(\boldsymbol{\theta}^{k+1}) \right).\label{desceprotty}
\end{align}
\textcolor[rgb]{0.00,0.00,0.00}{Lemma \ref{descentLyapunovLemma} implies that 
$\mathbb{L}(\boldsymbol{\theta}^{k})\mathbb{L}(\boldsymbol{\theta}^{k+1})\le(\mathbb{L}(\boldsymbol{\theta}^{k}))^2$ so that (\ref{desceprotty}) yields }
\begin{align}
\mathbb{L}(\boldsymbol{\theta}^{k})\mathbb{L}(\boldsymbol{\theta}^{k+1})\le\frac{N}{\min_k\{\sigma_0,\sigma_1/\eta_1\}}\left( \mathbb{L}(\boldsymbol{\theta}^{k})-\mathbb{L}(\boldsymbol{\theta}^{k+1}) \right)
\end{align}
\textcolor[rgb]{0.00,0.00,0.00}{or} 
\begin{align}\label{sumLay12}
\frac{1}{\mathbb{L}(\boldsymbol{\theta}^{k+1})} - \frac{1}{\mathbb{L}(\boldsymbol{\theta}^{k})} \ge\frac{\min_k\{\sigma_0,\sigma_1/\eta_1\}}{N}.
\end{align}
\textcolor[rgb]{0.00,0.00,0.00}{It follows that }
\begin{align}\label{}
\frac{1}{\mathbb{L}(\boldsymbol{\theta}^{K})}\ge\frac{1}{\mathbb{L}(\boldsymbol{\theta}^{K})} - \frac{1}{\mathbb{L}(\boldsymbol{\theta}^{0})} \ge\frac{K\min_k\{\sigma_0,\sigma_1/\eta_1\}}{N}, 
\end{align}
\textcolor[rgb]{0.00,0.00,0.00}{which completes the proof.}

\section{Proof of Theorem \ref{convergenceTheorem3}}\label{nonConvexCase}
\textcolor[rgb]{0.00,0.00,0.00}{This section again builds on \cite{chen2018lag}}.
\textcolor[rgb]{0.00,0.00,0.00}{From Lemma \ref{descentLyapunovLemma} }
\begin{align}
&\mathbb{L}(\boldsymbol{\theta}^{k+1}) -  \mathbb{L}(\boldsymbol{\theta}^{k}) \notag\\
&\le  -\sigma_0\left\|\nabla f\left(\boldsymbol{\theta}^{k}\right)\right\|^{2} - \sigma_1\left\| \boldsymbol{\theta}^{k}-\boldsymbol{\theta}^{k-1} \right\|^{2}\\
&\le -\min_k\{\sigma_0,\frac{\sigma_1}{\eta_1}\}\left(\left\|\nabla f\left(\boldsymbol{\theta}^{k}\right)\right\|^{2} + \eta_1\left\| \boldsymbol{\theta}^{k}-\boldsymbol{\theta}^{k-1} \right\|^{2}\right).\label{decreasLay11}
\end{align}
\textcolor[rgb]{0.00,0.00,0.00}{The last equation is  summed over $k$ from $1$ to $K$ to yield }
\begin{align}
&\mathbb{L}(\boldsymbol{\theta}^{1}) - \mathbb{L}(\boldsymbol{\theta}^{K+1})  \notag\\ &\ge\min_k\{\sigma_0,\frac{\sigma_1}{\eta_1}\}\sum_{k=1}^K\left(\left\|\nabla f\left(\boldsymbol{\theta}^{k}\right)\right\|^{2} + \eta_1\left\| \boldsymbol{\theta}^{k}-\boldsymbol{\theta}^{k-1} \right\|^{2}\right).
\end{align}
\textcolor[rgb]{0.00,0.00,0.00}{Note that  $\mathbb{L}(\boldsymbol{\theta}^{1}) - \mathbb{L}(\boldsymbol{\theta}^{K+1})\le\mathbb{L}(\boldsymbol{\theta}^{1})<\infty$ so that }
\begin{align}
\lim_{K\rightarrow\infty}\sum_{k=1}^K\left(\left\|\nabla f\left(\boldsymbol{\theta}^{k}\right)\right\|^{2} + \eta_1\left\| \boldsymbol{\theta}^{k}-\boldsymbol{\theta}^{k-1} \right\|^{2}\right)<\infty
\end{align}
\textcolor[rgb]{0.00,0.00,0.00}{or}
\begin{align}
\lim_{K\rightarrow\infty}\left\|\nabla f\left(\boldsymbol{\theta}^{K}\right)\right\|^{2}\rightarrow 0.
\end{align}

\section{Proof of Lemma \ref{ONBsavinglemma}}\label{CommSavingLemma}
Suppose that worker $m$ has communicated with the server at iteration $(k-1)$ which means $\hat{\boldsymbol{\theta}}_{m}^{k-1}=\boldsymbol{\theta}^{k-1}$, we can obtain
\begin{align}
\left\| \nabla f_m(\hat{\boldsymbol{\theta}}_{m}^{k-1}) - \nabla f_m({\boldsymbol{\theta}}^{k}) \right\|^{2} \le L_m^2\left\|\hat{\boldsymbol{\theta}}_{m}^{k-1}
-\boldsymbol{\theta}^{k}\right\|^{2}\label{propertysmooth}\\
=L_m^2\left\| \boldsymbol{\theta}^{k-1} - \boldsymbol{\theta}^{k} \right\|^{2}
\end{align}
where (\ref{propertysmooth}) is obtained using the definition of the smoothness constant in Assumption \ref{OHBassumption1}.

If (\ref{Lmstracincondition1}) is true, then $L_m^2\left\| \boldsymbol{\theta}^{k-1} - \boldsymbol{\theta}^{k} \right\|^{2}$ can be bounded by $\varepsilon_1\left\| \boldsymbol{\theta}^{k-1} - \boldsymbol{\theta}^{k} \right\|^{2}$. Thus, we have
\begin{align}\label{sto1ppingcondi}
\left\| \nabla f_m(\hat{\boldsymbol{\theta}}_{m}^{k-1}) - \nabla f_m({\boldsymbol{\theta}}^{k}) \right\|^{2}\le\varepsilon_1\left\| \boldsymbol{\theta}^{k-1} - \boldsymbol{\theta}^{k} \right\|^{2}
\end{align}
which is equivalent to the CHB-skip-transmission condition in (\ref{CHB-stopping condition}). The result in (\ref{sto1ppingcondi}) indicates that  worker $m$ will not transmit at iteration $k$ since the CHB-skip-transmission condition is satisfied. Thus, in the first $k$ iterations, the worker $m$ communicates with the server at most $k/2$ times if (\ref{Lmstracincondition1}) holds.

\ifCLASSOPTIONcaptionsoff
  \newpage
\fi



%
%
%

\bibliographystyle{IEEEtran}
\bibliography{refs}

\begin{thebibliography}{10}
\providecommand{\url}[1]{#1}
\csname url@samestyle\endcsname
\providecommand{\newblock}{\relax}
\providecommand{\bibinfo}[2]{#2}
\providecommand{\BIBentrySTDinterwordspacing}{\spaceskip=0pt\relax}
\providecommand{\BIBentryALTinterwordstretchfactor}{4}
\providecommand{\BIBentryALTinterwordspacing}{\spaceskip=\fontdimen2\font plus
\BIBentryALTinterwordstretchfactor\fontdimen3\font minus
  \fontdimen4\font\relax}
\providecommand{\BIBforeignlanguage}[2]{{%
\expandafter\ifx\csname l@#1\endcsname\relax
\typeout{** WARNING: IEEEtran.bst: No hyphenation pattern has been}%
\typeout{** loaded for the language `#1'. Using the pattern for}%
\typeout{** the default language instead.}%
\else
\language=\csname l@#1\endcsname
\fi
#2}}
\providecommand{\BIBdecl}{\relax}
\BIBdecl

\bibitem{liu2021approximate}
S.~Liu, N.~Gupta, and N.~H. Vaidya, ``Approximate byzantine fault-tolerance in
  distributed optimization,'' in \emph{Proceedings of the 2021 ACM Symposium on
  Principles of Distributed Computing}, 2021, pp. 379--389.

\bibitem{turan2022robust}
B.~Turan, C.~A. Uribe, H.-T. Wai, and M.~Alizadeh, ``Robust distributed
  optimization with randomly corrupted gradients,'' \emph{IEEE Transactions on
  Signal Processing}, 2022.

\bibitem{li2021distributed}
X.~Li, L.~Xie, and Y.~Hong, ``Distributed aggregative optimization over
  multi-agent networks,'' \emph{IEEE Transactions on Automatic Control}, 2021.

\bibitem{zhang2015distributed}
Y.~Zhang, Y.~Lou, Y.~Hong, and L.~Xie, ``Distributed projection-based
  algorithms for source localization in wireless sensor networks,'' \emph{IEEE
  Transactions on Wireless Communications}, vol.~14, no.~6, pp. 3131--3142,
  2015.

\bibitem{chen2019testing}
Y.~Chen, R.~S. Blum, B.~M. Sadler, and J.~Zhang, ``Testing the structure of a
  gaussian graphical model with reduced transmissions in a distributed
  setting,'' \emph{IEEE Transactions on Signal Processing}, vol.~67, no.~20,
  pp. 5391--5401, 2019.

\bibitem{chen2018ordered}
Y.~Chen, B.~M. Sadler, and R.~S. Blum, ``Ordered transmission for efficient
  wireless autonomy,'' in \emph{2018 52nd Asilomar Conference on Signals,
  Systems, and Computers}.\hskip 1em plus 0.5em minus 0.4em\relax IEEE, 2018,
  pp. 1299--1303.

\bibitem{chen2020optimal}
Y.~Chen, R.~S. Blum, and B.~M. Sadler, ``Optimal quickest change detection in
  sensor networks using ordered transmissions,'' in \emph{2020 IEEE 21st
  International Workshop on Signal Processing Advances in Wireless
  Communications (SPAWC)}.\hskip 1em plus 0.5em minus 0.4em\relax IEEE, 2020,
  pp. 1--5.

\bibitem{chen2021ordering}
------, ``Ordering for communication-efficient quickest change detection in a
  decomposable graphical model,'' \emph{IEEE Transactions on Signal
  Processing}, vol.~69, pp. 4710--4723, 2021.

\bibitem{binetti2014distributed}
G.~Binetti, A.~Davoudi, F.~L. Lewis, D.~Naso, and B.~Turchiano, ``Distributed
  consensus-based economic dispatch with transmission losses,'' \emph{IEEE
  Transactions on Power Systems}, vol.~29, no.~4, pp. 1711--1720, 2014.

\bibitem{park2021communication}
J.~Park, S.~Samarakoon, A.~Elgabli, J.~Kim, M.~Bennis, S.-L. Kim, and
  M.~Debbah, ``Communication-efficient and distributed learning over wireless
  networks: Principles and applications,'' \emph{Proceedings of the IEEE}, vol.
  109, no.~5, pp. 796--819, 2021.

\bibitem{vlaski2021distributed}
S.~Vlaski and A.~H. Sayed, ``Distributed learning in non-convex
  environments—part i: Agreement at a linear rate,'' \emph{IEEE Transactions
  on Signal Processing}, vol.~69, pp. 1242--1256, 2021.

\bibitem{imteaj2021survey}
A.~Imteaj, U.~Thakker, S.~Wang, J.~Li, and M.~H. Amini, ``A survey on federated
  learning for resource-constrained iot devices,'' \emph{IEEE Internet of
  Things Journal}, vol.~9, no.~1, pp. 1--24, 2021.

\bibitem{chen2022communication}
Y.~Chen, R.~S. Blum, and B.~M. Sadler, ``Communication efficient federated
  learning via ordered admm in a fully decentralized setting,'' in \emph{2022
  56th Annual Conference on Information Sciences and Systems (CISS)}.\hskip 1em
  plus 0.5em minus 0.4em\relax IEEE, 2022, pp. 96--100.

\bibitem{smith2017cocoa}
V.~Smith, S.~Forte, C.~Ma, M.~Tak{\'a}{\v{c}}, M.~I. Jordan, and M.~Jaggi,
  ``Cocoa: A general framework for communication-efficient distributed
  optimization,'' \emph{The Journal of Machine Learning Research}, vol.~18,
  no.~1, pp. 8590--8638, 2017.

\bibitem{chen2021ordered}
Y.~Chen, ``Ordered transmissions, estimation, and parameter learning,'' Ph.D.
  dissertation, Lehigh University, 2021.

\bibitem{zhou2021communication}
Y.~Zhou, Q.~Ye, and J.~Lv, ``Communication-efficient federated learning with
  compensated overlap-fedavg,'' \emph{IEEE Transactions on Parallel and
  Distributed Systems}, vol.~33, no.~1, pp. 192--205, 2021.

\bibitem{ozfatura2021time}
E.~Ozfatura, K.~Ozfatura, and D.~G{\"u}nd{\"u}z, ``Time-correlated
  sparsification for communication-efficient federated learning,'' in
  \emph{2021 IEEE International Symposium on Information Theory (ISIT)}.\hskip
  1em plus 0.5em minus 0.4em\relax IEEE, 2021, pp. 461--466.

\bibitem{jhunjhunwala2021adaptive}
D.~Jhunjhunwala, A.~Gadhikar, G.~Joshi, and Y.~C. Eldar, ``Adaptive
  quantization of model updates for communication-efficient federated
  learning,'' in \emph{ICASSP 2021-2021 IEEE International Conference on
  Acoustics, Speech and Signal Processing (ICASSP)}.\hskip 1em plus 0.5em minus
  0.4em\relax IEEE, 2021, pp. 3110--3114.

\bibitem{li2021communication}
C.~Li, G.~Li, and P.~K. Varshney, ``Communication-efficient federated learning
  based on compressed sensing,'' \emph{IEEE Internet of Things Journal},
  vol.~8, no.~20, pp. 15\,531--15\,541, 2021.

\bibitem{gao2021convergence}
H.~Gao, A.~Xu, and H.~Huang, ``On the convergence of communication-efficient
  local sgd for federated learning,'' in \emph{Proceedings of the AAAI
  Conference on Artificial Intelligence}, vol.~35, no.~9, 2021, pp. 7510--7518.

\bibitem{chen2022distributed}
Y.~Chen, R.~S. Blum, M.~Tak{\'a}{\v{c}}, and B.~M. Sadler, ``Distributed
  learning with sparsified gradient differences,'' \emph{IEEE Journal of
  Selected Topics in Signal Processing}, vol.~16, no.~3, pp. 585--600, 2022.

\bibitem{chen2020ordered}
Y.~Chen, B.~M. Sadler, and R.~S. Blum, ``Ordered gradient approach for
  communication-efficient distributed learning,'' in \emph{2020 IEEE 21st
  International Workshop on Signal Processing Advances in Wireless
  Communications (SPAWC)}.\hskip 1em plus 0.5em minus 0.4em\relax IEEE, 2020,
  pp. 1--5.

\bibitem{sattler2021cfd}
F.~Sattler, A.~Marban, R.~Rischke, and W.~Samek, ``Cfd: Communication-efficient
  federated distillation via soft-label quantization and delta coding,''
  \emph{IEEE Transactions on Network Science and Engineering}, vol.~9, no.~4,
  pp. 2025--2038, 2021.

\bibitem{honig2022dadaquant}
R.~H{\"o}nig, Y.~Zhao, and R.~Mullins, ``Dadaquant: Doubly-adaptive
  quantization for communication-efficient federated learning,'' in
  \emph{International Conference on Machine Learning}.\hskip 1em plus 0.5em
  minus 0.4em\relax PMLR, 2022, pp. 8852--8866.

\bibitem{fan2021communication}
X.~Fan, Y.~Wang, Y.~Huo, and Z.~Tian, ``Communication-efficient federated
  learning through 1-bit compressive sensing and analog aggregation,'' in
  \emph{2021 IEEE International Conference on Communications Workshops (ICC
  Workshops)}.\hskip 1em plus 0.5em minus 0.4em\relax IEEE, 2021, pp. 1--6.

\bibitem{wang2021quantized}
Y.~Wang, Y.~Xu, Q.~Shi, and T.-H. Chang, ``Quantized federated learning under
  transmission delay and outage constraints,'' \emph{IEEE Journal on Selected
  Areas in Communications}, vol.~40, no.~1, pp. 323--341, 2021.

\bibitem{elkordy2022heterosag}
A.~R. Elkordy and A.~S. Avestimehr, ``Heterosag: Secure aggregation with
  heterogeneous quantization in federated learning,'' \emph{IEEE Transactions
  on Communications}, vol.~70, no.~4, pp. 2372--2386, 2022.

\bibitem{chen2021dynamic}
S.~Chen, C.~Shen, L.~Zhang, and Y.~Tang, ``Dynamic aggregation for
  heterogeneous quantization in federated learning,'' \emph{IEEE Transactions
  on Wireless Communications}, vol.~20, no.~10, pp. 6804--6819, 2021.

\bibitem{panda2022sparsefed}
A.~Panda, S.~Mahloujifar, A.~N. Bhagoji, S.~Chakraborty, and P.~Mittal,
  ``Sparsefed: Mitigating model poisoning attacks in federated learning with
  sparsification,'' in \emph{International Conference on Artificial
  Intelligence and Statistics}.\hskip 1em plus 0.5em minus 0.4em\relax PMLR,
  2022, pp. 7587--7624.

\bibitem{mitra2021linear}
A.~Mitra, R.~Jaafar, G.~J. Pappas, and H.~Hassani, ``Linear convergence in
  federated learning: Tackling client heterogeneity and sparse gradients,''
  \emph{Advances in Neural Information Processing Systems}, vol.~34, pp.
  14\,606--14\,619, 2021.

\bibitem{stich2018sparsified}
S.~U. Stich, J.-B. Cordonnier, and M.~Jaggi, ``Sparsified sgd with memory,'' in
  \emph{Advances in Neural Information Processing Systems}, 2018, pp.
  4447--4458.

\bibitem{zinkevich2010parallelized}
M.~Zinkevich, M.~Weimer, L.~Li, and A.~J. Smola, ``Parallelized stochastic
  gradient descent,'' in \emph{Advances in neural information processing
  systems}, 2010, pp. 2595--2603.

\bibitem{zhang2013communication}
Y.~Zhang, J.~C. Duchi, and M.~J. Wainwright, ``Communication-efficient
  algorithms for statistical optimization,'' \emph{The Journal of Machine
  Learning Research}, vol.~14, no.~1, pp. 3321--3363, 2013.

\bibitem{mcdonald2009efficient}
R.~Mcdonald, M.~Mohri, N.~Silberman, D.~Walker, and G.~S. Mann, ``Efficient
  large-scale distributed training of conditional maximum entropy models,'' in
  \emph{Advances in neural information processing systems}, 2009, pp.
  1231--1239.

\bibitem{mcmahan2016federated}
H.~B. McMahan, E.~Moore, D.~Ramage, and B.~A. y~Arcas, ``Federated learning of
  deep networks using model averaging. corr abs/1602.05629 (2016),''
  \emph{arXiv preprint arXiv:1602.05629}, 2016.

\bibitem{shamir2014communication}
O.~Shamir, N.~Srebro, and T.~Zhang, ``Communication-efficient distributed
  optimization using an approximate newton-type method,'' in
  \emph{International conference on machine learning}, 2014, pp. 1000--1008.

\bibitem{agarwal2014reliable}
A.~Agarwal, O.~Chapelle, M.~Dud{\'\i}k, and J.~Langford, ``A reliable effective
  terascale linear learning system,'' \emph{The Journal of Machine Learning
  Research}, vol.~15, no.~1, pp. 1111--1133, 2014.

\bibitem{bordes2009sgd}
A.~Bordes, L.~Bottou, and P.~Gallinari, ``Sgd-qn: Careful quasi-newton
  stochastic gradient descent,'' \emph{Journal of Machine Learning Research},
  vol.~10, no. Jul, pp. 1737--1754, 2009.

\bibitem{byrd2016stochastic}
R.~H. Byrd, S.~L. Hansen, J.~Nocedal, and Y.~Singer, ``A stochastic
  quasi-newton method for large-scale optimization,'' \emph{SIAM Journal on
  Optimization}, vol.~26, no.~2, pp. 1008--1031, 2016.

\bibitem{moritz2016linearly}
P.~Moritz, R.~Nishihara, and M.~Jordan, ``A linearly-convergent stochastic
  l-bfgs algorithm,'' in \emph{Artificial Intelligence and Statistics}, 2016,
  pp. 249--258.

\bibitem{gower2016stochastic}
R.~Gower, D.~Goldfarb, and P.~Richt{\'a}rik, ``Stochastic block bfgs: Squeezing
  more curvature out of data,'' in \emph{International Conference on Machine
  Learning}, 2016, pp. 1869--1878.

\bibitem{zhang2015disco}
Y.~Zhang and X.~Lin, ``Disco: Distributed optimization for self-concordant
  empirical loss,'' in \emph{International conference on machine learning},
  2015, pp. 362--370.

\bibitem{reddi2016aide}
S.~J. Reddi, J.~Kone{\v{c}}n{\`y}, P.~Richt{\'a}rik, B.~P{\'o}cz{\'o}s, and
  A.~Smola, ``Aide: Fast and communication efficient distributed
  optimization,'' \emph{arXiv preprint arXiv:1608.06879}, 2016.

\bibitem{duchi2011dual}
J.~C. Duchi, A.~Agarwal, and M.~J. Wainwright, ``Dual averaging for distributed
  optimization: Convergence analysis and network scaling,'' \emph{IEEE
  Transactions on Automatic control}, vol.~57, no.~3, pp. 592--606, 2011.

\bibitem{scaman2018optimal}
K.~Scaman, F.~Bach, S.~Bubeck, L.~Massouli{\'e}, and Y.~T. Lee, ``Optimal
  algorithms for non-smooth distributed optimization in networks,'' in
  \emph{Advances in Neural Information Processing Systems}, 2018, pp.
  2740--2749.

\bibitem{he2018cola}
L.~He, A.~Bian, and M.~Jaggi, ``Cola: Decentralized linear learning,'' in
  \emph{Advances in Neural Information Processing Systems}, 2018, pp.
  4536--4546.

\bibitem{rago1996censoring}
C.~Rago, P.~Willett, and Y.~Bar-Shalom, ``Censoring sensors: A
  low-communication-rate scheme for distributed detection,'' \emph{IEEE
  Transactions on Aerospace and Electronic Systems}, vol.~32, no.~2, pp.
  554--568, 1996.

\bibitem{appadwedula2005energy}
S.~Appadwedula, V.~V. Veeravalli, and D.~L. Jones, ``Energy-efficient detection
  in sensor networks,'' \emph{IEEE Journal on Selected areas in
  communications}, vol.~23, no.~4, pp. 693--702, 2005.

\bibitem{marano2006cross}
S.~Marano, V.~Matta, P.~Willett, and L.~Tong, ``Cross-layer design of
  sequential detectors in sensor networks,'' \emph{IEEE Transactions on Signal
  Processing}, vol.~54, no.~11, pp. 4105--4117, 2006.

\bibitem{patwari2003hierarchical}
N.~Patwari, A.~Hero, and B.~M. Sadler, ``Hierarchical censoring sensors for
  change detection,'' in \emph{IEEE Workshop on Statistical Signal Processing,
  2003}.\hskip 1em plus 0.5em minus 0.4em\relax IEEE, 2003, pp. 21--24.

\bibitem{chen2016event}
W.~Chen and W.~Ren, ``Event-triggered zero-gradient-sum distributed consensus
  optimization over directed networks,'' \emph{Automatica}, vol.~65, pp.
  90--97, 2016.

\bibitem{lu2017event}
Q.~L{\"u} and H.~Li, ``Event-triggered discrete-time distributed consensus
  optimization over time-varying graphs,'' \emph{Complexity}, vol. 2017, 2017.

\bibitem{tsianos2013networked}
K.~I. Tsianos, S.~F. Lawlor, J.~Y. Yu, and M.~G. Rabbat, ``Networked
  optimization with adaptive communication,'' in \emph{2013 IEEE Global
  Conference on Signal and Information Processing}.\hskip 1em plus 0.5em minus
  0.4em\relax IEEE, 2013, pp. 579--582.

\bibitem{chen2018lag}
T.~Chen, G.~Giannakis, T.~Sun, and W.~Yin, ``Lag: Lazily aggregated gradient
  for communication-efficient distributed learning,'' in \emph{Advances in
  Neural Information Processing Systems}, 2018, pp. 5050--5060.

\bibitem{leen1994optimal}
T.~K. Leen and G.~B. Orr, ``Optimal stochastic search and adaptive momentum,''
  in \emph{Advances in neural information processing systems}, 1994, pp.
  477--484.

\bibitem{sutskever2013importance}
I.~Sutskever, J.~Martens, G.~Dahl, and G.~Hinton, ``On the importance of
  initialization and momentum in deep learning,'' in \emph{International
  conference on machine learning}, 2013, pp. 1139--1147.

\bibitem{Bottou:2018}
L.~Bottou, F.~E. Curtis, and J.~Nocedal, ``Optimization methods for large-scale
  machine learning,'' \emph{Siam Review}, vol.~60, no.~2, pp. 223--311, 2018.

\bibitem{nesterov2018lectures}
Y.~Nesterov, \emph{Lectures on convex optimization}.\hskip 1em plus 0.5em minus
  0.4em\relax Springer, 2018, vol. 137.

\bibitem{peressini1988mathematics}
A.~L. Peressini, F.~E. Sullivan, and J.~J. Uhl, \emph{The mathematics of
  nonlinear programming}.\hskip 1em plus 0.5em minus 0.4em\relax
  Springer-Verlag New York, 1988.

\bibitem{Ghadimi:2015}
E.~Ghadimi, H.~R. Feyzmahdavian, and M.~Johansson, ``Global convergence of the
  heavy-ball method for convex optimization,'' \emph{arXiv preprint
  arXiv:1412.7457}, 2014.

\bibitem{geron2019hands}
A.~G{\'e}ron, \emph{Hands-on machine learning with Scikit-Learn, Keras, and
  TensorFlow: Concepts, tools, and techniques to build intelligent
  systems}.\hskip 1em plus 0.5em minus 0.4em\relax O'Reilly Media, 2019.

\bibitem{prokhorov2001ijcnn}
D.~Prokhorov, ``Ijcnn 2001 neural network competition,'' \emph{Slide
  presentation in IJCNN}, vol.~1, no.~97, p.~38, 2001.

\bibitem{lecun1998gradient}
Y.~LeCun, L.~Bottou, Y.~Bengio, and P.~Haffner, ``Gradient-based learning
  applied to document recognition,'' \emph{Proceedings of the IEEE}, vol.~86,
  no.~11, pp. 2278--2324, 1998.

\end{thebibliography}

%








\end{document}